\documentclass[11pt]{article}
\usepackage[utf8]{inputenc}
\usepackage{float}

\usepackage{soul} %
\usepackage{setspace} 
\usepackage{shellesc}
\usepackage{mathrsfs,amsthm,xcolor,verbatim,bbm,amsmath,amsfonts,amssymb,nicefrac,enumitem,hyperref,bm,mathtools,xparse,etoolbox,textcomp,comment,cite}
\usepackage[outer=2cm, inner=3cm,top=4cm,bottom=5cm, twoside, a4paper]{geometry}
\usepackage[sort,capitalise]{cleveref} 
\usepackage{xurl}
\usepackage{frontespizio}
\usepackage{caption, booktabs} %
\usepackage{subfigure}
\counterwithin{figure}{section}
\counterwithin{table}{section}

\usepackage[disable]{todonotes}
\newcommand{\todoc}[1]{}
\newcommand{\todoDavide}[1]{}
\newcommand{\todoPhil}[1]{}

\newcommand{\todoSecond}[1]{}

\usepackage{tikz}
\usetikzlibrary{matrix}
\usepackage{tikz-3dplot}
\usetikzlibrary{automata,chains}

\usepackage{caption, booktabs}
\usepackage{standalone}
\usepackage{import}
\usetikzlibrary{quotes,arrows.meta}
\usetikzlibrary{positioning}

\usepackage{Ball}
\usepackage{Box}
\usepackage{RightBandedBox}

\crefname{enumi}{item}{items}

\crefname{equation}{}{}
\crefname{subsection}{Subsection}{Subsections}
\crefname{figure}{Figure}{Figures}

\definecolor{darkblue}{rgb}{0,0,0.75}
\definecolor{darkgreen}{rgb}{0,0.75,0}
\definecolor{darkred}{rgb}{0.75,0,0}
\hypersetup{colorlinks=true,linkcolor=darkblue,citecolor=darkgreen,urlcolor=darkblue}

\theoremstyle{plain}
\newtheorem{theorem}{Theorem}[section]
\newtheorem{lemma}[theorem]{Lemma}
\newtheorem{prop}[theorem]{Proposition}
\newtheorem{cor}[theorem]{Corollary}

\newtheorem{setting}[theorem]{Setting}
\newtheorem{method}[theorem]{Method}

\newtheorem{definition}[theorem]{Definition}

\newtheorem{remark}[theorem]{Remark}

\theoremstyle{definition}

\DeclareMathAlphabet{\mathpzc}{OT1}{pzc}{m}{it}

\DeclareFontEncoding{LS1}{}{}
\DeclareFontSubstitution{LS1}{stix}{m}{n}
\DeclareMathAlphabet{\mathscr}{LS1}{stixscr}{m}{n}

\newcommand{\E}{\mathbb{E}}

\newcommand{\R}{\mathbb{R}}
\newcommand{\N}{\mathbb{N}}

\renewcommand{\d}[1]{\mathfrak{d}^{#1}}

\DeclarePairedDelimiterX{\infdivx}[2]{(}{)}{%
   #1\;\delimsize\|\;#2 %
}
\newcommand{\infdivcomplex}{\cfadd{def:KL_divergence}D_{KL}}
\DeclarePairedDelimiterXPP{\infdiv}[2]{\cfadd{def:KL_divergence}D_{KL}}{(}{)}{}{#1\|#2} %
\newcommand{\Eb}[2]{\E_{#1}\!\left[#2\right]}

\DeclarePairedDelimiterXPP{\infdivdisc}[2]{\cfadd{def:KL_divergence_discrete}D_{KL}}{(}{)}{}{#1\|#2} %

\newcommand{\smallsum}{\textstyle\sum}

\newcommand{\Prob}[1]{\mathbb{P}^{#1}} %
\newcommand{\Xt}[1]{X^{#1}} %
\newcommand{\Zt}[1]{Z_{#1}} %

\newcommand{\pjoint}[1]{p^{#1}} %
\newcommand{\p}[2]{\mathfrak{p}^{#1}_{#2}} %
\newcommand{\pcond}[3]{\mathscr{p}^{#1}_{#2\vert#3}} %
\newcommand{\argpcond}[2]{{#1\vert#2}} 
\newcommand{\n}{\mathcal{N}\cfadd{def:gaussian}} %
\newcommand{\f}[1]{f^{#1}} %
\newcommand{\noise}{\mathcal{E}} %
\newcommand{\tk}{\tau} %
\newcommand{\opttheta}{\vartheta}

\newcommand{\bern}{\mathcal{B}} %
\DeclarePairedDelimiterXPP{\negloglike}[2]{\cfadd{def:exp_neg_log}\mathcal{H}}{(}{)}{}{#1\|#2} %
\newcommand{\sis}{\mathcal{S}} %

\newcommand{\nn}{N}  %

\newcommand{\tal}{\tilde{\alpha}} %
\newcommand{\al}{\alpha}

\newcommand{\tbe}{\tilde{\beta}}
\newcommand{\bV}{\mathbbm{V}} %
\newcommand{\bv}{\mathbbm{v}} %

\newcommand{\datax}{\mathcal{X}} %
\newcommand{\datac}{\mathcal{C}} %
\newcommand{\datal}{\mathcal{Y}} %

\newcommand{\Di}{D} %
\newcommand{\di}{d} %
\newcommand{\randT}{\mathcal{T}} %
\newcommand{\backX}{X} %
\newcommand{\bI}{\mathbbm{I}} %

\newcommand{\ds}{M}  %

\newcommand{\inx}{\mathscr{x}}
\newcommand{\iny}{\mathscr{y}}
\newcommand{\heads}{\mathscr{h}}
\newcommand{\headsdim}{\mathscr{d}}
\newcommand{\context}{c}
\newcommand{\maxtok}{l}
\newcommand{\emb}{e}

\newcommand{\cB}{\mathcal{B}}

\newcommand{\cD}{\mathcal{D}}
\newcommand{\cE}{\mathcal{E}}
\newcommand{\cF}{\mathcal{F}}

\newcommand{\fG}{\mathfrak{G}}

\newcommand{\fL}{\mathfrak{L}}

\newcommand{\fd}{\mathfrak{d}}

\renewcommand{\emptyset}{\varnothing}

\renewcommand{\P}{\mathbb{P}}

\DeclarePairedDelimiter{\norm}{\lVert}{\rVert}

\DeclarePairedDelimiter{\pr}{(}{)}

\renewcommand{\d}{ \mathrm{d}}

\newcommand{\qand}{\qquad\text{and}}
\newcommand{\qandq}{\qquad\text{and}\qquad}
\newcommand{\andq}{\text{and}\qquad}

\ExplSyntaxOn

\ExplSyntaxOn

\bool_new:N \g_noteobserve

\NewDocumentCommand{\setnote}{}{
  \bool_gset_true:N \g_noteobserve
}

\NewDocumentCommand{\setobserve}{}{
  \bool_gset_false:N \g_noteobserve
}

\NewDocumentCommand{\nobs}{ o }{
  \IfValueT{#1}{
    \str_if_eq:noTF {note} {#1} {
      \bool_gset_true:N \g_noteobserve
    } {
      \str_if_eq:noTF {Note} {#1} {
        \bool_gset_true:N \g_noteobserve
      } {
        \bool_gset_false:N \g_noteobserve
      }
    }
  }
  \bool_if:nTF { \g_noteobserve } {
    \bool_gset_false:N \g_noteobserve
    note
  } {
    \bool_gset_true:N \g_noteobserve
    observe
  }
  \IfValueF{#1}{~}
}

\NewDocumentCommand{\Nobs}{ o }{
  \IfValueT{#1}{
    \str_if_eq:noTF {note} {#1} {
      \bool_gset_true:N \g_noteobserve
    } {
      \str_if_eq:noTF {Note} {#1} {
        \bool_gset_true:N \g_noteobserve
      } {
        \bool_gset_false:N \g_noteobserve
      }
    }
  }
  \bool_if:nTF { \g_noteobserve } {
    \bool_gset_false:N \g_noteobserve
    Note
  } {
    \bool_gset_true:N \g_noteobserve
    Observe
  }
  \IfValueF{#1}{~}
}

\int_new:N \g_furthermore

\NewDocumentCommand{\Moreover}{ o o }{
  \IfValueT{#1}{
    \str_case:nn {#1} {
      {Furthermore} {\int_set:Nn {\g_furthermore} {0}}
      {Moreover} {\int_set:Nn {\g_furthermore} {1}}
      {In~addition} {\int_set:Nn {\g_furthermore} {2}}
      {note} {\bool_gset_true:N \g_noteobserve}
      {observe} {\bool_gset_false:N \g_noteobserve}
    }
    \IfValueT{#2}{
      \str_case:nn {#2} {
        {Furthermore} {\int_set:Nn {\g_furthermore} {0}}
        {Moreover} {\int_set:Nn {\g_furthermore} {1}}
        {In~addition} {\int_set:Nn {\g_furthermore} {2}}
        {note} {\bool_gset_true:N \g_noteobserve}
        {observe} {\bool_gset_false:N \g_noteobserve}
      }
    }
  }
  \int_case:nn { \int_mod:nn {\g_furthermore} {3} } {
    { 0 } { Furthermore,~\nobs that}
    { 1 } { Moreover,~\nobs that}
    { 2 } { In~addition,~\nobs that}
  }
  \int_incr:N \g_furthermore
  \IfValueF{#1}{~}
}

\bool_new:N \g_hencetherefore

\NewDocumentCommand{\hence}{}{
  \bool_if:nTF { \g_hencetherefore } {
    \bool_gset_false:N \g_hencetherefore
    hence~
  } {
    \bool_gset_true:N \g_hencetherefore
    therefore~
  }
}

\NewDocumentCommand{\Hence}{}{
  \bool_if:nTF { \g_hencetherefore } {
    \bool_gset_false:N \g_hencetherefore
    Hence,~we~obtain~
  } {
    \bool_gset_true:N \g_hencetherefore
    Therefore,~we~obtain~
  }
}

\ExplSyntaxOff

\ExplSyntaxOn

\seq_new:N \g_cflist_loaded
\seq_new:N \g_cflist_pending

\NewDocumentCommand{\cfadd} { m } {
  \seq_if_in:NnF \g_cflist_loaded { #1 } {
    \seq_if_in:NnF \g_cflist_pending { #1 } {
      \seq_gput_right:Nn \g_cflist_pending { #1 }
    }
  }
}

\NewDocumentCommand{\cfconsiderloaded} { m } {
  \seq_gput_right:Nn \g_cflist_loaded {#1}
}

\NewDocumentCommand{\cfremove} { m } {
  \seq_gremove_all:Nn \g_cflist_pending { #1 }
}

\NewDocumentCommand{\cfload} { o } {
  \seq_if_empty:NTF \g_cflist_pending {\unskip\IfValueT{#1}{\ignorespaces}} {
    (cf.\ \cref{\seq_use:Nn \g_cflist_pending {,}})\IfValueTF{#1}{#1~}{\unskip}
    \seq_gconcat:NNN \g_cflist_loaded \g_cflist_loaded \g_cflist_pending
    \seq_gclear:N \g_cflist_pending
    \IfValueT{#1}{\ignorespaces}
  }
}

\NewDocumentCommand{\cfclear} {} {
  \seq_gclear:N \g_cflist_loaded
  \seq_gclear:N \g_cflist_pending
}

\NewDocumentCommand{\cfout} { o } {
  \seq_if_empty:NTF \g_cflist_pending {\unskip\IfValueT{#1}{\ignorespaces}} {
    (cf.\ \cref{\seq_use:Nn \g_cflist_pending {,}})\IfValueTF{#1}{#1~}{\unskip}
    \seq_gclear:N \g_cflist_pending
    \IfValueT{#1}{\ignorespaces}
  }
}

\NewDocumentCommand{\ifnocf} { m } {
  \seq_if_empty:NT \g_cflist_pending { #1 }
}

\ExplSyntaxOff

\ExplSyntaxOn

\bool_new:N \g_forexample

\NewDocumentCommand{\eg}{ o }{
\IfValueT{#1}{
\str_if_eq:noTF {fe} {#1} {
\bool_gset_true:N \g_forexample
} {\bool_gset_false:N \g_forexample}
}
\bool_if:nTF { \g_forexample } {
\bool_gset_false:N \g_forexample
for~example,~
}{
\bool_gset_true:N \g_forexample
for~instance,~
}
}

\ExplSyntaxOff

\ExplSyntaxOn

\bool_new:N \g_Forexample

\NewDocumentCommand{\Eg}{ o }{
\IfValueT{#1}{
\str_if_eq:noTF {fe} {#1} {
\bool_gset_true:N \g_forexample
} {\bool_gset_false:N \g_forexample}
}
\bool_if:nTF { \g_forexample } {
\bool_gset_false:N \g_forexample
For~example,~
}{
\bool_gset_true:N \g_forexample
For~instance,~
}
}

\ExplSyntaxOff

\NewDocumentEnvironment{cproof}{m}
{\begin{proof}[Proof of \cref{#1}]}%
{\noindent The proof of~\cref{#1} is thus complete.
\end{proof}}

\NewDocumentEnvironment{cproof2}{m}
{\begin{proof}[Proof of \cref{#1}]}%
{\noindent This completes the proof of~\cref{#1}.
\end{proof}}

\makeatletter
\ExplSyntaxOn
\seq_new:N \g_abbrs
\prop_new:N \g_abbr_counts
\tl_new:N \l_abbr_count_tl

\NewDocumentCommand{\abbr}{m m O{#1} m m O{#4}}{
    \expandafter\newcommand\csname#3\endcsname[1][]{
        \seq_if_in:NnTF \g_abbrs {#1} {
            \prop_get:NnN \g_abbr_counts {#1} \l_abbr_count_tl
            \prop_gput:Nnx \g_abbr_counts {#1} {\int_eval:n {\l_abbr_count_tl + 1}}
            \hyperref[#1]{#1}
        } {
            \seq_gput_left:Nn \g_abbrs {#1}
            \prop_gput:Nnn \g_abbr_counts {#1} {1}
            \expandafter\gdef\csname#1@def\endcsname{#2}
            \phantomsection\label{#1}
            \str_if_eq:nnTF{##1}{}{\emph{#2}}{##1}~(\hyperref[#1]{#1})
        }
    }
    \expandafter\newcommand\csname#6\endcsname[1][]{
        \seq_if_in:NnTF \g_abbrs {#1} {
            \prop_get:NnN \g_abbr_counts {#1} \l_abbr_count_tl
            \prop_gput:Nnx \g_abbr_counts {#1} {\int_eval:n {\l_abbr_count_tl + 1}}
            \hyperref[#1]{#4}
        } {
            \expandafter\gdef\csname#1@def\endcsname{#5}
            \seq_gput_left:Nn \g_abbrs {#1}
            \prop_gput:Nnn \g_abbr_counts {#1} {1}
            \phantomsection\label{#1}
            \str_if_eq:nnTF{##1}{}{\emph{#5}}{##1}~(\hyperref[#1]{#4})
        }
    }
}

\ExplSyntaxOff
\makeatother

\abbr{ANN}{artificial neural network}{ANNs}{artificial neural networks}
\abbr{SGD}{stochastic gradient descent}{SGDs}{Stochastic Gradient Descents}
\abbr{PDE}{partial differential equation}{PDEs}{partial differential equations}
\abbr{ODE}{ordinary differential equation}{ODEs}{ordinary differential equations}
\abbr{FNO}{Fourier neural operator}{FNOs}{Fourier neural operators}
\abbr{ADANN}{Algorithmically Designed Artificial Neural Networks}{ADANNs}{Algorithmically Designed Artificial Neural Networks}
\abbr{LIRK}{linearly implicit Runge-Kutta}{LIRKs}{Linearly Implicit Runge-Kutta}
\abbr{GELU}{Gaussian Error Linear Unit}{GELUs}{Gaussian Error Linear Units}
\abbr{DeepONet}{deep operator network}{DeepONets}{deep operator networks}
\abbr{MLP}{full history recursive multilevel Picard}{MLPs}{full history recursive multilevel Picards}
\abbr{MC}{Monte Carlo}{MCs}{Monte Carlos}
\abbr{MLMC}{multilevel Monte Carlo}{MLMCs}{multilevel Monte Carlos}
\abbr{QMC}{quasi-Monte Carlo}{QMCs}{quasi-Monte Carlos}
\abbr{LRV}{learning the random variables}{LRVs}{learning the random variables}
\abbr{SDE}{stochastic differential equation}{SDEs}{stochastic differential equations}
\abbr{BSDE}{backward stochastic differential equation}{BSDEs}{backward stochastic differential equations}
\abbr{PDF}{probability density function}{PDFs}{probability density functions}
\abbr{ENLL}{expected negative log-likelihood}{ENLLs}{expected negative log-likelihoods}
\abbr{KL}{Kullback-Leibler}{KLs}{Kullback-Leibler}

\abbr{DDPM}{denoising diffusion probabilistic model}{DDPMs}{denoising diffusion probabilistic models}
\abbr{DDIM}{denoising diffusion implicit model}{DDIMs}{denoising diffusion implicit models}
\abbr{FID}{Fréchet inception distance}{FIDs}{Fréchet inception distances}
\abbr{IS}{inception score}{ISs}{inception scores}
\abbr{AdaGN}{adaptive group normalization}{AdaGNs}{adaptive group normalizations}
\abbr{SSIM}{structured similarity index metric}{SSIMs}{structured similarity index metrics}
\abbr{PSNR}{peak signal-to-noise ratio}{PSNRs}{peak signal-to-noise ratios}
\abbr{LPIPS}{Learned Perceptual Image Patch Similarity}{LPIPSs}{Learned Perceptual Image Patch Similarities}
\abbr{GAN}{generative adversarial network}{GANs}{generative adversarial networks}
\abbr{VAE}{variational autoencoder}{VAEs}{variational autoencoders}

\begin{document}

\title{An overview of diffusion models \\
for generative artificial intelligence}

\author{Davide Gallon$^{1}$, Arnulf Jentzen$^{2,3}$, and  Philippe von Wurstemberger$^{4,5}$
\bigskip
    \\
	\small{$^1$Applied Mathematics: Institute for Analysis}
	\\
    \small{and Numerics, University of M\"unster,}
    \\
	\small{Germany, e-mail: davide.gallon@uni-muenster.de} 
	\smallskip
	\\
	\small{$^2$School of Data Science and Shenzhen Research Institute of }
	\\
	\small{Big Data, The Chinese University of Hong Kong, Shenzhen}
    \\
    \small{(CUHK-Shenzhen), China, e-mail: ajentzen@cuhk.edu.cn} 
	\smallskip
	\\
	\small{$^3$Applied Mathematics: Institute for Analysis and Numerics,}
	\\
	\small{University of M\"unster, Germany, e-mail: ajentzen@uni-muenster.de} 
	\smallskip
	\\
    \small{$^4$Risklab, Department of Mathematics, ETH Zurich,}
	\\
	\small{Switzerland, e-mail: philippe.vonwurstemberger@math.ethz.ch} 
    \smallskip
	\\
	\small{$^5$School of Data Science, The Chinese University of}
	\\
	\small{Hong Kong, Shenzhen (CUHK-Shenzhen),} 
    \\
    \small{China, e-mail: philippevw@cuhk.edu.cn} 
	\smallskip
	\\
}

\date{\today}

\maketitle

\begin{abstract}
\noindent

This article provides a mathematically rigorous introduction to \emph{denoising diffusion probabilistic models} (DDPMs), sometimes also referred to as \emph{diffusion probabilistic models} or \emph{diffusion models}, for generative artificial intelligence.
We provide a detailed basic mathematical framework for DDPMs and explain the main ideas behind training and generation procedures. 
In this overview article we also review selected extensions and improvements of the basic framework from the literature such as improved DDPMs, denoising diffusion implicit models, classifier-free diffusion guidance models, and latent diffusion models.

\end{abstract}

\tableofcontents
\section{Introduction}

The goal of generative modelling is to generate new data samples from an unknown underlying distribution based on a dataset of samples from that distribution.
Many different machine learning approaches for this goal have been proposed, 
such as 
\GANs\ \cite{goodfellow2014generative}, 
\VAEs\ \cite{DBLP:journals/corr/KingmaW13},
autoregressive models \cite{oord2016conditionalimagegenerationpixelcnn}, %
normalizing flows \cite{rezende2016variationalinferencenormalizingflows}, %
and energy-based models \cite{lecun2006tutorial}. %
In this article, we provide an introduction to \DDPMs, a class of generative methods (sometimes also called \emph{diffusion models} or \emph{diffusion probabilistic models}) which is based on the idea to reconstruct a diffusion process, which starts at the underlying distribution and gradually adds noise to its state until it arrives at a terminal state that is purely noise, backwards.
Through this backward reconstruction, pure noise is transformed into meaningful data, and as such \DDPMs\ provide a natural generative framework.
We aim to provide a basic but rigorous understanding of the motivating ideas behind \DDPMs\ and precise descriptions of some of the most influential \DDPM-based methods in the literature.

\DDPMs\ were originally introduced in \cite{sohl2015deep} and further popularized in \cite{ho2020denoising} and have been able to achieve state of the art results in many domains like image synthesis and editing
\cite{9878449,pmlr-v139-ramesh21a,ramesh2022hierarchical,nichol2022glide,saharia2022photorealistic},  
video generation \cite{ho2022video,yang2022diffusion},
natural language processing \cite{austin2023structured,li2022diffusionlm},
and
anomaly detection \cite{Wyatt_2022_CVPR,wolleb2022diffusion}.
In the canonical formulation, a \DDPM\ is a framework consisting of two stochastic processes, a forward process and a backward process. 
The forward process -- the \emph{diffusion} process -- starts at the initial time step at the (approximate) underlying distribution (\eg its initial state could be a random sample from the dataset) and then gradually adds noise to its state so that its state at the terminal time step is (approximately) purely noise.
The backward process -- the \emph{denoising} process -- is a parametric process which starts (at the terminal time step) at a purely noisy state.
The idea in the context of \DDPMs\ is to learn parameters for this backward process such that the distribution at each time step of the backward process is approximately the same as the distribution at the corresponding time step of the forward process.
If this is achieved, the backward process can be interpreted to gradually remove noise from its initial state until it is at the initial distribution of the forward process.
In that sense, the backward process gradually \emph{denoises} its purely noisy initial state.
Once appropriate parameters for the backward process have been found, the generative procedure consists in sampling realizations of the backwards process.

We rigorously set up a general mathematical framework for \DDPMs\ and explain the ideas behind the training of the backward process and the creation of generative samples in \cref{sec:ddpm}.
We then consider the most common special case of this framework when the noise is Gaussian and the backward process is governed by a denoising \ANN\ in \cref{sec:DDPM_gaussian}.
In \cref{sec:evaluation} we thereafter discuss some metrics from the literature on how to evaluate the quality of generated samples.
We conclude in \cref{sec:advanced} with a discussion of some of the most popular \DDPM-based methods that have been proposed in the literature such as Improved \DDPMs\ (see \cite{ho2020denoising}), 
\DDIMs\ (see \cite{song2022denoising}), 
classifier-free diffusion guidance models 
(see \cite{ho2022classifierfree}), and latent diffusion models (see \cite{9878449}).
In particular, classifier-free diffusion guidance models and latent diffusion models show how to guide the backward  process to generate data from different classes and based on a given text, respectively. 
Code supporting this article is available at \url{https://github.com/deeplearningmethods/diffusion_model}.

\section{Denoising diffusion probabilistic models (DDPMs)}
\label{sec:ddpm}

In this section we introduce the main ideas behind \DDPMs. 
Specifically, we introduce and discuss a general mathematical framework for \DDPMs\ and elaborate some of its elementary properties in \cref{sec:ddpm_problemformulation},
we discuss the training objective with which \DDPMs\ aim to achieve the goal of generative modelling in \cref{sec:training_obj}, and
we present a simplified \DDPM\ methodology based on this training objective in \cref{sec:ddpm_simple}.

\subsection{General framework for DDPMs}\label{sec:ddpm_problemformulation}

\begin{setting}[General framework for \DDPMs]\label{setting0}
Let $\di , \fd, T \in \N$, let  $(\Omega, \cF, \mathbb{P})$ be a probability space, 
for every $\theta \in (\R^{\fd}\cup \{ \emptyset \})$ let  $\Xt{\theta} = (\Xt{\theta}_t)_{t \in \{0,1,\ldots,T\}} \colon \{0,1,\ldots,T\} \times \Omega \to \R^{ \di }$ 
be a stochastic process, assume that $(\Xt{\theta})_{\theta \in \R^{\fd}}$ and $\Xt{\emptyset}$ are independent,
for every $\theta \in (\R^{\fd}\cup \{ \emptyset \})$ let $\pjoint{\theta} \colon (\R^{\di})^{T+1}\to (0,\infty)$ be a measurable function which satisfies\footnote{Note that for every tolopogical space $(E, \cE)$ it holds that $\cB(E)$ is the Borel $\sigma$-algebra of $E$ (the smallest $\sigma$-algebra that contains $\cE$).} for all $B_0, B_1, \ldots, B_T \in \cB(\R^{\di})$ that
\begin{equation}
    \mathbb{P}\bigl(\Xt{\theta}_0 \in B_0,\Xt{\theta}_1 \in B_1, \ldots, \Xt{\theta}_T \in B_T\bigr)= \int_{B_0}\int_{B_1}\ldots \int_{B_T}\pjoint{\theta}(x_0,x_1,\ldots,x_T) \, \d x_0 \, \d x_1 \ldots \, \d x_T,
\end{equation}
for every $\theta \in ( \R^{\fd}\cup \{ \emptyset \})$, $S \in \{1, \ldots, T \}$, $a_1,\ldots,a_{T+1} \in \N_0$ with $\{a_1,\ldots,a_{T+1}\}=\{0,1,\ldots,T\}$
let $\p{\theta}{a_1,\ldots,a_S} \colon (\R^{\di})^S\to (0,\infty)$ satisfy for all $x_{a_1}, \dots,x_{a_S} \in \R^{\di}$
that
\begin{equation}
\begin{split}
    &\p{\theta}{a_1,\ldots,a_S}(x_{a_1},\ldots,x_{a_S}) \\
    & = 
    \begin{cases}
          \displaystyle \int_{\R^{\di}} \int_{\R^{\di}} \ldots \int_{\R^{\di}} \pjoint{\theta}(x_0,x_1,\ldots,x_T) \, \d x_{a_{S+1}} \, \d x_{a_{S+2}}  \ldots \, \d x_{a_{T+1}}   & \colon S \leq T \\
          \displaystyle \pjoint{\theta}(x_0,x_1,\ldots,x_T) & \colon S=T+1,
    \end{cases} 
\end{split}
\end{equation}
 for every $\theta \in (\R^{\fd}\cup \{ \emptyset \})$, $S,K \in  \{1, \ldots, T \}$, $a_1,\ldots,a_{S+K} \in \{0,1,\ldots,T\}$ with $\vert\{ a_1,\ldots,a_{S+K} \}  \vert=S + K$ let $\pcond{\theta}{a_1,\ldots,a_S}{a_{S+1},\ldots, a_{S+K}} = (\pcond{\theta}{a_1,\ldots,a_S}{a_{S+1},\ldots, a_{S+K}}(\textbf{x}\vert \textbf{y}))_{(\textbf{x},\textbf{y})\in (\R^{\di})^{S} \times(\R^{\di})^{K}} \colon \allowbreak (\R^{\di})^{S} \times(\R^{\di})^{K} \to (0,\infty)$ satisfy for all %
$x_{a_1},\ldots,x_{a_{S+K}} \in \R^{\di}$
that
\begin{equation}\label{setting0:pcond}
        \pcond{\theta}{a_1,\ldots,a_S}{ a_{S+1},\ldots,a_{S+K}}\argpcond{(x_{a_1},\ldots,x_{a_S}}{x_{a_{S+1}},\ldots,x_{a_{S+K}}) } = \frac{\p{\theta}{a_1,\ldots,a_{S+K}}(x_{a_1},\ldots,x_{a_{S+K}})}{\p{\theta}{a_{S+1},\ldots,a_{S+K}}(x_{a_{S+1}},\ldots,x_{a_{S+K}})},
\end{equation}
let $\Pi \colon \R^\di \to (0,\infty)$ be a function, and assume for all $\theta \in \R^\fd$ that $\p{\theta}{T}=\Pi$.
\end{setting}

\begin{remark}[Explanations for \cref{setting0}]
    \label{remark:setting0}
In this remark we provide some intuitive interpretations for the mathematical objects appearing in \cref{setting0} and roughly explain their role in the context of \DDPMs\ for generative modelling.
Roughly speaking, we note that
\begin{enumerate}[label=(\roman*)]
    \item
    we think of 
    $\di$
    as the dimension of the objects we want to generate (\eg the number of pixels in an image),
    \item
    we think of $T$ as the numbers of time steps in the \DDPM,
    \item
    we think of $\Xt{\emptyset} = (\Xt{\emptyset}_t)_{t \in \{0,1,\ldots,T\}}$ as the forward process in the \DDPM\ which gradually adds noise to an initial state $\Xt{\emptyset}_0$,
    \item
    we think of the initial state $\Xt{\emptyset}_0$ of the forward process as a random variable with the (approximate) distribution from which we would like to generate samples (\eg the initial state could correspond to a random image from a training dataset),
    \item
    we think of $\fd$ as the number of trainable parameters in the \DDPM,
    \item
    we think of $(\Xt{\theta})_{\theta \in \R^\fd} = ((\Xt{\theta}_t)_{t \in \{0,1,\ldots,T\}})_{\theta \in \R^\fd}$ as the parametric backward process in the \DDPM\ parametrized by parameters $\theta \in \R^\fd$ which aims to gradually remove noise from its initial state $\Xt{\theta}_T$, and
    \item
    we think of the \PDF\ $\Pi$ of the initial state $(\Xt{\theta}_T)_{\theta \in \R^\fd}$ of the backward process as a \PDF\ of a noisy distribution (\eg a multivariate Gaussian distribution).
\end{enumerate}
In addition to the objects described above, we also introduce notations for the joint, marginal, and conditional \PDFs\ of the forward and backward processes.
Specifically, note for every 
    $\theta \in (\R^{\fd}\cup \{ \emptyset \})$,
    $a_1, \ldots, a_{T+1} \in \{0,1,\ldots,T\}$,
    $S, K \in \{1, \ldots, T\}$
with 
    $\{a_1, \ldots, a_{T+1}\} = \{0,1,\ldots,T\}$ 
and 
    $S + K \leq T$
that
\begin{enumerate}[label=(\roman*)]  
    \item 
    we think of 
    $
        \pjoint{\theta}
    $
    as the joint \PDF\ of the process $\Xt{\theta}$,
    
    \item 
    we think of
    $
        \p{\theta}{a_1,\ldots,a_S}
    $
    as the marginal \PDF\ of the process $\Xt{\theta}$ for the time steps $a_1, \ldots, a_S$, and

    \item
    we think of
    $
        \pcond{\theta}{a_1,\ldots,a_S}{a_{S+1},\ldots,a_{S+K}}
    $
    as the conditional \PDF\ of the process $\Xt{\theta}$ for the time steps $a_1, \ldots, a_S$ given the time steps $a_{S+1}, \ldots, a_{S+K}$.
\end{enumerate}
Loosely speaking, in the context of \DDPMs\ the goal in \cref{setting0} is to find parameters $\opttheta \in \R^\fd$ such that the terminal value $\Xt{\opttheta}_0$ of the backward process is approximately distributed like the initial state $\Xt{\emptyset}_0$ of the forward process, or, in other terms,
\begin{equation}
\label{remark:setting0:eq1}
\begin{split}
    \p{\opttheta}{0}
\approx
    \p{\emptyset}{0}.
\end{split}
\end{equation}
The idea of \DDPMs\ is to achieve this goal by training the parameter $\theta \in \R^\fd$ such that the backward process $\Xt{\theta}$
is approximately distributed like the forward process $\Xt{\emptyset}$.
For this, we think that the distribution of the terminal state $\Xt{\emptyset}_T$ of the forward process roughly has the same distribution as the initial state $(\Xt{\theta}_T)_{\theta \in \R^\fd}$ of the backward process, that is,
\begin{equation}
\label{remark:setting0:eq2}
\begin{split}
    \p{\emptyset}{T}
\approx
    \Pi.
\end{split}
\end{equation}
A practical interpretation of this assumption is that the forward process $\Xt{\emptyset}$ 
adds noise to its initial state $\Xt{\emptyset}_0$ until it reaches a completely noisy state $\Xt{\emptyset}_T$
(cf., \eg \cref{remark:prodq} for a discussion of this assumption in the context of \DDPMs\ with Gaussian noise).
\end{remark}

In many applications, the forward process $\Xt{\emptyset}$ and the backward process $(\Xt{\theta})_{\theta \in \R^\fd}$ in \cref{setting0} are constructed to be Markov processes. 
We add this assumption to \cref{setting0} in the following framework.

\begin{setting}[General framework for \DDPMs\ with Markov assumptions]\label{setting01}
    Assume \cref{setting0} and assume for all $\theta \in \R^{\fd}$, $t \in \{1,\ldots,T\}$, $x_0,x_1,\ldots,x_T \in \R^{\di}$ that
\begin{equation}\label{setting0:prob}
        \pcond{\emptyset}{t}{t-1,t-2,\ldots,0}
        \argpcond{(x_t}{x_{t-1},x_{t-2},\ldots, x_{0}) }
        = \pcond{\emptyset}{t}{t-1}\argpcond{(x_t}{x_{t-1})}  
\end{equation}
\begin{equation}\label{setting0:prob2}
        \andq   \pcond{\theta}{t-1}{t,t+1,\ldots,T}
        \argpcond{(x_{t-1}}{x_{t},x_{t+1},\ldots,x_{T}) } = \pcond{\theta}{t-1}{t}
        \argpcond{(x_{t-1}}{x_{t})}. 
\end{equation}
\end{setting}

\begin{figure}[H]
\centering
\begin{tikzpicture}[start chain=going left,node distance=3.6cm, every state/.style={minimum size=3em}]
\node[state, on chain] (N+1) {$X_{T}^\emptyset$};
\node[on chain] (g) {$\cdots$};
\node[state, on chain] (1) {$X_1^\emptyset$};
\node[state, on chain] (0) {$X_0^\emptyset$};
\draw[
    >=latex,
    auto=right,
    loop above/.style={out=-75,in=-105,loop},
    every loop,
]
(0) edge node {$\pcond{\emptyset}{1}{0}\argpcond{(X_1^\emptyset}{X_0^\emptyset)}$} (1)
(1) edge node {$\pcond{\emptyset}{2}{1}\argpcond{(X_2^\emptyset}{X_1^\emptyset)}$} (g)
(g) edge node {$ \pcond{\emptyset}{T}{T-1}\argpcond{(X_T^\emptyset}{X_{T-1}^\emptyset)}$} (N+1);
\end{tikzpicture}
\centering
\begin{tikzpicture}[start chain=going left,node distance=3.6cm, every state/.style={minimum size=3em}]
\node[state, on chain] (N+1) {$X_{T}^\theta$};
\node[state, on chain] (N) {$X_{T-1}^\theta$};
\node[on chain] (g) {$\cdots$};
\node[state, on chain] (0) {$X_0^\theta$};
\draw[
    >=latex,
    auto=right,
    loop below/.style={out=75,in=105,loop},
    every loop,
]
(N+1) edge node {$\pcond{\theta}{T-1}{T}\argpcond{(X_{T-1}^\theta}{X_{T}^\theta)}$} (N)
(N) edge[] node {$\pcond{\theta}{T-2}{T-1}\argpcond{(X_{T-2}^\theta}{X_{T-1}^\theta)}$} (g)
(g) edge node {$\pcond{\theta}{0}{1}\argpcond{(X_0^\theta}{X_{1}^\theta)}$} (0);
\end{tikzpicture}
  \caption{Graphical illustration the forward process $\Xt{\emptyset}$ and the backward process $(\Xt{\theta})_{\theta \in \R^\fd}$ in \DDPMs\ with Markov assumptions in \cref{setting01}.
  }
  \label{fig:dd}
\end{figure}

\begin{remark}[Transition kernels and transition densities in \cref{setting01}]
    \label{remark:transition}
    Assume \cref{setting01}.
    Roughly speaking, the assumptions in \cref{setting0:prob,setting0:prob2} imply that 
    for both the forward and backward processes, the distribution of the process at any step, conditioned on all previous steps of the respective process,
    only depends on the distribution of the immediately preceding step.
    In other words, the forward process $\Xt{\emptyset}$ is a Markov process and the backward process $(\Xt{\theta})_{\theta \in \R^\fd}$ is a backward Markov process.
    Specifically, we have for all 
        $\theta \in \R^{\fd}$, 
        $t \in \{1,2,\ldots,T\}$,
        $B \in \cB(\R^{\di})$
    that
    \begin{equation}
    \label{T_B_D}
    \begin{split}
        \P(\Xt{\emptyset}_t \in B \mid \Xt{\emptyset}_{t-1}, \Xt{\emptyset}_{t-2}, \ldots, \Xt{\emptyset}_0)
    =
        \P(\Xt{\emptyset}_t \in B \mid \Xt{\emptyset}_{t-1})
    \end{split}
    \end{equation}
    \begin{equation}
    \label{T_B_D}
    \begin{split}
        \andq
        \P(\Xt{\theta}_{t-1} \in B \mid \Xt{\theta}_{t}, \Xt{\theta}_{t+1}, \ldots, \Xt{\theta}_T)
    =
        \P(\Xt{\theta}_{t-1} \in B \mid \Xt{\theta}_{t}).
    \end{split}
    \end{equation}
    In this Markovian context
    we refer 
    to the functions
    \begin{equation}
    \label{T_B_D}
    \begin{split}
            \R^{\di} \times \cB(\R^{\di}) \ni (x_{t-1},B) 
        &\mapsto 
            \int_B \pcond{\emptyset}{t}{t-1}\argpcond{(x_t}{x_{t-1})} \, \d x_t 
        \in  
            [0,1]
    \end{split}        
    \end{equation}
    for $t \in \{1, 2, \ldots, T\}$
    as the \emph{transition kernels} for the forward process,
    we refer to the functions
    \begin{equation}
    \label{T_B_D}
    \begin{split}
            \R^{\di} \times \cB(\R^{\di})  \ni (x_{t},B) 
        &\mapsto 
            \int_B \pcond{\theta}{t-1}{t}\argpcond{(x_{t-1}}{x_{t})} \, \d x_{t-1} 
        \in 
            [0,1]            
    \end{split}
    \end{equation}
    for $t \in \{1, 2, \ldots, T\}$, $\theta \in \R^{\fd}$
    as the \emph{transition kernels} for the backward process,
    we refer to the functions
    \begin{equation}
    \label{T_B_D}
    \begin{split}
            \R^{\di} \times \R^{\di} \ni (x_{t-1},x_{t})
        &\mapsto
            \pcond{\emptyset}{t}{t-1}\argpcond{(x_{t}}{x_{t-1})} 
        \in 
            [0,\infty) 
    \end{split}
    \end{equation}
    for $t \in \{1, 2, \ldots, T\}$
    as the transition densities for the forward process,
    and
    we refer to the functions
    \begin{equation}
    \label{T_B_D}
    \begin{split}
            \R^{\di} \times \R^{\di}
        \ni  
            (x_{t},x_{t-1})
        &\mapsto
            \pcond{\theta}{t-1}{t}\argpcond{(x_{t-1}}{x_{t})}
        \in  
            [0,\infty) 
    \end{split}
    \end{equation}
    for $t \in \{1, 2, \ldots, T\}$, $\theta \in \R^{\fd}$
    as the transition densities for the backward process.
    An illustration of the forward process $\Xt{\emptyset}$, the backward process $(\Xt{\theta})_{\theta \in \R^\fd}$, and the role of the respective transition densities is provided in \cref{fig:dd}.
\end{remark}

Under the Markov assumptions of \cref{setting01}, the marginal \PDFs\ of the forward and backward processes admit a representation in terms of the respective transition densities. This is the subject of the next lemma.
\begin{lemma}[Representation for marginal \PDFs\ in \DDPMs\ with Markov assumptions]\label{cor:multprod}
Assume \cref{setting01}.
Then it holds for all $\theta \in \R^{\fd}$, $t \in \{1,\ldots,T\}$, $x_0,x_1,\ldots,x_T \in \R^{\di}$ that
\begin{equation}\label{cor:multprod:thesis}
        \p{\emptyset}{0,1,\ldots,t}(x_0, x_1,\ldots,x_t)= \p{\emptyset}{0}(x_0) \left[ \textstyle \prod_{s=1}^{t} \pcond{\emptyset}{s}{s-1} \argpcond{(x_{s}}{ x_{s-1} )} \right] 
\end{equation}
\begin{equation}
         \andq \p{\theta}{t-1,t,\ldots,T}(x_{t-1},x_{t},\ldots,x_T) = \p{\theta}{T}(x_T) \left[ \textstyle \prod_{s=t}^{T} \pcond{\theta}{s-1}{s} \argpcond{(x_{s-1}}{ x_{s} )} \right].
\end{equation}
\end{lemma}
\begin{cproof}{cor:multprod}
    \Nobs that \cref{setting0:pcond} implies that for all $\theta \in \R^{\fd}$, $t \in \{1,\ldots,T\}$, $x_0,x_1,\ldots,x_T \in \R^{\di}$ it holds that
    \begin{equation}
\begin{split}
        \p{\emptyset}{0,1,\ldots,t}(x_0, x_1,\ldots,x_t)&= \p{\emptyset}{0}(x_0) \left[\textstyle\prod_{s=1}^{t} \pcond{\emptyset}{s}{s-1,s-2,\ldots,0} \argpcond{(x_{s}}{x_{s-1},x_{s-2}, \ldots, x_{0} )}\right]   \\    
        \andq \p{\theta}{t-1,t,\ldots,T}(x_{t-1},x_{t},\ldots,x_T) &= \p{\theta}{T}(x_T) \left[ \textstyle \prod_{s=t}^{T} \pcond{\theta}{s-1}{s,s+1,\ldots,T} \argpcond{(x_{s-1}}{x_{s}, x_{s+1},\ldots, x_{T} )} \right].\\
\end{split}
\end{equation}
This and the fact that for all $\theta \in \R^{\fd}$, $t \in \{1,\ldots,T\}$, $x_0,x_1,\ldots,x_T \in \R^{\di}$ it holds that
\begin{equation}
\begin{split}
        \pcond{\emptyset}{t}{t-1,t-2,\ldots,0}\argpcond{(x_t}{x_{t-1},x_{t-2},\ldots, x_{0})} & = \pcond{\emptyset}{t}{t-1}\argpcond{(x_t}{x_{t-1})}  \\
        \andq \pcond{\theta}{t-1}{t,t+1,\ldots,T}\argpcond{(x_{t-1}}{x_{t},x_{t+1},\ldots,x_{T})} & = \pcond{\theta}{t-1}{t}\argpcond{(x_{t-1}}{x_{t})} 
\end{split}
\end{equation}
demonstrate that for all $\theta \in \R^{\fd}$, $t \in \{1,\ldots,T\}$, $x_0,x_1,\ldots,x_T \in \R^{\di}$ it holds that
\begin{equation}
\begin{split}
        \p{\emptyset}{0,1,\ldots,t}(x_0, x_1,\ldots,x_t)&= \p{\emptyset}{0}(x_0) \left[ \textstyle \prod_{s=1}^{t} \pcond{\emptyset}{s}{s-1} \argpcond{(x_{s}}{ x_{s-1} )} \right]\\
        \andq \p{\theta}{t-1,t,\ldots,T}(x_{t-1},x_{t},\ldots,x_T) &= \p{\theta}{T}(x_T) \left[ \textstyle \prod_{s=t}^{T} \pcond{\theta}{s-1}{s} \argpcond{(x_{s-1}}{ x_{s} )} \right].\\
\end{split}
\end{equation}
\end{cproof}

\subsection{Training objective in DDPMs}
\label{sec:training_obj}

In this section we discuss the objective used to train the parameters of the backward process in \cref{setting0}. As discussed in \cref{remark:setting0}, the goal in the context of \DDPMs\ is to find parameters for the backward process
such that the terminal value of the backward process is approximately distributed like the initial value of the forward process (cf.\ \cref{remark:setting0:eq1} in \cref{remark:setting0}).
To achieve this, \cite{sohl2015deep} propose to minimize the \ENLL\ 
(sometimes called cross-entropy in the context of information theory)
of the \PDF\ of the initial value of the forward process
with respect to the \PDF\ of the terminal value of the backward process 
(see \cite[Section 5.5]{Goodfellow-et-al-2016} for an introduction to minimizing the \ENLL\ in the context of machine learning).
Roughly speaking, this \ENLL\ measures how similar the distribution of the terminal value of the backward process is to the distribution of the initial value of the forward process.

We start this section by introducing the concept of the \ENLL\ in \cref{def:exp_neg_log} and the related concept of the \KL\ divergence (see \cite{kullback1951information}) in \cref{def:KL_divergence}.
We then justify the choice of the \ENLL\ as a training objective in \cref{lemma:loglike}.
Thereafter, in \cref{lemma:upperboundE,remark:loglike} we discuss an upper bound for the \ENLL\ in the context of \cref{setting01} which can be used as an alternative training objective for the parameters of the backward process.

\begin{definition}[\ENLL]\label{def:exp_neg_log}
    Let $\di \in \N$ and
    for every $i \in \{1,2\}$
    let $p_i \colon \R^\di \to (0,\infty)$ be a measurable function which satisfies $\int_{\R^\di} p_i(x)\, \d x =1$. 
    Then we denote by $\negloglike{p_1}{p_2} \in \R \cup \{\infty\}$ the number given by
    \begin{equation}
        \negloglike{p_1}{p_2} 
    = 
        \int_{\R^\di}  -\ln\!\left({p_2(x)}\right) p_1(x) \, \d x
    \end{equation}
    and we call $\negloglike{p_1}{p_2}$ the \ENLL\ of $p_2$ with respect to $p_1$
    (we call $\negloglike{p_1}{p_2}$ the cross-entropy from $p_1$ to $p_2$).
\end{definition}

\begin{definition}[\KL\ divergence]\label{def:KL_divergence}
    Let $\di \in \N$ and
    for every $i \in \{1,2\}$
    let $p_i \colon \R^\di \to (0,\infty)$ be a measurable function which satisfies $\int_{\R^\di} p_i(x)\, \d x =1$. 
    Then we denote by $\infdiv{p_1}{p_2} \in \R \cup \{-\infty, \infty\}$ the extended real number given by 
    \begin{equation}
    \infdiv{p_1}{p_2} = \int_{\R^\di}  \ln\!\left(\frac{p_1(x)}{p_2(x)}\right)p_1(x) \, \d x
\end{equation}
    and we call $\infdiv{p_1}{p_2}$ the \KL\ divergence of $p_1$ from $p_2$.
\end{definition}

\cfclear
\begin{lemma}[Properties of the \ENLL\ and the \KL\ divergence]\label{lemma:loglike}
    Let $\di \in \N$, 
    for every $i \in \{1, 2\}$
    let $p_i \colon \R^\di \to (0,\infty)$ be a measurable function which satisfies $\int_{\R^\di} p_i(x)\, \d x =1$, 
    let $(\Omega, \cF, \mathbb{P})$ be a probability space, and
    let $X \colon \Omega \to \R^\di$ satisfy for all $B \in \cB(\R^{\di})$ that $\mathbb{P}(X \in B) = \int_{B} p_1(x)\, \d x $.
    Then 
    \begin{enumerate}[label=(\roman*)]
    \item \label{lemma:loglike:item1} 
    it holds that $\negloglike{p_1}{p_2} = \Eb{}{-\ln\!\left(p_2(X)\right)}$,
    \item \label{lemma:loglike:item2}
    it holds that $\infdiv{p_1}{p_2} = \Eb{}{\ln\!\left(\frac{p_1(X)}{p_2(X)}\right)}$,
    \item \label{lemma:loglike:item3} 
    it holds that 
    $\negloglike{p_1}{p_2} - \negloglike{p_1}{p_1} = \infdiv{p_1}{p_2}\geq0$, 
    and 
    \item \label{lemma:loglike:item4} it holds that the following three statements are equivalent: 
        \begin{enumerate}[label=(iv.\Roman*)]
            \item \label{lemma:loglike:item2-1} It holds that $\infdiv{p_1}{p_2}=0$.
            \item \label{lemma:loglike:item2-2} It holds that $\negloglike{p_1}{p_2} =\negloglike{p_1}{p_1}$.
            \item \label{lemma:loglike:item2-3} It holds Lebesgue-almost everywhere that $p_1=p_2$
        \end{enumerate}
    \end{enumerate}
     \cfout.
\end{lemma}
\begin{cproof}{lemma:loglike}
    \Nobs that the fact that for all $B \in \cB(\R^{\di})$ it holds that $\mathbb{P}(X \in B) = \int_{B} p_1(x)\, \d x $ shows that 
    \begin{equation}\label{lemma:loglike:proof:1}
        \negloglike{p_1}{p_2} 
        = 
        \int_{\R^\di}  -\ln\!\left({p_2(x)}\right) p_1(x) \, \d x 
        = 
        \Eb{}{-\ln\!\left(p_2(X)\right)}
    \end{equation}
    \begin{equation}
        \andq  \infdiv{p_1}{p_2} = \int_{\R^\di}  \ln\!\left(\frac{p_1(x)}{p_2(x)}\right)p_1(x) \, \d x
        =\Eb{}{\ln\!\left(\frac{p_1(X)}{p_2(X)}\right)}
    \end{equation}
    \cfload.
    This and \cref{lemma:loglike:proof:1} prove \cref{lemma:loglike:item1,lemma:loglike:item2}.
    \Nobs that 
    \begin{equation}
    \begin{split}
        &\infdiv{p_1}{p_2} = \int_{\R^\di}  \ln\!\left(\frac{p_1(x)}{p_2(x)}\right)p_1(x) \, \d x = \int_{\R^\di}  \big(\ln(p_1(x))- \ln(p_2(x))\big)p_1(x) \, \d x \\
        & = \negloglike{p_1}{p_2} - \negloglike{p_1}{p_1}.
    \end{split}
    \end{equation}
    This and, \eg \cite[Section 8.2.1]{barber2012bayesian} imply \cref{lemma:loglike:item3}.
     Moreover, \nobs that \ref{lemma:loglike:item3} ensures that $(\ref{lemma:loglike:item2-1}\leftrightarrow\ref{lemma:loglike:item2-2})$.
     In addition, \nobs that, \eg \cite[(8.2.36)]{barber2012bayesian} demonstrates that $(\ref{lemma:loglike:item2-1}\leftrightarrow\ref{lemma:loglike:item2-3})$.
\end{cproof}

\cfclear
\begin{lemma}[Upper bounds for \ENLL\ objective in \DDPMs] \label{lemma:upperboundE}
    Assume \cref{setting01}.
    Then it holds for all $\theta \in \R^{\fd}$ that
    \begin{equation}\label{lemma:upperboundE:thesis}
    \begin{split}
         &\negloglike[\big]{\p{\emptyset}{0}}{\p{\theta}{0}} 
         =
         \Eb{}{- \ln \big(\p{\theta}{0}(X_{0}^\emptyset)\big) } \\
         &\leq
        \Eb{} {\infdiv{\pcond{\emptyset}{T}{0}\argpcond{(\cdot}{X_0^\emptyset)}}{\Pi}} -\Eb{}{\ln \big( \pcond{\theta}{0}{1}\argpcond{(X_0^\emptyset}{X_1^\emptyset)}\big)}\\
        &\quad + \sum_{t = 2}^T\Eb{}{\infdiv{\pcond{\emptyset}{t-1}{t,0}\argpcond{(\cdot}{X_t^\emptyset,X_0^\emptyset)}}{\pcond{\theta}{t-1}{t}\argpcond{(\cdot}{X_{t}^\emptyset)}}}
    \end{split}
    \end{equation}
    \cfout.
\end{lemma}
\begin{cproof}{lemma:upperboundE}

\Nobs that Jensen’s inequality imply that for all $\theta \in \R^{\fd}$, $x_0 \in \R^{\di}$ it holds that
\begin{equation}
\begin{split}
     &\ln \big(\p{\theta}{0}(x_{0})\big) =\ln \bigg( \int_{\R^{\di}} \ldots \int_{\R^{\di}} \pjoint{\theta}(x_0,x_1,\ldots,x_T) \, \d x_{1}  \ldots \, \d x_{T} \bigg)\\
     &=  \ln \bigg(  \int_{\R^{\di}} \ldots \int_{\R^{\di}} \pcond{\emptyset}{1,\ldots,T}{0}\argpcond{(x_1,\ldots, x_T}{ x_{0})} \frac{\pjoint{\theta}(x_0,x_1,\ldots,x_T) }{\pcond{\emptyset}{1,\ldots,T}{0}\argpcond{(x_1,\ldots, x_T}{ x_{0})} }\, \d x_{1}  \ldots \, \d x_{T} \bigg) \\
     & \geq   \int_{\R^{\di}} \ldots \int_{\R^{\di}} \pcond{\emptyset}{1,\ldots,T}{0}\argpcond{(x_1,\ldots, x_T}{ x_{0})} \ln \bigg( \frac{\pjoint{\theta}(x_0,x_1,\ldots,x_T) }{\pcond{\emptyset}{1,\ldots,T}{0}\argpcond{(x_1,\ldots, x_T}{ x_{0})} }\bigg) \, \d x_{1}  \ldots \, \d x_{T}.  \\
\end{split} 
\end{equation}
This assures that for all $\theta \in \R^{\fd}$ it holds that
\begin{equation}
\begin{split}
     &\Eb{}{\ln \big(\p{\theta}{0}(X_{0}^\emptyset)\big)  } = 
      \int_{\R^{\di}} \p{\emptyset}{0}(x_{0}) \ln \big(\p{\theta}{0}(x_{0})\big) \, \d x_{0}\\
     & \geq \int_{\R^{\di}} \int_{\R^{\di}} \ldots \int_{\R^{\di}} \p{\emptyset}{0}(x_{0})  \pcond{\emptyset}{1,\ldots,T}{0}\argpcond{(x_1,\ldots, x_T}{ x_{0})} \\
     & \quad \ln \bigg(\frac{\pjoint{\theta}(x_0,x_1,\ldots,x_T) }{\pcond{\emptyset}{1,\ldots,T}{0}\argpcond{(x_1,\ldots, x_T}{ x_{0})} }\bigg) \, \d x_{0} \, \d x_{1}  \ldots \, \d x_{T} \\
     & = \Eb{}{\ln \bigg(\frac{\pjoint{\theta}(X_0^\emptyset,X_1^\emptyset,\ldots,X_T^\emptyset)}{\pcond{\emptyset}{1,\ldots,T}{0}\argpcond{(X_1^\emptyset,\ldots, X_T^\emptyset}{ X_{0}^\emptyset)} }\bigg) }.
\end{split} 
\end{equation}
This and \cref{cor:multprod} demonstrate that for all $\theta \in \R^{\fd}$ it holds that
\begin{equation}\label{lemma:upperboundE:first:res}
    \begin{split}
        & \Eb{}{\ln (\p{\theta}{0}(X_{0})) } \geq  \Eb{}{\ln \bigg(\frac{\pjoint{\theta}(X_0^\emptyset,X_1^\emptyset,\ldots,X_T^\emptyset)}{\pcond{\emptyset}{1,\ldots,T}{0}\argpcond{(X_1^\emptyset,\ldots, X_T^\emptyset}{ X_{0}^\emptyset)} }\bigg) } \\
        & =  \Eb{}{\ln \bigg( \frac{\p{\theta}{T}(X_{T}^\emptyset) \prod_{t=1}^{T} \pcond{\theta}{t-1}{t}\argpcond{(X_{t-1}^\emptyset}{X_{t}^\emptyset)}}{\prod_{t=1}^{T}\pcond{\emptyset}{t}{t-1}\argpcond{(X_{t}^\emptyset}{X_{t-1}^\emptyset)}} \bigg)}\\
        &=  \Eb{}{\ln \bigg(  \frac{\p{\theta}{T}(X_{T}^\emptyset)
        \pcond{\theta}{0}{1}\argpcond{(X_{0}^\emptyset}{X_{1}^\emptyset)}}{\pcond{\emptyset}{1}{0}\argpcond{(X_{1}^\emptyset}{X_{0}^\emptyset)}}
        \prod_{t=2}^{T}\frac{
        \pcond{\theta}{t-1}{t}\argpcond{(X_{t-1}^\emptyset}{X_{t}^\emptyset)}}{ \pcond{\emptyset}{t}{t-1}\argpcond{(X_{t}^\emptyset}{X_{t-1}^\emptyset)}}\bigg)}\\
          &= \Eb{}{ \ln \big(\p{\theta}{T}(X_{T}^\emptyset)\big)  
          + \ln\big(\pcond{\theta}{0}{1}\argpcond{(X_0^\emptyset}{X_1^\emptyset)}\big)
          - \ln\big(\pcond{\emptyset}{1}{0}\argpcond{(X_1^\emptyset}{X_0^\emptyset)} \big)       
          + \sum_{t = 2}^T \ln \frac{
        \pcond{\theta}{t-1}{t}\argpcond{(X_{t-1}^\emptyset}{X_{t}^\emptyset)}}{ \pcond{\emptyset}{t}{t-1}\argpcond{(X_{t}^\emptyset}{X_{t-1}^\emptyset)}} }. \\
        \end{split}
        \end{equation}
        This and the fact that for all $t \in \{2,3,\ldots,T\}$, $x_0,x_{t-1},x_t \in \R^{\di}$ it holds that
 \begin{equation}
         \pcond{\emptyset}{t}{t-1}\argpcond{(x_{t}}{x_{t-1})}=
        \pcond{\emptyset}{t-1}{t,0}\argpcond{(x_{t-1}}{x_t,x_0)}
       \frac{\pcond{\emptyset}{t}{0}\argpcond{(x_{t}}{x_0)}}{\pcond{\emptyset}{t-1}{0}\argpcond{(x_{t-1}}{x_0)}}
    \end{equation}
    show that for all $\theta \in \R^{\fd}$ it holds that
        \begin{equation}
        \begin{split}
          & \Eb{}{\ln (\p{\theta}{0}(X_{0})) }\\
          & \geq \Eb{}{ \ln \big(\p{\theta}{T}(X_{T}^\emptyset)\big)  
          + \ln\big(\pcond{\theta}{0}{1}\argpcond{(X_0^\emptyset}{X_1^\emptyset)}\big)
          - \ln\big(\pcond{\emptyset}{1}{0}\argpcond{(X_1^\emptyset}{X_0^\emptyset)} \big)       
          + \sum_{t = 2}^T \ln \frac{
        \pcond{\theta}{t-1}{t}\argpcond{(X_{t-1}^\emptyset}{X_{t}^\emptyset)}}{ \pcond{\emptyset}{t}{t-1}\argpcond{(X_{t}^\emptyset}{X_{t-1}^\emptyset)}} } \\
          &=  \E \Bigg[ \ln \big(\p{\theta}{T}(X_{T}^\emptyset)\big)  
          + \ln\big(\pcond{\theta}{0}{1}\argpcond{(X_0^\emptyset}{X_1^\emptyset)}\big)
          - \ln\big(\pcond{\emptyset}{1}{0}\argpcond{(X_1^\emptyset}{X_0^\emptyset)} \big) \\
          & \quad \left. + \sum_{t = 2}^T \ln \bigg(\frac{\pcond{\theta}{t-1}{t}\argpcond{(X_{t-1}^\emptyset}{X_{t}^\emptyset)} \pcond{\emptyset}{t-1}{0}\argpcond{(X_{t-1}^\emptyset}{X_0^\emptyset)}}{\pcond{\emptyset}{t-1}{t,0}\argpcond{(X_{t-1}^\emptyset}{X_t^\emptyset,X_0^\emptyset)} \pcond{\emptyset}{t}{0}\argpcond{(X_t^\emptyset}{X_0^\emptyset)}}\bigg) \right] \\
          &=  \E \Bigg[ \ln \big(\p{\theta}{T}(X_{T}^\emptyset)\big)  
          + \ln\big(\pcond{\theta}{0}{1}\argpcond{(X_0^\emptyset}{X_1^\emptyset)}\big)
          - \ln\big(\pcond{\emptyset}{1}{0}\argpcond{(X_1^\emptyset}{X_0^\emptyset)} \big) \\
          & \quad \left. + \ln\big(\pcond{\emptyset}{1}{0}\argpcond{(X_1^\emptyset}{X_0^\emptyset)} \big) - \ln\big(\pcond{\emptyset}{T}{0}\argpcond{(X_T^\emptyset}{X_0^\emptyset)}\big) + \sum_{t = 2}^T \ln \bigg(\frac{\pcond{\theta}{t-1}{t}\argpcond{(X_{t-1}^\emptyset}{X_{t}^\emptyset)} }{\pcond{\emptyset}{t-1}{t,0}\argpcond{(X_{t-1}^\emptyset}{X_t^\emptyset,X_0^\emptyset)} }\bigg) \right] \\
          &= \E_{} \left[ \ln \bigg(\frac{\p{\theta}{T}(X_{T}^\emptyset)}{\pcond{\emptyset}{T}{0}\argpcond{(X_T^\emptyset}{X_0^\emptyset)}}\bigg) + 
          \ln\big(\pcond{\theta}{0}{1}\argpcond{(X_0^\emptyset}{X_1^\emptyset)}\big)
          + \sum_{t = 2}^T \ln \bigg(\frac{\pcond{\theta}{t-1}{t}\argpcond{(X_{t-1}^\emptyset}{X_{t}^\emptyset)} }{\pcond{\emptyset}{t-1}{t,0}\argpcond{(X_{t-1}^\emptyset}{X_t^\emptyset,X_0^\emptyset)} }\bigg) \right]\\
          & = \Eb{}{\ln \bigg(\frac{\p{\theta}{T}(X_{T}^\emptyset)}{\pcond{\emptyset}{T}{0}\argpcond{(X_T^\emptyset}{X_0^\emptyset)}}\bigg)}+ \Eb{}{\ln\big(\pcond{\theta}{0}{1}\argpcond{(X_0^\emptyset}{X_1^\emptyset)}\big)} \\
          & \quad + \sum_{t = 2}^T\Eb{}{\ln \bigg(\frac{\pcond{\theta}{t-1}{t}\argpcond{(X_{t-1}^\emptyset}{X_{t}^\emptyset)} }{\pcond{\emptyset}{t-1}{t,0}\argpcond{(X_{t-1}^\emptyset}{X_t^\emptyset,X_0^\emptyset)} }\bigg)}\\
          & = 
          \int_{\R^{\di}}\int_{\R^{\di}} \ln \bigg(\frac{\p{\theta}{T}(x_{T})}{\pcond{\emptyset}{T}{0}\argpcond{(x_T}{x_0)}}\bigg) \pcond{\emptyset}{T}{0}\argpcond{(x_T}{x_0)} \p{\emptyset}{0}(x_{0})
          \, \d x_{0} \, \d x_{T}  
          + \Eb{}{\ln\big(\pcond{\theta}{0}{1}\argpcond{(X_0^\emptyset}{X_1^\emptyset)}\big)} \\
          & \quad + \sum_{t = 2}^T
          \int_{\R^{\di}}\int_{\R^{\di}}\int_{\R^{\di}}
          \ln \bigg(\frac{\pcond{\theta}{t-1}{t}\argpcond{(x_{t-1}}{x_{t})}}{\pcond{\emptyset}{t-1}{t,0}\argpcond{(x_{t-1}}{x_t,x_0)}}\bigg) \pcond{\emptyset}{t-1}{t,0}\argpcond{(x_{t-1}}{x_t,x_0)} \\
          & \quad \p{\emptyset}{0,t}(x_{0},x_{t}) 
          \, \d x_{0} \, \d x_{t-1} \, \d x_{t}  \\
          & = -\Eb{} {\infdiv{\pcond{\emptyset}{T}{0}\argpcond{(\cdot}{X_0^\emptyset)}}{\Pi}} +\Eb{}{\ln \pcond{\theta}{0}{1}\argpcond{(X_0^\emptyset}{X_1^\emptyset)}}\\
        &\quad - \sum_{t = 2}^T\Eb{}{\infdiv{\pcond{\emptyset}{t-1}{t,0}\argpcond{(\cdot}{X_t^\emptyset,X_0)}}{\pcond{\theta}{t-1}{t}\argpcond{(\cdot}{X_{t}^\emptyset)}}}
    \end{split}
\end{equation}
\cfload.
\end{cproof}

\begin{remark}[Explanations for \cref{lemma:upperboundE}]
\label{remark:loglike}
In this remark we 
explain the relevance of \cref{lemma:upperboundE} in the context of \DDPMs\ with Markov assumptions
and
provide intuitive explanations for the terms appearing in \cref{lemma:upperboundE:thesis}.
Roughly speaking, 
\cref{lemma:upperboundE} provides 
for all $\theta \in \R^{\fd}$
an upper bound for 
the \ENLL\ 
$
    \negloglike[\big]{\p{\emptyset}{0}}{\p{\theta}{0}} 
$
of the \PDF\ of the initial value of the forward process 
$\p{\emptyset}{0}$
with respect to the \PDF\ of the terminal value of the backward process 
$\p{\theta}{0}$.
As illustrated in \cref{lemma:loglike:item4,lemma:loglike:item3} in \cref{lemma:loglike} the 
\ENLL\ 
$
    (
        \R^{\fd}
        \ni 
        \theta 
        \mapsto
        \negloglike[\big]{\p{\emptyset}{0}}{\p{\theta}{0}}
        \in 
        \R
    )
$
can be considered to be a natural training objective for the goal explained in \cref{remark:setting0} of finding parameters 
$\opttheta \in \R^\fd$ for the backward process such that
\begin{equation}
\begin{split}
    \p{\opttheta}{0}
\approx
    \p{\emptyset}{0}
\end{split}
\end{equation}
(cf.\ \cref{remark:setting0:eq1} in \cref{remark:setting0}).
The estimate in \cref{lemma:upperboundE} now allows to minimize this training objective 
by minimizing the upper bound.
The upper bound, in turn, can be minimized by separately minimizing each term appearing in it.
Crucially, each term in the upper bound only depends on a single step transition probability of the backward process, and the resulting training objectives to be minimized are therefore much simpler than the original one.

Such a loss term decomposition was first proposed in \cite{sohl2015deep} and further refined in \cite{ho2020denoising} to the bound in \cref{lemma:upperboundE}.
We illustrate how it can be used to train the parameters of the backward process in \cref{sec:ddpm_simple} below.

We now provide some very rough interpretations for the new terms appearing in the upper bound.
\begin{enumerate}[label=(\roman*)]
    
    \item
    The terms
    $
        \Eb{}{\infdiv{\pcond{\emptyset}{t-1}{t,0}\argpcond{(\cdot}{X_t^\emptyset,X_0^\emptyset)}}{\pcond{\theta}{t-1}{t}\argpcond{(\cdot}{X_{t}^\emptyset)}}}
    $,
    $t \in \{2,3,\ldots,T\}$,
    $\theta \in \R^{\fd}$,
    measure 
    the difference between 
    backward transition kernels of the forward process given the initial value of the forward process
    and
    transition kernels of the backward process.
    Minimizing these terms should make the distribution of the backward process approximate the distribution of the forward process.

    \item
    The terms
    $-\Eb{}{\ln \big( \pcond{\theta}{0}{1}\argpcond{(X_0^\emptyset}{X_1^\emptyset)}\big)}
        =
    \Eb{}{\negloglike[\big] {\pcond{\emptyset}{0}{1}\argpcond{(\cdot}{X_1^\emptyset)}}{\pcond{\theta}{0}{1}\argpcond{(\cdot}{X_1^\emptyset)}}}$, 
    $\theta \in \R^{\fd}$,
    measure how accurately the backward process can recover the initial value from the first noisy step. Minimizing this term encourages the model to learn an effective denoising process for the final step, where it aims to reconstruct the original input from its slightly noisy version.

    \item 
    The term 
        $\Eb{} {\infdiv{\pcond{\emptyset}{T}{0}\argpcond{(\cdot}{X_0^\emptyset)}}{\Pi}}$
    measures how much the distribution of the terminal value of the forward process differs from the  distribution of the initial value of the backward process.
    This term has no learnable parameters and consequently can be ignored during training.
\end{enumerate}

\end{remark}

\subsection{A first simplified DDPM generative method}
\label{sec:ddpm_simple}

In this section we discuss in \cref{setting_base,remark:setting_base} a \DDPM\ methodology which makes use of the upper bound in \cref{lemma:upperboundE} to minimize the \ENLL\ of the \PDF\ of the initial value of the forward process
with respect to the \PDF\ of the terminal value of the backward process in \cref{setting01}.
\cref{setting_base} can be regarded as a simplified version of the \DDPM\ methodologies proposed in \cite{ho2020denoising,sohl2015deep}.

\cfclear
\begin{method}[A simplified \DDPM\ generative method]
    \label{setting_base}
    Assume \cref{setting01}, assume $T>1$, let $\ds \in \N$, $\gamma \in (0, \infty)$, 
let  
$\fL \colon \R^\fd \times \{1,\ldots,T\} \times \R^\di \times \R^\di \times \ldots \times \R^\di  \to \R$
satisfy 
for all $ \theta \in \R^{ \fd } $, $x_0, x_1,\ldots,x_T \in \R^\di$
that
\begin{equation}
      \fL( \theta, t, x_0, x_1,\ldots,x_T)  = 
      \begin{cases}
      -\ln (\pcond{\theta}{0}{1}\argpcond{(x_0}{x_1)})
      &  \colon t=1\\
      \infdiv{\pcond{\emptyset}{t-1}{t,0}\argpcond{(\cdot}{x_t,x_0)}}{\pcond{\theta}{t-1}{t}\argpcond{(\cdot}{x_{t})}}
      & \colon t>1,
\end{cases}
\end{equation}
let $\fG \colon  \R^\fd \times \{1,\ldots,T\} \times \R^\di \times \R^\di \times \ldots \times \R^\di \to \R^{ \fd }$ satisfy for all 
$t \in \{1, \ldots, T\}$,
$x_0, x_1,\ldots,x_T \in \R^\di$,
$\theta \in \R^\fd$ with $\fL( \cdot, t, x_0, x_1,\ldots,x_T)$ differentiable at $\theta$
that
\begin{equation}
\fG( \theta, t, x_0, x_1,\ldots,x_T) =  ( \nabla_{ \theta } \fL )( \theta, t, x_0, x_1,\ldots,x_T),
\end{equation}
let
$\datax_{n,i}= (\datax_{n,i,t})_{t\in \{0,1,\ldots,T\}} \colon \allowbreak \{0,1,\ldots,T\} \times \Omega \to \R^\di$, $ n, i \in \N$,
be identically distributed stochastic processes, assume that $\datax_{1,1}$ and $X^{\emptyset}$ are identically distributed,
assume that $(\datax_{n,i})_{(n,i) \in \N^2}$ and $(X^{\theta})_{\theta \in \R^\fd}$ are independent,
let  
$\randT_{n} \colon \Omega \to \{1,2,\ldots,T\}$, $ n\in \N$,
be independent $\mathcal{U}_{\{1,2,\ldots,T\}}$-distributed random variables,
and let 
$ \Theta  \colon \N_0 \times \Omega \to \R^{ \fd } 
$
be a stochastic process which satisfies 
for all $n\in\N$ that
\begin{equation}
\Theta_{n}=\Theta_{n-1}-\gamma \left[   \frac{ 1 }{ \ds }   \smallsum\limits_{ i= 1 }^{ \ds }
  \fG ( \Theta_{n-1}, \randT_{n}, \datax_{n,i,0}, \datax_{n,i,1}, \ldots, \datax_{n,i,T} )
 \right]
\end{equation}
\cfout.
\end{method}

\begin{remark}[Explanations for \cref{setting_base}]
\label{remark:setting_base}
In this remark we provide some intuitive interpretations for the mathematical objects appearing in \cref{setting_base} and roughly explain in what sense \cref{setting_base} can be used for generative modelling.

Roughly speaking, in \cref{setting_base} we aim to train the parameters of the backward process $(X^{\theta})_{\theta \in \R^\fd}$
by minimizing 
the training objective 
$
    (
        \R^{\fd}
        \ni 
        \theta 
        \mapsto
        \negloglike[\big]{\p{\emptyset}{0}}{\p{\theta}{0}}
        \in 
        \R
    )
$
in \cref{lemma:upperboundE}.
From this perspective, observe that 
\begin{enumerate}[label=(\roman*)]
    \item we think of
    $\fL$ as the loss used in the training which is based on the trainable terms of the upper bound in \cref{lemma:upperboundE},
    \item we think of 
    $\fG$ as the generalized gradient of the loss $\fL$ with respect to the trainable parameters,
    \item we think of
    $\datax_{n,i}$, $ n, i \in \N$, as random samples of the forward process used for training,
    \item we think of
    $\randT_{n}$, $ n\in \N$, as random times used to determine which terms of the upper bound are considered in each training step,
    \item we think of
    $(\Theta_{n})_{n\in\N_0}$ as the training process for the parameters of the backward process given by an \SGD\ process for the generalized gradient $\fG$
    with learning rate $\gamma$,
    batch size $\ds$, and
    training data 
    $(\randT_{n}, \datax_{n,i,0}, \datax_{n,i,1}, \ldots, \datax_{n,i,T})_{(n,i) \in \N^2}$.
\end{enumerate}
Note that in \cref{setting_base} we choose for simplicity the \SGD\ method to train the parameters of the backward process.
In practice typically other, more sophisticated, \SGD-type methods are used
(cf., \eg 
    \cite[Section 5]{bach2024learning},
    \cite[Section 7]{jentzen2023mathematical},
    \cite{ruder2016overview}, and
    \cite[Section 14]{shalev2014understanding}
for introductions to such \SGD-type methods).

Note that the objective that the \SGD\ process aims to minimize is given
for all 
    $\theta \in \R^{ \fd } $
by
\begin{equation}
\label{T_B_D}
\begin{split}
    &\Eb{}{\fL(\theta,\randT_{1}, \datax_{1,1,0}, \datax_{1,1,1}, \ldots, \datax_{1,1,T})} \\
&=
    \frac{1}{T}
    \pr*{
        -\Eb{}{\ln \big( \pcond{\theta}{0}{1}\argpcond{(X_0^\emptyset}{X_1^\emptyset)}\big)}
        + \sum_{t = 2}^T\Eb{}{\infdiv{\pcond{\emptyset}{t-1}{t,0}\argpcond{(\cdot}{X_t^\emptyset,X_0^\emptyset)}}{\pcond{\theta}{t-1}{t}\argpcond{(\cdot}{X_{t}^\emptyset)}}}
    }
. 
\end{split}
\end{equation}
The upper bound in \cref{lemma:upperboundE} indicates that minimizing this objective roughly allows to minimize the \ENLL\ 
$
    (
        \R^{\fd}
        \ni 
        \theta 
        \mapsto
        \negloglike[\big]{\p{\emptyset}{0}}{\p{\theta}{0}}
        \in 
        \R
    )
$
of the \PDF\ of the initial value of the forward process with respect to the \PDF\ of the terminal value of the backward process.

For large enough $N \in \N$ we therefore expect that 
$X^{\Theta_{N}}_0$ is roughly distributed according to the distribution we would like to sample from (cf. \cref{lemma:loglike:item3,lemma:loglike:item4} in \cref{lemma:loglike} and \cref{remark:loglike}).
Loosely speaking, creating a new generative sample in the context of \cref{setting_base} then corresponds to sampling a random realization of $X^{\Theta_{N}}_0$.
\end{remark}

\section{DDPMs with Gaussian noise}
\label{sec:DDPM_gaussian}

In this section we consider \DDPMs\ with Markov assumptions 
when the transition kernels are given by Gaussian distributions. %
The setup and methodology considered in this section essentially correspond to the one proposed in \cite{ho2020denoising}.
Intuitively speaking, in this setup we think that the forward process gradually adds Gaussian noise to a training sample
which the backward process then aims to gradually remove to recover the original training sample.

We first discuss some elementary properties of Gaussian distributions in \cref{sec:gaussian}.
We then motivate and describe a \DDPM\ framework involving such Gaussian distributions as transition kernels in \cref{sec:DDPM_gaussian_FW}.
Thereafter, we discuss some consequences of this choice of transition kernels on distributions of the forward process in \cref{sec:cond_forward} and on the upper bound for the training objective from \cref{lemma:upperboundE} above in \cref{sec:Obj_gaussian}.
Motivated by the previous sections we then describe a training and generation scheme for \DDPMs\ with Gaussian noise in \cref{sec:scheme_gaussian}.
Finally, in \cref{sec:architecture} we point to some possible choices of architectures for the \ANNs\ appearing in the method description in \cref{sec:scheme_gaussian}.

\subsection{Properties of Gaussian distributions}
\label{sec:gaussian}

In this section we recall some elementary and well-known properties of Gaussian distributions which will be used in the definition of transition kernels throughout \cref{sec:DDPM_gaussian}.
We start by recalling the definition of \PDFs\ of Gaussian distributions.

\begin{definition}[Gaussian \PDFs]\label{def:gaussian}
    Let $ \di \in \N$ and\footnote{
        Note that for every $n,m \in \N$, $A \in \R^{n \times m}$ we have that $A^{*} \in \R^{m \times n}$ is the transpose of $A$.
    } let
    $\sis=\{Q \in \R^{\di \times \di } \colon Q^{*}=Q \text{ and } (\forall\, v \in \R^\di \backslash \{0\} \colon v^*Qv > 0)\}$. %
    Then we denote by $\n \colon \R^{\di} \times \R^{\di} \times \sis \to \R$ the function which satisfies
     for all $x,v \in \R^{\di}$, $Q \in \sis$ 
    that
\begin{equation}
    \n(x,v,Q)= (2\pi)^{-\frac{\di}2} \det(Q)^{-\frac12} \exp \bigl(- \tfrac{1}{2} (x-v)^{*} Q^{-1} (x-v)  \bigr)
\end{equation}
    and for every
    $v \in \R^{\di}$, $Q \in \sis$ 
    we call 
    $\n(\cdot,v,Q) \colon \R^{\di} \to \R$
    the \PDF\ of the Gaussian distribution with mean $v$ and covariance matrix $Q$.
\end{definition}

\subsubsection{On Gaussian transition kernels}

The next two results illustrate how distributions propagate in Markov chains with transition kernels involving Gaussian distributions. 
We first present a result on the level of \PDFs\ in \cref{lemma:condprob} and then state the consequence on the level of random variables in \cref{cor:condprob}.

\cfclear
\begin{lemma}
    \label{lemma:condprob}
    Let $\di \in \N$, let 
    $\sis=\{Q \in \R^{\di \times \di } \colon Q^{*}=Q \text{ and } (\forall\, v \in \R^\di \backslash \{0\} \colon v^*Qv > 0)\}$,
    and
    let $\mu_1, \mu_2 \in \R^{\di}$, 
    $A \in \R^{\di \times \di }$, 
    $\Sigma_1, \Sigma_2 \in \sis$.
    Then it holds for all $x \in \R^\di$ that
    \begin{equation}\label{lemma:condprob:thesis}
    \begin{split}
    \int_{\R^\di} \n(x, A y+ \mu_1 , \Sigma_1 ) \n(y,\mu_2, \Sigma_2 ) \, \d y  = \n(x,A \mu_2 + \mu_1, A \Sigma_2 A^* + \Sigma_1 )
    \end{split}
    \end{equation}
    \cfout.
\end{lemma}
\begin{cproof}{lemma:condprob}
    \Nobs that, \eg \cite[(2.115)]{10.5555/1162264} shows \cref{lemma:condprob:thesis}.
\end{cproof}

\cfclear
\begin{cor}
    \label{cor:condprob}
    Let $\di \in \N$, let 
    $\sis=\{Q \in \R^{\di \times \di } \colon Q^{*}=Q \text{ and } (\forall\, v \in \R^\di \backslash \{0\} \colon v^*Qv > 0)\}$,
    let 
    $\mu_1, \mu_2 \in \R^{\di}$, 
    $A \in \R^{\di \times \di }$, 
    $\Sigma_1, \Sigma_2 \in \sis$,
    let $(\Omega, \cF, \P)$ be a probability space, 
    let $X \colon \Omega \to \R^\di$ and $Y \colon \Omega \to \R^\di$ be random variables,
    and assume for all
        $B \in \cB(\R^\di)$
    that
    \begin{equation}
    \label{T_B_D}
    \begin{split}
        \P(Y \in B) = \int_B \n(y, \mu_2, \Sigma_2) \, \d y
        \qandq
        \P(X \in B | Y) \stackrel{\P\text{-a.s.}}{=} \int_B \n(x, A Y + \mu_1, \Sigma_1) \, \d x
    \end{split}
    \end{equation}
    \cfload. 
    Then it holds for all
        $B \in \cB(\R^\di)$
    that
    \begin{equation}
    \label{cor:condprob:concl1}
    \begin{split}
        \P(X \in B)
    =
        \int_B \n(x, A \mu_2 + \mu_1, A \Sigma_2 A^* + \Sigma_1) \, \d x
        .
    \end{split}
    \end{equation}
    
\end{cor}

\begin{cproof}{cor:condprob}
    \Nobs that \cref{lemma:condprob} establishes \eqref{cor:condprob:concl1}.
\end{cproof}

\subsubsection{Explicit constructions for Gaussian transition kernels}
The result below shows an explicit way to simulate a step in a Markov chain with Gaussian transition kernels based on realizations of standard normal random variables.

\cfclear
\begin{lemma}\label{lemma:indZ}
    Let $\di \in \N$, let 
    $\sis=\{Q \in \R^{\di \times \di } \colon Q^{*}=Q \text{ and } (\forall\, v \in \R^\di \backslash \{0\} \colon v^*Qv > 0)\}$,
    let 
    $\mu \colon \R^{\di}  \to \R^{\di}$ and 
    $\Sigma \colon \R^{\di}  \to \sis$ be functions,
    let $(\Omega, \cF, \P)$ be a probability space, 
    let $X \colon \Omega \to \R^\di$, $Y \colon \Omega \to \R^\di$, and $Z \colon \Omega \to \R^\di$ be random variables,
    and assume for all
        $B \in \cB(\R^\di)$
    that
    \begin{equation}
    \label{lemma:indZ:Z}
    \begin{split}
        \P(X \in B | Y) \stackrel{\P\text{-a.s.}}{=} \int_B \n(x, \mu(Y), \Sigma(Y)) \, \d x
        \qandq X = \mu (Y) + \big(\Sigma(Y)\big)^{\nicefrac12}Z
    \end{split}
    \end{equation}
    \cfload.
    Then 
    \begin{enumerate}[label=(\roman*)]
        \item \label{lemma:indZ:1} it holds for all $B \in \cB(\R^{\di})$  that
    $\Prob{}(Z \in B)= \int_B \n(x,0,\mathbb{I})\, \d x$ and
    \item \label{lemma:indZ:2} it holds that $Z$ and $Y$ are independent.
    \end{enumerate}
\end{lemma}
\begin{cproof}{lemma:indZ}
    \Nobs that \cref{lemma:indZ:Z}, the fact that for all
    $y \in \R^{\di}$ it holds that $\Sigma(y)=\Sigma^*(y)$,
    and, \eg \cite[Theorem 8.38]{klenke2013probability} show that for all measurable and bounded $f\colon \R^{\di} \to \R^{\di} $ it holds $\P$-a.s.\ that
    \begin{equation}\label{lemma:indZ:exp} 
    \begin{split}
        &\E[f(Z)|Y]=\E\big[{f\big([\Sigma(Y)}]^{-\nicefrac12}(X - \mu (Y) )\big)|Y\big]\\
        & = \int_{\R^{\di}} f\big([\Sigma(Y)]^{-\nicefrac12}(x - \mu (Y) )\big) \n(x, \mu(Y), \Sigma(Y)) \, \d x \\
        & =  \int_{\R^{\di}} f(z) \n\big(\mu (Y) + [\Sigma(Y)]^{\nicefrac12}z, \mu(Y), \Sigma(Y)\big) \det(\Sigma(Y))^{\nicefrac12} \, \d z\\
        & = \int_{\R^{\di}} f(z) 
        (2\pi)^{-\nicefrac{\di}2} \det(\Sigma(Y))^{-\nicefrac12} \exp \Big(- \tfrac{1}{2} \big(\mu (Y) + [\Sigma(Y)]^{\nicefrac12}z-\mu (Y)\big)^{*} [\Sigma(Y)]^{-1}\\
        &\quad \big(\mu (Y) + [\Sigma(Y)]^{\nicefrac12}z-\mu (Y)\big)  \Big) \det(\Sigma(Y))^{\nicefrac12} \, \d z\\
        & = \int_{\R^{\di}} f(z) 
        (2\pi)^{-\nicefrac{\di}2}  \exp \big(- \tfrac{1}{2} ( z)^{*} \mathbb{I} ( z)\big)= \int_{\R^{\di}} f(z) \n(z, 0, \mathbb{I}) \, \d z = \E[f(Z)].
    \end{split}
    \end{equation}
    This assures that for all $B \in \cB(\R^{\di})$ it holds that
    \begin{equation}
        \Prob{}(Z \in B) = \E[\mathbbm{1}_{B}(Z)] = \E[\mathbbm{1}_{B}(Z)| Y] = \int_{B} \n(z, 0, \mathbb{I}) \, \d z.
    \end{equation}
    This demonstrates  \cref{lemma:indZ:1}.
    \Moreover \cref{lemma:indZ:exp} proves that for all measurable and bounded $f\colon \R^{\di} \to \R^{\di} $ and $g\colon \R^{\di} \to \R^{\di} $ it holds $\P$-a.s.\ that
    \begin{equation}
    \begin{split}
            &\E[f(Z)g(Y)] =\E[\E[f(Z)g(Y)|Y]]=\E[g(Y)\E[f(Z)|Y]]=\E[g(Y)\E[f(Z)]] \\ &=\E[g(Y)]\E[f(Z)].
    \end{split}
    \end{equation}
    This and, \eg \cite[Theorem 3D]{grimmett2014probability} imply that $Z$ and $Y$ are independent. This establishes 
    \cref{lemma:indZ:2}.
\end{cproof}

\subsubsection{Bayes rule for Gaussian distributions}

The next two results illustrate an explicit form of the Bayes rule for Gaussian distributions.
We first present a result on the level of \PDFs\ in \cref{lemma:prob_bayes} and then state the consequence on the level of random variables in \cref{cor:prob_bayes}.

\cfclear
\begin{lemma}%
    \label{lemma:prob_bayes}
    Let $\di \in \N$,
    let 
    $\sis = \{Q \in \R^{\di \times \di } \colon Q^{*}=Q \text{ and } (\forall\, v \in \R^\di \backslash \{0\} \colon v^*Qv > 0)\}$,
    let $\mu_1, \mu_2 \in \R^{\di}$, $A \in \R^{\di \times \di }$, $\Sigma_1, \Sigma_2 \in \sis$,
    and let $\Sigma_3 \in \R^{\di \times \di }$ satisfy $\Sigma_3=\Sigma_2 A^*(A \Sigma_2 A^* + \Sigma_1)^{-1}$. 
    Then it holds for all $x,y \in \R^\di$ that
    \begin{equation}\label{lemma:prob_bayes:thesis}
    \begin{split}
    \frac{\n(x, A y+ \mu_1 , \Sigma_1 )\n(y,\mu_2, \Sigma_2 )}{\n(x,A \mu_2 + \mu_1, A \Sigma_2 A^* + \Sigma_1 )}
    = \n(y, \Sigma_3 (x- A^* \mu_2-\mu_1) +\mu_2, \Sigma_2-\Sigma_3 A \Sigma_2^*)
    \end{split}
    \end{equation}
    \cfout.
\end{lemma}
\begin{cproof}{lemma:prob_bayes}
    \Nobs that, \eg \cite[(2.116)]{10.5555/1162264} implies \cref{lemma:prob_bayes:thesis}.
\end{cproof}

\cfclear
\begin{cor}
    \label{cor:prob_bayes}
 Let $\di \in \N$, let 
    $\sis=\{Q \in \R^{\di \times \di } \colon Q^{*}=Q \text{ and } (\forall\, v \in \R^\di \backslash \{0\} \colon v^*Qv > 0)\}$,
    let 
    $\mu_1, \mu_2 \in \R^{\di}$, 
    $A \in \R^{\di \times \di }$, 
    $\Sigma_1, \Sigma_2 \in \sis$, let $\Sigma_3 \in \R^{\di \times \di }$ satisfy $\Sigma_3=\Sigma_2 A^*(A \Sigma_2 A^* + \Sigma_1)^{-1}$,
    let $(\Omega, \cF, \P)$ be a probability space, 
    let $X \colon \Omega \to \R^\di$ and $Y \colon \Omega \to \R^\di$ be random variables,
    and assume for all
        $B \in \cB(\R^\di)$
    that
    \begin{equation}
    \label{T_B_D}
    \begin{split}
        \P(Y \in B) = \int_B \n(y, \mu_2, \Sigma_2) \, \d y
        \qandq
        \P(X \in B | Y) \stackrel{\P\text{-a.s.}}{=} \int_B \n(x, A Y + \mu_1, \Sigma_1) \, \d x
    \end{split}
    \end{equation}
    \cfload.
    Then it holds for all
        $B \in \cB(\R^\di)$
    that
    \begin{equation}
    \label{cor:prob_bayes:concl1}
    \begin{split}
        \P(Y \in B| X) \stackrel{\P\text{-a.s.}}{=}
        \int_B \n(y, \Sigma_3 (X- A^* \mu_2-\mu_1) +\mu_2, \Sigma_2-\Sigma_3 A \Sigma_2^*) \, \d x
        .
    \end{split}
    \end{equation}
\end{cor}
\begin{cproof}{cor:prob_bayes}
    \Nobs that \cref{cor:condprob}, \cref{lemma:prob_bayes},  and Bayes' Theorem establish \eqref{cor:prob_bayes:concl1}.
\end{cproof}

\subsubsection{KL divergence between Gaussian distributions}
In the next result we recall a formula for the KL divergence between two \PDFs\ of Gaussian distributions.

\cfclear
\begin{lemma}[KL divergence between Gaussian distributions]
    \label{lemma_kl_gaussian}
Let $\di \in \N$, let $\sis=\{Q \in \R^{\di \times \di } \colon Q^{*}=Q \text{ and } (\forall\, v \in \R^\di \colon v^*Qv> 0)\}$, and let $\mu_1, \mu_2 \in \R^{\di}$, $\Sigma_1, \Sigma_2 \in \sis$. Then
\begin{equation}\label{lemma_kl_gaussian:thesis}
    \begin{split}
        &\infdiv{\n(\cdot, \mu_1,\Sigma_1)}{\n( \cdot,\mu_2,\Sigma_2)} \\
        & \quad =\tfrac{1}{2}\left[\ln\Big(\frac{\det\Sigma_2}{\det\Sigma_1}\Big) - \di + \text{tr}(\Sigma_2^{-1}\Sigma_1)
        + (\mu_2-\mu_1)^* \Sigma_2^{-1} (\mu_2-\mu_1)\right]
    \end{split}
\end{equation}
\cfout.
\end{lemma}
\begin{cproof}{lemma_kl_gaussian}
    \Nobs that, \eg \cite[Section 9]{duchi2007derivations} establishes \cref{lemma_kl_gaussian:thesis}.
\end{cproof}

\subsection{Framework for DDPMs with Gaussian noise}
\label{sec:DDPM_gaussian_FW}

In this section we present in \cref{setting1} a framework for \DDPMs\ with Markov assumptions when the transition kernels are given by Gaussian distributions.
In \cref{lemma:constructive} we then show a constructive way to sample the forward and backward processes in this setting using standard normal random variables.

\cfclear
\begin{setting}[\DDPMs\ with Gaussian transition kernels]
    \label{setting1}
    Assume \cref{setting01}, let
    $\sis=\{Q \in \R^{\di \times \di } \colon Q^{*}=Q \text{ and } (\forall\, v \in \R^\di \backslash \{0\} \colon v^*Qv > 0)\}$,
    let 
    $
        \al_1,\ldots,\al_T
    \in 
        [0,1)
    $,
    for every $\theta \in \R^{\fd}$ let $ \mu^{\theta}=( \mu^{\theta}_t)_{t \in\{1,\ldots,T\}}\colon \R^{\di} \times\{1,\ldots, T\} \to  \R^{\di}$ and $ \Sigma^{\theta}=( \Sigma^{\theta}_t)_{t \in\{1,\ldots,T\}}\colon \R^{\di} \times\{1,\ldots, T\} \to \sis$ be measurable functions,
    and
    assume for all $t \in \{1,\ldots,T\}$, $x_{t-1},x_t  \in \R^{\di}$ that
    \begin{equation}\label{setting1:prob}
        \pcond{\emptyset}{t}{t-1}\argpcond{(x_t}{x_{t-1})}=\n(x_t, \sqrt{\al_t} x_{t-1}, (1-\al_t) \mathbb{I}),
    \end{equation}
    \begin{equation}\label{setting1:prob1}
        \Pi=\n(\cdot,0,\mathbb{I}),
        \qandq
        \pcond{\theta}{t-1}{t}\argpcond{(x_{t-1}}{x_{t})}=\n(x_{t-1}, \mu^{\theta}_t( x_{t}), \Sigma^{\theta}_t( x_{t})) 
    \end{equation}
    \cfout.
\end{setting}

\begin{lemma}[Constructive forward and backward processes in \DDPMs] \label{lemma:constructive}
    Assume \cref{setting1}.
    Then for all $\theta \in \R^\fd \cup \{\emptyset\}$ there exist 
    i.i.d.\ standard normal random variables
    $\Zt{t}^{\theta}\colon \Omega \to \R^{ \di }$, $t \in\{0,1,\ldots,T+1\}$, such that
    \begin{enumerate}[label=(\roman*)]
        \item \label{lemma:constructive:item1}
        for all $t \in\{1,\ldots,T\}$ it holds that $\Xt{\emptyset}_{t-1}$ and $\Zt{t}^{\emptyset}$ are independent and
        \begin{equation}
    \Xt{\emptyset}_{t}= \sqrt{\al_t}\Xt{\emptyset}_{t-1} + \sqrt{1-\al_t} \Zt{t}^{\emptyset}
        \end{equation}
        and
        \item \label{lemma:constructive:item2}
         for all $t \in\{1,\ldots,T\}$ it holds that 
        \begin{equation}
    \Xt{\theta}_{t-1}=\mu^{\theta}_t(\Xt{\theta}_{t}) +  \big(\Sigma^{\theta}_t(\Xt{\theta}_{t})\big)^{\nicefrac12} \Zt{t}^{\theta} \qandq \Xt{\theta}_{T}=\Zt{T+1}^{\theta}.
        \end{equation}
    \end{enumerate}
\end{lemma}
\begin{cproof}{lemma:constructive}
    \Nobs that \cref{lemma:indZ} and \cref{setting1:prob} assure \cref{lemma:constructive:item1}. \Moreover \cref{lemma:indZ} and \cref{setting1:prob1} show \cref{lemma:constructive:item2}.
\end{cproof}

\begin{remark}[Explanations for \cref{setting1}]
    \label{remark:setting1}
    In \cref{setting1} we specify the transition densities in \DDPMs\ with Markov assumptions in \cref{setting01} as certain Gaussian \PDFs. 

    \Cref{lemma:constructive:item1} in \cref{lemma:constructive} shows that the distribution of the forward process specified in \cref{setting1:prob} can be realized by gradually perturbing the state of the forward process with Gaussian noise.
    In particular, for every $t \in \{1, 2, \ldots, T\}$ the number $(1-\al_t)$ measures the amount of Gaussian noise added in the $t$-th step of the forward process.

    On the other hand, \cref{lemma:constructive:item2} in \cref{lemma:constructive} shows that the distribution of the backward process specified in \cref{setting1:prob1} can be realized by starting at a standard normally distributed random variable and then proceeding with transformations involving Gaussian noise.
    For every 
        $t \in \{1, 2, \ldots, T\}$ 
    the functions $(\mu^\theta_t)_{\theta \in \R^{\fd}}$ specify the mean transformation in the $t$-th step of the backward process 
    and 
    the functions $(\Sigma^\theta_t)_{\theta \in \R^{\fd}}$ specify the Gaussian noise added in the  $t$-th step of the backward process.

\end{remark}

\subsection{Distributions of the forward process in DDPMs with Gaussian noise}
\label{sec:cond_forward}

In this section we discuss some consequences of the choice of transition densities in \cref{setting1} on \PDFs\ of the forward process.

\subsubsection{Conditional distributions going forward}

In \cref{lemma:prodq} below we show that in \cref{setting1} the conditional distribution of any time step of the forward process given the initial value of the forward process is again given by a Gaussian distribution.
As a consequence of \cref{lemma:prodq}, we obtain in \cref{cor:randomvar:prodq} that to sample a realization of an arbitrary step of the forward process it suffices to sample a random variable from the initial distribution and a further independent standard normal random variable.

\begin{lemma}[Multi-step transition density of the forward process]\label{lemma:prodq}
Assume \cref{setting1} and let  $ \tal_1, \ldots, \tal_T \in [0,1)$
    satisfy for all $t \in \{1, \ldots, T\}$ that
        $\tal_t= \textstyle\prod_ {s=1}^t\al_s$.
Then it holds for all $t \in \{1,\ldots,T\}$, $x_0,x_t\in \R^{\di}$ that
\begin{equation}\label{lemma:prodq:tesi}
         \pcond{\emptyset}{t}{0}\argpcond{(x_{t}}{x_{0})}=\n(x_{t},\sqrt{\tal_t} x_{0}, (1-\tal_t) \mathbb{I}).
\end{equation}
\end{lemma}
\begin{cproof}{lemma:prodq}
    We prove \cref{lemma:prodq:tesi} by induction.
    \Nobs that the fact that for all $t \in \{1,\ldots,T\}$, $x_t,x_{t-1} \in \R^{\di}$ it holds that 
    \begin{equation}
        \pcond{\emptyset}{t}{t-1}\argpcond{(x_t}{x_{t-1})}=\n(x_t, \sqrt{\al_t} x_{t-1}, (1-\al_t) \mathbb{I})
    \end{equation}
    implies that
 for all $x_1,x_0 \in \R^{\di}$ it holds that
\begin{equation}
        \pcond{\emptyset}{1}{0}\argpcond{(x_{1}}{x_{0})}=\n(x_{1}, \sqrt{\tal_1} x_{0}, (1-\tal_1) \mathbb{I}).
\end{equation}
For the induction step let $t \in \{2,3,\ldots,T\}$ and assume that for all $x_{t-1},x_0 \in \R^{\di}$ it holds that
\begin{equation}\label{lemma:prodq:induction_step}
         \pcond{\emptyset}{t-1}{0}\argpcond{(x_{t-1}}{x_{0})}=\n(x_{t-1},\sqrt{\tal_{t-1}} x_{0}, (1-\tal_{t-1}) \mathbb{I}).
\end{equation}
\Nobs that \cref{lemma:prodq:induction_step} and \cref{lemma:condprob} assure that for all $x_{t},x_0 \in \R^{\di}$ it holds that
\begin{equation}\label{lemma:prodq:integral}
\begin{split}
    \pcond{\emptyset}{t}{0}\argpcond{(x_{t}}{x_{0})} &=\int_{\R^\di} \pcond{\emptyset}{t}{t-1,0}\argpcond{(x_{t}}{x_{t-1},x_{0})}\pcond{\emptyset}{t-1}{0}\argpcond{(x_{t-1}}{x_{0})} \, \d x_{t-1}\\
    & = \int_{\R^\di} \pcond{\emptyset}{t}{t-1}\argpcond{(x_{t}}{x_{t-1})}\pcond{\emptyset}{t-1}{0}\argpcond{(x_{t-1}}{x_{0})} \, \d x_{t-1}\\
    & = \int_{\R^\di} \n(x_t, \sqrt{\al_t} x_{t-1}, (1-\al_t) \mathbb{I}) \n(x_{t-1},\sqrt{\tal_{t-1}} x_{0}, (1-\tal_{t-1}) \mathbb{I})
    \, \d x_{t-1}\\
     & = \n(x_{t}, \sqrt{\tal_t} x_{0}, (1-\tal_t) \mathbb{I}).
\end{split}
\end{equation}
Induction thus establishes \cref{lemma:prodq:tesi}.
\end{cproof}
\begin{cor}[Gaussian random variables]\label{cor:randomvar:prodq}
    Assume \cref{setting1}, let  $ \tal_1, \ldots, \tal_T \in [0,1)$
    satisfy for all $t \in \{1, \ldots, T\}$ that
        $\tal_t= \textstyle\prod_ {s=1}^t\al_s$, and for all $t \in \{1,\ldots,T\}$ let $\mathcal{E}_t \colon \Omega \to \R^{ \di }$ satisfy $X_t^\emptyset = \sqrt{\tal_t}X_0^\emptyset + \sqrt{1-\tal_t} \mathcal{E}_t$.
    Then 
    \begin{enumerate}[label=(\roman*)]
        \item \label{cor:randomvar:prodq:1} it holds for all 
            $t \in \{1,\ldots,T\}$,
            $B \in \cB(\R^{\di})$ that
            $\Prob{}(\mathcal{E}_t \in B)= \int_B \n(x,0,\mathbb{I})\, \d x$ and
        \item \label{cor:randomvar:prodq:2} it holds for all 
            $t \in \{1,\ldots,T\}$
            that $\mathcal{E}_t$ and $X_0^\emptyset$ are independent.    
    \end{enumerate}
\end{cor}
\begin{cproof}{cor:randomvar:prodq}
\Nobs that \cref{lemma:indZ} and \cref{lemma:prodq} prove \cref{cor:randomvar:prodq:1} and \cref{cor:randomvar:prodq:2}.
\end{cproof}

\subsubsection{Terminal distributions}

In this section we illustrate a consequence of \cref{lemma:prodq} on the distribution of the terminal value of the forward process.
We first prove in \cref{cor:prodq} an auxiliary result which then allows us to explain in \cref{remark:prodq} that the terminal distribution of the forward process tends towards a standard normal distribution when, roughly speaking, we add enough Gaussian noise throughout the forward process.

\cfclear
\begin{lemma}%
    \label{cor:prodq}
Let $d \in \N$, let $ \textbf{p} \colon \R^d \to (0,\infty)$ satisfy $\int_{\R^d}  \textbf{p}(x) \, \d x =1$, and let $(\tal_t)_{t \in \N} \subseteq [0,1)$ satisfy $\lim_{t \to \infty} \tal_t = 0$.
Then it holds for all $x \in \R^d$ that
\begin{equation}\label{cor:prodq:tesi}
        {\textstyle{\lim_{t \to \infty }}} \int_{\R^\di} \textbf{p}(x_0) \n(x, \sqrt{\tal_t} x_{0}, (1-\tal_t) \mathbb{I}) \, \d x_0
         = \n(x,0, \mathbb{I})
\end{equation}
\cfout.
\end{lemma}
\begin{cproof}{cor:prodq}
    \Nobs that for all $t \in \N$, $x, x_0 \in \R^d$ it holds that 
    \begin{equation}
        \| \textbf{p}(x_0) \n(x, \sqrt{\tal_t} x_{0}, (1-\tal_t) \mathbb{I}) \| = |\textbf{p}(x_0)|  \| \n(x, \sqrt{\tal_t} x_{0}, (1-\tal_t) \mathbb{I}) \| \leq \textbf{p}(x_0) \sqrt{d}
    \end{equation}
    \cfload.
    This and Lebesgue’s dominated convergence theorem demonstrate that
    \begin{equation}
    \begin{split}
         &{\textstyle{\lim_{t \to \infty }}} \int_{\R^\di} \textbf{p}(x_0) \n(x, \sqrt{\tal_t} x_{0}, (1-\tal_t) \mathbb{I}) \, \d x_0 \\
         &=
        \int_{\R^\di} {\textstyle{\lim_{t \to \infty }}} \textbf{p}(x_0) \n(x, \sqrt{\tal_t} x_{0}, (1-\tal_t) \mathbb{I}) \, \d x_0 = \int_{\R^\di}  \textbf{p}(x_0) \n(x,0, \mathbb{I}) \, \d x_0 = \n(x,0, \mathbb{I}).
    \end{split}
    \end{equation}
\end{cproof}
\begin{remark}[Limiting distribution of the forward process]
    \label{remark:prodq}
    Assume \cref{setting1} and let  $ \tal_1, \ldots, \tal_T \in [0,1)$
    satisfy for all $t \in \{1, \ldots, T\}$ that
        $\tal_t= \textstyle\prod_ {s=1}^t\al_s$.
    Note that 
    \cref{lemma:prodq:tesi}
    implies that
    for all $x_T \in \R^\di$ it holds that
    \begin{equation}
    \begin{split}
        \p{\emptyset}{T}(x_T) &=\int_{\R^\di} \p{\emptyset}{0}(x_0) \pcond{\emptyset}{T}{0}\argpcond{(x_{T}}{x_{0})} \, \d x_{0} = \int_{\R^\di} \p{\emptyset}{0}(x_0) \n(x_T, \sqrt{\tal_T} x_{0}, (1-\tal_T) \mathbb{I}) \, \d x_{0}.
    \end{split}
    \end{equation}
    \cref{cor:prodq} therefore suggests that if $\tal_T \approx 0$ we can expect for all $x_T \in \R^\di$ that
    \begin{equation}
    \begin{split}
        \p{\emptyset}{T}(x_T) 
        \approx \n(x_T,0, \mathbb{I}) 
        = 
        \Pi(x_T).
    \end{split}
    \end{equation}
    Roughly speaking, this shows that the assumption 
    that the terminal distribution of the forward process is approximately the same as the initial distribution of the backward process 
    (cf.\ in \cref{remark:setting0:eq2} in \cref{remark:setting0})
    is satisfied in \cref{setting1} when $\tal_T \approx 0$.
    Intuitively speaking, in \cref{setting1} we think in this situation that the forward process gradually adds Gaussian noise to its initial value
     until it arrives at a standard normal distribution.
\end{remark}

\subsubsection{Conditional distributions going backwards}

In this section we show that the conditional distribution of any time step of the forward process given the next value of the forward process and the initial value of the forward process is again given by a certain Gaussian distribution.
The considered conditional distributions are precisely the ones appearing in the upper bound in \cref{lemma:upperboundE}.

\begin{lemma}[Backward transition density of the forward process given the initial value]\label{lemma:qgaussian}
    Assume \cref{setting1}, let  $ \tal_1, \ldots, \tal_T, \tbe_2,\tbe_3, \ldots, \tbe_T \in (0,1)$,
    assume for all $t \in \{1, \ldots, T\}$ that $\tal_t= \textstyle\prod_ {s=1}^t\al_s$, 
    assume for all $t \in \{2,3, \ldots, T\}$ that $\tbe_t=\left[\frac{1-{\tal}_{t-1}}{1-{\tal}_t}\right](1-\al_t)$,
        and for every $t \in \{2,3,\ldots,T\}$ let $\tilde{\mu}_t \colon \R^{\di}  \times \R^{\di} \to \R^{\di}$ satisfy for all $x,y \in \R^{\di}$ that 
\begin{equation}\label{setting11:tilde_mu}
    \tilde{\mu}_t(x,y)=\left[\frac{\sqrt{\al_t}(1- {\tal}_{t-1})}{1-{\tal}_t}\right] x + \left[\frac{\sqrt{{\tal}_{t-1}}(1-\al_t) }{1-{\tal}_t}\right] y.
\end{equation}
    Then it holds for all $t \in \{2,3,\ldots,T\}$, $x_0,x_{t-1},x_t \in \R^{\di}$ that
 \begin{equation}
     \pcond{\emptyset}{t-1}{t,0}\argpcond{(x_{t-1}}{x_t,x_0)}=\n(x_{t-1}, \tilde{\mu}_{t}( x_{t},x_0), \tbe_t \mathbb{I}).
 \end{equation}
\end{lemma}
\begin{cproof}{lemma:qgaussian}
    \Nobs that \cref{setting0:pcond}, %
    \cref{setting1:prob}, and \cref{lemma:prodq}
    imply that for all $t \in \{2,3,\ldots,T\}$, $x_0,x_{t-1},x_t \in \R^{\di}$ it holds that
    \begin{equation}\label{lemma:qgaussian:bayes}
    \begin{split}
        &\pcond{\emptyset}{t-1}{t,0}\argpcond{(x_{t-1}}{x_t,x_0)}=
        \pcond{\emptyset}{t}{t-1,0}\argpcond{(x_{t}}{x_{t-1},x_0)}
        \frac{\pcond{\emptyset}{t-1}{0}\argpcond{(x_{t-1}}{x_0)}}{\pcond{\emptyset}{t}{0}\argpcond{(x_{t}}{x_0)}} \\
        &=\frac{\n(x_t, \sqrt{\al_t} x_{t-1}, (1-\al_t) \mathbb{I})\n(x_{t-1},\sqrt{\tal_{t-1}} x_{0}, (1-\tal_{t-1}) \mathbb{I})}{\n(x_{t},\sqrt{\tal_t} x_{0}, (1-\tal_t) \mathbb{I})}.
      \end{split}
    \end{equation}
    This and \cref{lemma:prob_bayes} %
demonstrate that for all $t \in \{2,3,\ldots,T\}$, $x_0,x_{t-1},x_t \in \R^{\di}$ it holds that
 \begin{equation}
 \begin{split}
     &\pcond{\emptyset}{t-1}{t,0}\argpcond{(x_{t-1}}{x_t,x_0)}=
     \n\bigg(x_{t-1}, \bigg[\big((1-\tal_{t-1}) \mathbb{I} \big) \big(\sqrt{\al_t}  \mathbb{I}\big) \Big(\big(\sqrt{\al_t}  \mathbb{I}\big) \big((1-\tal_{t-1}) \mathbb{I}\big) \big(\sqrt{\al_t}  \mathbb{I}\big) \\
     & \quad + (1-\al_t) \mathbb{I}\Big)^{-1}\big(x_{t-1}-\sqrt{\al_t}  \mathbb{I}(\sqrt{\tal_{t-1}} x_{0})\big) + \sqrt{\tal_{t-1}} x_{0}\bigg], \bigg[ (1-\tal_{t-1}) \mathbb{I}- \big((1-\tal_{t-1}) \mathbb{I}\big) \\
     & \quad \big(\sqrt{\al_t}  \mathbb{I}\big) \Big(\big(\sqrt{\al_t}  \mathbb{I}\big) \big((1-\tal_{t-1}) \mathbb{I}\big) \big(\sqrt{\al_t}  \mathbb{I}\big) + (1-\al_t) \mathbb{I}\Big)^{-1} \big(\sqrt{\al_t} \mathbb{I}\big) \big((1-\tal_{t-1}) \mathbb{I}\big)\bigg]\bigg)    \\
     & =\n\bigg(x_{t-1}, \bigg[\sqrt{\al_t} (1-\tal_{t-1})(1-\tal_{t})^{-1} \big(x_{t-1}-\sqrt{\tal_t} x_{0}\big) + \sqrt{\tal_{t-1}} x_{0}\bigg] ,\\
     & \quad \bigg[(1-\tal_{t-1}) \mathbb{I} - \al_t (1-\tal_{t-1})^2(1-\tal_{t})^{-1}  \mathbb{I}\bigg] \bigg)\\
     & = \n\bigg(x_{t-1},\bigg[\sqrt{\al_t} (1-\tal_{t-1})(1-\tal_{t})^{-1}x_{t-1} + \sqrt{\tal_{t-1}}(1-\al_{t})(1-\tal_{t})^{-1} x_{0}\bigg], \\
     & \quad \bigg[(1-\tal_{t-1})(1-\al_{t})(1-\tal_{t})^{-1}  \mathbb{I}\bigg] \bigg)\\
     & = \n(x_{t-1}, \tilde{\mu}_{t}( x_{t},x_0), \tbe_t \mathbb{I}).
 \end{split}
 \end{equation}
\end{cproof}

\subsection{Reformulated training objective in DDPMs with Gaussian noise}
\label{sec:Obj_gaussian}

The goal in this section is to choose suitable functions 
$(\mu^\theta)_{\theta \in \R^{\fd}}$ and $(\Sigma^\theta)_{\theta \in \R^{\fd}}$ in \cref{setting1}
such that the upper bound for the training objective in \cref{lemma:upperboundE} admits a convenient expression which can be used for the training of the backward process. 
The resulting upper bound is presented in \cref{prop:ddpm:noise}.

We first show in \cref{lemma:mean,lemma:mean2} below that choosing suitable variances 
$(\Sigma^\theta)_{\theta \in \R^{\fd}}$
which do not depend on the parameter $\theta  \in \R^{\fd}$ simplifies the trainable terms in the upper bound in \cref{lemma:upperboundE}.

\cfclear
\begin{lemma}[KL divergence between desired and approximated backward distribution]
    \label{lemma:mean}
Assume \cref{setting1}, 
let  $ \tal_1, \ldots, \tal_T, \tbe_2,\tbe_3, \ldots, \tbe_T \in (0,1)$,
    assume for all $t \in \{1, \ldots, T\}$ that $\tal_t= \textstyle\prod_ {s=1}^t\al_s$, 
    assume for all $t \in \{2,3, \ldots, T\}$ that $\tbe_t=\left[\frac{1-{\tal}_{t-1}}{1-{\tal}_t}\right](1-\al_t)$,
for every $t \in \{2,3,\ldots,T\}$ let $\tilde{\mu}_t \colon \R^{\di}  \times \R^{\di} \to \R^{\di}$ satisfy for all $x,y \in \R^{\di}$ that 
\begin{equation}
    \tilde{\mu}_t(x,y)=\left[\frac{\sqrt{\al_t}(1- {\tal}_{t-1})}{1-{\tal}_t}\right] x + \left[\frac{\sqrt{{\tal}_{t-1}}(1-\al_t) }{1-{\tal}_t}\right] y,
\end{equation} and assume for all $\theta \in \R^\fd$, $t \in \{2,3,\ldots,T\}$, $x_t\in \R^\di$ that $\Sigma^{\theta}_t( x_{t})=\tbe_t \mathbb{I}$. Then it holds for all $\theta \in \R^\fd$, $t \in \{2,3,\ldots,T\}$, $x_0,x_t \in \R^{\di}$ that
\begin{equation}\label{lemma:mean:thesis}
    \infdiv{\pcond{\emptyset}{t-1}{t,0}\argpcond{(\cdot}{x_t,x_0)}}{\pcond{\theta}{t-1}{t}\argpcond{(\cdot}{x_{t})}}=\frac{1}{2\tbe_t} \norm{\tilde{\mu}_{t}(x_t,x_0)-\mu^{\theta}_t(x_t)}_2^{2}
\end{equation}
\cfout.
\end{lemma}
\begin{cproof}{lemma:mean}
    \cfconsiderloaded{def:gaussian}
    \Nobs that \cref{setting1:prob}, \cref{lemma_kl_gaussian}, and \cref{lemma:qgaussian} assure that for all $\theta \in \R^\fd$, $t \in \{2,3,\ldots,T\}$, $x_0,x_t \in \R^{\di}$ it holds that
    \begin{equation}
    \begin{split}
        &\infdiv{\pcond{\emptyset}{t-1}{t,0}\argpcond{(\cdot}{x_t,x_0)}}{\pcond{\theta}{t-1}{t}\argpcond{(\cdot}{x_{t})}}  = \infdiv{\n(\cdot, \tilde{\mu}_{t}( x_{t},x_0), \tbe_t \mathbb{I})}{
        \n(\cdot, \mu^{\theta}_t( x_{t}),\tbe_t \mathbb{I}) 
        } \\
        & =\frac{1}{2}\bigg[\di\ln\bigg( \frac{\tbe_t}{\tbe_t }\bigg) - \di + \di (\tbe_t^{-1}\tbe_t)
        + \big(\mu^{\theta}_t(x_t)-\tilde{\mu}_{t}(x_t,x_0)\big)^* \tbe_t^{-1}\mathbb{I} \big(\mu^{\theta}_t(x_t)-\tilde{\mu}_{t}(x_t,x_0)\big)\bigg] \\
        &  = \frac{1}{2\tbe_t} \norm{\tilde{\mu}_{t}(x_t,x_0)-\mu^{\theta}_t(x_t)}_2^{2}
    \end{split}
    \end{equation}
    \cfload.
\end{cproof}

\cfclear
\begin{lemma}\label{lemma:mean2}
    Assume \cref{setting1},
    let  $  \tbe_1 \in (0,1)$,
    and assume for all $\theta \in \R^\fd$, $x_1 \in \R^\di$ that $\Sigma^{\theta}_1( x_{1})=\tbe_1 \mathbb{I}$. Then it holds for all $\theta \in \R^\fd$, $x_0,x_1 \in \R^{\di}$ that
\begin{equation}\label{lemma:mean2:thesis}
   -\ln\big(\pcond{\theta}{0}{1}\argpcond{(x_0}{x_{1})}\big)=\frac{\di}2\ln\big(2\pi \tbe_1 \big) + \frac{1}{2\tbe_1} \norm{x_0 - \mu^{\theta}_1(x_1)}_2^{2}.
\end{equation}
\end{lemma}
\begin{cproof}{lemma:mean2}
    \cfconsiderloaded{def:gaussian}
    \Nobs that \cref{setting1:prob1} demonstrates that for all $\theta \in \R^\fd$, $x_0,x_1 \in \R^{\di}$ it holds that
    \begin{equation}
        \begin{split}
            & -\ln\big(\pcond{\theta}{0}{1}\argpcond{(x_0}{x_{1})}\big)=-
            \ln
            \big(
            \n(x_0, \mu^{\theta}_1(x_1),  \tbe_1 \mathbb{I})
            \big) \\
            &= -\ln\Big((2\pi \tbe_1)^{-\frac{\di}2}  \exp \bigl(- \tfrac{1}{2} (x_0-\mu^{\theta}_1(x_1))^{*} (\tbe_1 \mathbb{I})^{-1} (x_0-\mu^{\theta}_1(x_1))  \bigr) \Big)\\
            &= \frac{\di}2\ln\big(2\pi \tbe_1 \big) + \frac{1}{2\tbe_1} \norm{x_0 - \mu^{\theta}_1(x_1)}_2^{2}.
        \end{split}
    \end{equation}
    \cfload
\end{cproof}

Motivated by \cref{lemma:mean,lemma:mean2} we now choose a specific form for the means 
$(\mu^\theta)_{\theta \in \R^{\fd}}$ in \cref{setting1} allowing the terms in  \cref{lemma:mean,lemma:mean2} (respectively in the upper bound in \cref{lemma:upperboundE}) to be further simplified.

\cfclear
\cfconsiderloaded{def:gaussian}
\begin{lemma}[KL divergence between desired and approximated backward distribution]\label{lemma:ddpm:noise}
    Assume \cref{setting1},
    let  $ \tal_0,\tal_1, \ldots, \tal_T, \tbe_1,\ldots, \tbe_T \in (0,1)$
    satisfy for all $t \in \{1, \ldots, T\}$ that
        $\tal_t= \textstyle\prod_ {s=1}^t\al_s$ and $\tbe_t=\left[\frac{1-{\tal}_{t-1}}{1-{\tal}_t}\right](1-\al_t)$,
    for every $\theta \in \R^\fd$ let $\bV^{\theta} \colon \R^{\di} \times \R \to \R^{\di}$ be measurable, and assume for all $\theta \in \R^\fd$, $t \in \{1,\ldots,T\}$, $x_t \in \R^{\di}$ that
        \begin{equation}\label{lemma:ddpm:noise:mu_theta}
    \mu^{\theta}_t(x_t) = \frac{1}{\sqrt{\al_t}}\left(x_t - \frac{1 - \al_t}{\sqrt{1 - {\tal}_t}}\bV^{\theta}(x_t, t)\right) \qandq \Sigma^{\theta}_t( x_{t})=\tbe_t \mathbb{I}.
        \end{equation}
Then 
 \begin{enumerate}[label=(\roman*)]
        \item \label{lemma:ddpm:noise:item1}
        it holds for all $\theta \in \R^\fd$, $x_0, x_1, \varepsilon_1 \in \R^\di$ with $x_1= \sqrt{\tal_1}x_0 + \sqrt{1-\tal_1} \varepsilon_1$ that
        \begin{equation}
            -\ln\big(\pcond{\theta}{0}{1}\argpcond{(x_0}{x_{1})}\big)=\frac{\di}2\ln\big(2\pi \tbe_1 \big) + \frac{1}{2\tbe_1}\frac{(1-\al_1)^2}{ (1-\tal_1)\al_1} \lVert \varepsilon_1 - \bV^{\theta}(x_1, 1)\rVert_2^{2}
        \end{equation}
        and
        \item \label{lemma:ddpm:noise:item2}
        it holds for all $\theta \in \R^\fd$, $t \in \{2,3,\ldots,T\}$, $x_0, x_t, \varepsilon_t \in \R^\di$ with $x_t= \sqrt{\tal_t}x_0 + \sqrt{1-\tal_t} \varepsilon_t$ that
            \begin{equation}\label{lemma:ddpm:noise:eq}
            \begin{split}
                 \infdiv{\pcond{\emptyset}{t-1}{t,0}\argpcond{(\cdot}{x_t,x_0)}}{\pcond{\theta}{t-1}{t}\argpcond{(\cdot}{x_{t})}} = \frac{1}{2\tbe_t}\frac{(1 - \al_t)^2}{(1 - {\tal}_t)\al_t} \norm{\varepsilon_t - \bV^{\theta} (x_t, t)}_2^2
            \end{split}
            \end{equation}
 \end{enumerate}
\cfout.
\end{lemma}
\begin{cproof}{lemma:ddpm:noise}
\Nobs that  \cref{lemma:ddpm:noise:mu_theta} and \cref{lemma:mean2} ensure that for all $\theta \in \R^\fd$,  $x_0, x_1, \varepsilon_1 \in \R^\di$ with $x_1= \sqrt{\tal_1}x_0 + \sqrt{1-\tal_1} \varepsilon_1$
it holds that
\begin{equation}
\begin{split}
    &-\ln\big(\pcond{\theta}{0}{1}\argpcond{(x_0}{x_{1})}\big)=\frac{\di}2\ln\big(2\pi \tbe_1 \big) + \frac{1}{2\tbe_1} \norm{x_0 - \mu^{\theta}_1(x_1)}_2^{2}\\
    & =\frac{\di}2\ln\big(2\pi \tbe_1 \big) + \frac{1}{2\tbe_1} \Big\lVert x_0 - \frac{1}{\sqrt{\al_1}}\left(x_1 - \sqrt{1 - {\al}_1}\bV^{\theta}(x_1, 1)\right)\Big\rVert_2^{2}\\
     & =\frac{\di}2\ln\big(2\pi \tbe_1 \big) + \frac{(1-\al_1)}{2\tbe_1 \al_1} \Big\lVert \frac{\sqrt{\al_1}}{\sqrt{1 - {\al}_1}} x_0 - \frac{1}{\sqrt{1 - {\al}_1}} x_1 + \bV^{\theta}(x_1, 1)\Big\rVert_2^{2}\\
     & =\frac{\di}2\ln\big(2\pi \tbe_1 \big) + \frac{(1-\al_1)}{2\tbe_1 \al_1} \lVert \varepsilon_1 - \bV^{\theta}(x_1, 1)\rVert_2^{2}\\
     & =\frac{\di}2\ln\big(2\pi \tbe_1 \big) + \frac{(1-\al_1)^2}{2\tbe_1 (1-\tal_1)\tal_1} \lVert \varepsilon_1 - \bV^{\theta}(x_1, 1)\rVert_2^{2}.
\end{split}
\end{equation}
This establishes \cref{lemma:ddpm:noise:item1}.
Throughout this proof for every $t \in \{0,1,\ldots,T\}$
let $\tilde{\mu}_t \colon \R^{\di}  \times \R^{\di} \to \R^{\di}$ satisfy for all $x,y \in \R^{\di}$ that
    \begin{equation}\label{lemma:ddpm:noise:mu}
    \tilde{\mu}_t(x,y)= \left[\frac{\sqrt{\al_t}(1- {\tal}_{t-1})}{1-{\tal}_t}\right] x + \left[\frac{\sqrt{{\tal}_{t-1}}(1-\al_t) }{1-{\tal}_t}\right] y.
\end{equation}
\Nobs that \cref{lemma:ddpm:noise:mu_theta}, \cref{lemma:ddpm:noise:mu}, and $\cref{lemma:mean} $ show that for all $\theta \in \R^\fd$, $t \in \{2,3,\ldots,T\}$, $x_0, x_t, \varepsilon_t \in \R^\di$ with $x_t= \sqrt{\tal_t}x_0 + \sqrt{1-\tal_t} \varepsilon_t$
it holds that
    \begin{equation}
    \begin{split}
         &\infdiv{\pcond{\emptyset}{t-1}{t,0}\argpcond{(\cdot}{x_t,x_0)}}{\pcond{\theta}{t-1}{t}\argpcond{(\cdot}{x_{t})}} = \frac{1}{2\tbe_t} \norm{\tilde{\mu}_{t}(x_t,x_0)-\mu^{\theta}_t(x_t)}_2^{2} \\
         & =\frac{1}{2\tbe_t}\left\|  \frac{\sqrt{\al_t}(1- {\tal}_{t-1})}{1-{\tal}_t} x_t + \frac{\sqrt{{\tal}_{t-1}}(1-\al_t) }{1-{\tal}_t} x_0 
         -\frac{1}{\sqrt{\al_t}}x_t + \frac{1 - \al_t}{\sqrt{1 - {\tal}_t}\sqrt{\al_t}}\bV^{\theta}(x_t, t)
         \right\|_2^{2}\\ 
         & = \frac{1}{2\tbe_t} \left\| \frac{\al_t(1- {\tal}_{t-1})- 1+{\tal}_t}{(1-{\tal}_t)\sqrt{\al_t}}x_t + \frac{\sqrt{{\tal}_{t-1}}(1-\al_t) }{(1-{\tal}_t)\sqrt{{\tal}_{t}}} (x_t - \sqrt{1-{\tal}_t}\varepsilon_t) \right.  \\ 
         & \quad  \left. + \frac{1 - \al_t}{\sqrt{1 - {\tal}_t}\sqrt{\al_t}}\bV^{\theta}(x_t, t) \right\|_2^{2}  \\
         &= \frac{1}{2\tbe_t}\left\| - \frac{1 - \al_t}{\sqrt{1 - {\tal}_t}\sqrt{\al_t}}   \varepsilon_t + \frac{1 - \al_t}{\sqrt{1 - {\tal}_t}\sqrt{\al_t}}\bV^{\theta}(x_t, t)\right\|_2^{2}  \\
         &=\frac{(1 - \al_t)^2}{2\tbe_t(1 - {\tal}_t)\al_t} \norm{\varepsilon_t - \bV^{\theta} (x_t, t)}_2^2
    \end{split}
    \end{equation}
    \cfload.
    This demonstrates \cref{lemma:ddpm:noise:item2}.
\end{cproof}

Using the choices for
$(\mu^\theta)_{\theta \in \R^{\fd}}$ and $(\Sigma^\theta)_{\theta \in \R^{\fd}}$ 
in \cref{setting1}
elaborated in \cref{lemma:ddpm:noise}, we now present in \cref{prop:ddpm:noise} below the resulting reformulation for the upper bound in \cref{lemma:upperboundE}.
In addition, we also add two items in \cref{prop:ddpm:noise} which illustrate how to sample from the forward and backward processes, so that the result provides a complete theoretical basis for the scheme described in \cref{setting:dnn3}.

\cfclear
\cfconsiderloaded{def:gaussian}
\begin{prop}[Reformulation of the upper bound for the \ENLL]
    \label{prop:ddpm:noise}
    Assume \cref{setting1},
    let  $ \tal_0,\tal_1, \ldots, \tal_T, \tbe_1,\ldots, \tbe_T \in (0,1)$
    satisfy for all $t \in \{1, \ldots, T\}$ that
        $\tal_t= \textstyle\prod_ {s=1}^t\al_s$ and $\tbe_t=\left[\frac{1-{\tal}_{t-1}}{1-{\tal}_t}\right](1-\al_t)$,
    for every $t \in \{0,1,\ldots,T\}$ let $\mathcal{E}_t \colon \Omega \to \R^{ \di }$  %
    satisfy
    $X_t^\emptyset = \sqrt{\tal_t}X_0^\emptyset + \sqrt{1-\tal_t} \mathcal{E}_t$,
    for every $\theta \in \R^\fd$ let $\bV^{\theta} \colon \R^{\di} \times \R \to \R^{\di}$ be measurable, and assume for all $\theta \in \R^\fd$, $t \in \{1,\ldots,T\}$, $x_t \in \R^{\di}$ that
\begin{equation}\label{prop:ddpm:noise:mu_theta}
    \mu^{\theta}_t(x_t) = \frac{1}{\sqrt{\al_t}}\left(x_t - \frac{1 - \al_t}{\sqrt{1 - {\tal}_t}}\bV^{\theta}(x_t, t)\right) \qandq \Sigma^{\theta}_t( x_{t})=\tbe_t \mathbb{I}.
\end{equation}
Then 
\begin{enumerate} [label=(\roman*)]
    \item \label{prop:ddpm:noise:item1}
    it holds for all $\theta \in \R^\fd$
that
    \begin{equation}\label{prop:ddpm:noise:eq}
    \begin{split}
        \negloglike[\big]{\p{\emptyset}{0}}{\p{\theta}{0}}
        &=
         \Eb{}{- \ln (\p{\theta}{0}(X_{0}^\emptyset)) }\\
         &\leq   \Eb{} {\infdiv{\pcond{\emptyset}{T}{0}\argpcond{(\cdot}{X_0^\emptyset)}}{\Pi}} + \frac{\di}2\ln\big(2\pi \tbe_1 \big) \\ %
        &\quad + \sum_{t = 1}^T \frac{1}{2\tbe_t}\frac{(1 - \al_t)^2}{(1 - {\tal}_t)\al_t} \Eb{}{\norm{\mathcal{E}_t - \bV^{\theta} (\sqrt{\tal_t}X_0^\emptyset + \sqrt{1-\tal_t} \mathcal{E}_t , t)}_2^2},
    \end{split}
    \end{equation}
    \item \label{prop:ddpm:noise:item2}
        it holds for all 
            $t \in \{1,\ldots,T\}$,
            $B \in \cB(\R^{\di})$ that $\mathcal{E}_t$ and $X_0^\emptyset$ are independent and
            $\Prob{}(\mathcal{E}_t \in B)= \int_B \n(x,0,\mathbb{I})\, \d x$, and 
    \item \label{prop:ddpm:noise:item3}
     for all $\theta \in \R^\fd$ there exist i.i.d.\ random variables $Z^{\theta}_t \colon \Omega \to \R^\di$, $t \in \{1,\ldots,T+1\}$, such that for all $t \in \{1,\ldots,T\}$, $B \in \cB(\R^\di)$ it holds that 
    \begin{equation}
        \Prob{}(Z^{\theta}_1 \in B)= \int_B \n(z,0,\mathbb{I})\, \d z, \quad \Xt{\theta}_{T}=Z^{\theta}_{T+1},
        \qand
    \end{equation}
    \begin{equation} 
    \Xt{\theta}_{t-1}= \frac1{\sqrt{\al_t}}\bigg(\Xt{\theta}_{t}-\frac{1-\al_t}{\sqrt{1-\tal_t}}\bV^{\theta}(\Xt{\theta}_{t}, t)\bigg) +  \sqrt{\tbe_t} Z_t^{\theta}
\end{equation}
\end{enumerate}
\cfout.
\end{prop}
\begin{cproof}{prop:ddpm:noise}
    \Nobs that \cref{lemma:upperboundE}, \cref{lemma:ddpm:noise}, and the fact that for all $t \in \{1,\ldots,T\}$ it holds that $ X_t^\emptyset = \sqrt{\tal_t}X_0^\emptyset + \sqrt{1-\tal_t} \mathcal{E}_t$ demonstrate that for all $\theta \in \R^\fd$
    it holds that
    \begin{equation}
        \begin{split}
         &\negloglike[\big]{\p{\emptyset}{0}}{\p{\theta}{0}} 
         =
         \Eb{}{- \ln \big(\p{\theta}{0}(X_{0}^\emptyset)\big) } \\
         &\leq
        \Eb{} {\infdiv{\pcond{\emptyset}{T}{0}\argpcond{(\cdot}{X_0^\emptyset)}}{\Pi}} -\Eb{}{\ln \big( \pcond{\theta}{0}{1}\argpcond{(X_0^\emptyset}{X_1^\emptyset)}\big)}\\
        &\quad + \sum_{t = 2}^T\Eb{}{\infdiv{\pcond{\emptyset}{t-1}{t,0}\argpcond{(\cdot}{X_t^\emptyset,X_0^\emptyset)}}{\pcond{\theta}{t-1}{t}\argpcond{(\cdot}{X_{t}^\emptyset)}}}\\
        &=
        \Eb{} {\infdiv{\pcond{\emptyset}{T}{0}\argpcond{(\cdot}{X_0^\emptyset)}}{\Pi}} + \frac{\di}2\ln\big(2\pi \tbe_1 \big) + \frac{1}{2\tbe_1}\frac{(1-\al_1)^2}{ (1-\tal_1)\al_1} \Eb{}{\lVert \varepsilon_1 - \bV^{\theta}(x_1, 1)\rVert_2^{2}}\\
        &\quad + \sum_{t = 2}^T \frac{1}{2\tbe_t}\frac{(1 - \al_t)^2}{(1 - {\tal}_t)\al_t} \Eb{}{\norm{\mathcal{E}_t - \bV^{\theta} (X_t^\emptyset, t)}_2^2}\\
         &=
        \Eb{} {\infdiv{\pcond{\emptyset}{T}{0}\argpcond{(\cdot}{X_0^\emptyset)}}{\Pi}}  + \frac{\di}2\ln\big(2\pi \tbe_1 \big)\\
        &\quad + \sum_{t = 1}^T \frac{1}{2\tbe_t}\frac{(1 - \al_t)^2}{(1 - {\tal}_t)\al_t} \Eb{}{\norm{\mathcal{E}_t - \bV^{\theta} (\sqrt{\tal_t}X_0^\emptyset + \sqrt{1-\tal_t} \mathcal{E}_t , t)}_2^2}
    \end{split}
    \end{equation}
    \cfload.
    This establishes \cref{prop:ddpm:noise:item1}. \Moreover \cref{cor:randomvar:prodq} demonstrates \cref{prop:ddpm:noise:item2}. \Moreover \cref{lemma:constructive} and \cref{prop:ddpm:noise:mu_theta} show \cref{prop:ddpm:noise:item3}.
\end{cproof}

\begin{remark}[Explanations for \cref{prop:ddpm:noise}]
\label{remark:ddpm:noise}
    In this remark we provide some interpretations for the mathematical objects appearing in \cref{prop:ddpm:noise} and discuss some intuitive consequences of \cref{prop:ddpm:noise} for the training of the backward process.
    
    In \cref{prop:ddpm:noise} we specify the terms 
    $(\mu^{\theta})_{\theta \in \R^{\fd}}$ and $(\Sigma^{\theta})_{\theta \in \R^{\fd}}$
    in \cref{setting1} 
    such that the upper bound in \cref{lemma:upperboundE} for the training objective  
    $
    (
        \R^{\fd}
        \ni 
        \theta 
        \mapsto
        \negloglike[\big]{\p{\emptyset}{0}}{\p{\theta}{0}}
        \in 
        \R
    )
    $
    admits a convenient expression involving the cumulative noise $(\mathcal{E}_t)_{t \in \{1,\ldots,T\}}$ added to the initial value in the forward process.
    We think of the function $(\bV^{\theta})_{\theta \in \R^{\fd}}$ appearing in the definition of $(\mu^{\theta})_{\theta \in \R^{\fd}}$ as a denoising \ANN.
    Roughly speaking, minimizing terms in the upper bound in \cref{prop:ddpm:noise:item1} in \cref{prop:ddpm:noise} should result in \ANN\ parameters 
    $\opttheta \in \R^{\fd}$ such that for all
    $t \in \{1, 2, \ldots, T\}$ we have that 
    \begin{equation}
    \label{T_B_D}
    \begin{split}
        \textstyle
        \bV^{\opttheta} (X_t^{\emptyset} , t) 
    =
        \bV^{\opttheta} (\sqrt{\tal_t}X_0^\emptyset + \sqrt{1-\tal_t} \mathcal{E}_t , t) 
    \approx
       \mathcal{E}_t.
    \end{split}
    \end{equation}
    This can be interpreted as the \ANN\ $(\bV^{\theta})_{\theta \in \R^{\fd}}$ learning to extract the noise component
    $\mathcal{E}_t$ from the noisy data $X_t^{\emptyset}$
    of the forward process for all time steps $t \in \{1, 2, \ldots, T\}$.

    \Cref{prop:ddpm:noise:item2,prop:ddpm:noise:item3} in \cref{prop:ddpm:noise} show how the forward and backward processes can be sampled using independent standard normal random variables.

    We note that in \cref{prop:ddpm:noise}
    the number $\tal_0 \in (0,1)$ and $\tbe_1 \in (0,1)$ are not given as functions of $\al_1,\al_2,\ldots,\al_T$.
    The natural choice for $\tal_0$ would be 
    \begin{equation}
        \tal_0 = \prod_{s=1}^0 \al_s = 1
    \end{equation}
    and the corresponding choice for $\tbe_1$ would be
    \begin{equation}
        \tbe_1 = \left[\frac{1-{\tal}_{0}}{1-{\tal}_1}\right](1-\al_1) = 0.
    \end{equation}
    This would, however, not be admissible since the density of the normal distribution is not defined for zero variance
    and the bound in \cref{prop:ddpm:noise:item1} would involve a division by zero.
    Nonetheless, in \cref{setting:dnn3} below we will act as if we can choose $\tal_0 = 1$ and $\tbe_1 = 0$ as this does result in a practical and effective scheme.
\end{remark}

\subsection{DDPM generative method with Gaussian noise}
\label{sec:scheme_gaussian}

We now formulate a generative method for \DDPMs\ with Gaussian noise which is based on the upper bound for the training objective in \cref{prop:ddpm:noise}.
This scheme was proposed in \cite{ho2020denoising}.

\begin{method}[\DDPM\ generative method with Gaussian noise]
\label{setting:dnn3}
Let $\di,\fd, \ds \in \N$, $T\in \N \backslash \{1\}$,  $\gamma \in (0, \infty)$, $\al_1,\ldots,\al_T \in (0,1)$, $ \tal_0, \tal_1, \ldots, \tal_T, \tbe_1,\ldots, \tbe_T \in [0,1]$, assume for all $t \in \{0,1, \ldots, T\}$ that $\tal_t= \textstyle\prod_{s=1}^t\al_s$,  assume for all $t \in \{1, \ldots, T\}$ that $\tbe_t=\left[\frac{1-{\tal}_{t-1}}{1-{\tal}_t}\right](1-\al_t)$,
for every $\theta \in \R^{\fd}$ let 
$\bV^\theta \colon \R^{\di} \times \{1,\ldots,T\} \to\R^{\di}$
be a function,
let  
$\fL \colon \R^\fd \times \R^\di \times \R^\di \times \{1,\ldots,T\} \to \R$
satisfy 
for all $ \theta \in \R^{ \fd } $, $x, \varepsilon \in \R^\di$, $t \in \{1,\ldots,T\}$
that
\begin{equation}\label{setting:dnn3:loss}
\textstyle  \fL( \theta, x, \varepsilon, t ) =  \bigl\| \varepsilon - \bV^{ \theta }\bigl( \sqrt{\tal_t} x + \sqrt{1-\tal_t} \varepsilon
, t \bigr)  \bigr\|^2,
\end{equation}
let $\fG \colon \R^\fd \times \R^\di \times \R^\di \times \{1,\ldots,T\} \to \R^{ \fd }$ satisfy for all 
$x, \varepsilon \in \R^\di$,
$t \in \{1,\ldots,T\}$,
$\theta \in \R^\fd$ with $\fL( \cdot,x, \varepsilon, t)$ differentiable at $\theta$
that
\begin{equation}
\fG( \theta, x, \varepsilon, t) =  ( \nabla_{ \theta } \fL )( \theta, x, \varepsilon, t),
\end{equation}
let $(\Omega, \cF, \mathbbm{P})$ be a probability space, 
let
$\datax_{n,i} \colon \Omega \to \R^\di$, $ n, i \in \N$, be random variables,
let $\mathcal{E}_{n,i} \colon \Omega \to \R^{ \di }$, $ n, i \in \N$, 
be i.i.d.\ standard normal random variables,
let
$\randT_{n} \colon \Omega \to \{1,2,\ldots,T\}$, $ n\in \N$,
be independent $\mathcal{U}_{\{1,2,\ldots,T\}}$-distributed random variables,
let 
$ \Theta  \colon \N_0 \times \Omega \to \R^{ \fd }
$
be a stochastic process which satisfies 
for all $n\in\N$ that
\begin{equation}\label{setting:dnn3:sgd}
\Theta_{n}=\Theta_{n-1}-\gamma \left[ \frac{ 1 }{ \ds }   \smallsum\limits_{ i= 1 }^{ \ds }
 \fG( \Theta_{n-1}, \datax_{n,i} , \mathcal{E}_{n_,i}, \randT_n ) \right],
\end{equation}
let $\nn \in \N$, let $Z_t \colon \Omega \to \R^{ \di }$, $t\in\{1,\ldots,T+1\}$,
be i.i.d.\ standard normal random variables,
let $\backX = (\backX_t)_{t \in \{0,1,\ldots,T\}} \colon \allowbreak \{0,1,\ldots,T\} \times \Omega \to \R^{ \di }$ be a stochastic process,
and assume for all $t \in \{1,\ldots,T\}$ that
\begin{equation} \label{setting:dnn3:backx}
\backX_{T} = Z_{T+1} \qandq \backX_{t-1}= \frac1{\sqrt{\al_t}}\bigg(\backX_t-\frac{1-\al_t}{\sqrt{1-\tal_t}}\bV^{\Theta_{\nn}}(\backX_t, t)\bigg) +\sqrt{\tbe_t}Z_t.
\end{equation}
\end{method}

\begin{remark}[Explanations for \cref{setting:dnn3}]
\label{remark:dnn3}
In this remark we provide some intuitive and theoretical explanations for \cref{setting:dnn3} and we roughly explain in what sense the scheme in \cref{setting:dnn3} can be used for generative modelling.

Roughly speaking, the scheme in \cref{setting:dnn3} is based on the idea to minimize the upper bound in \cref{prop:ddpm:noise}.
One major advantage of the upper bound in \cref{prop:ddpm:noise} 
compared to the one in \cref{lemma:upperboundE} 
is that the trainable terms 
in the upper bound in \cref{prop:ddpm:noise}
only depend in a straight forward way on the initial value of the forward process (\eg a random element from a training dataset)
and on a noise component 
instead of depending on whole trajectories of the forward process and on 
conditional \PDFs\ as in \cref{lemma:upperboundE}.
Specifically, 
the upper bound in \cref{prop:ddpm:noise} suggest to
train an \ANN\ to extract the noise component from the noisy data of the forward process at each time step, which is what \cref{setting:dnn3} aims to do.

In light of this,
we note that 
\begin{enumerate}[label=(\roman*)]
    \item we think of $(\bV^{\theta})_{\theta \in \R^{\fd}}$ as the \ANN\ which is trained to predict the noise component of the noisy data at each time step,
    \item we think of
    $\fL$ 
as the loss used in the training,
    \item we think of
    $\fG$ 
    as the generalized gradient of the loss $\fL$ with respect to the trainable parameters,
    \item we think of 
    $\datax_{n,i}$, $ n, i \in \N$, 
    as random samples of the initial value of the forward process used for training,
    \item we think of
    $\mathcal{E}_{n,i}$, $ n, i \in \N$,
    as the noise components of the forward process used for training,
    \item we think of 
    $\randT_{n}$, $ n\in \N$,
    as random times used to determine which terms of the upper bound are considered in each training step,
    \item we think of
    $(\Theta_{n})_{n\in\N_0}$
    as the training process for the parameters of the backward process given by an \SGD\ process for the generalized gradient $\fG$
with learning rate $\gamma$,
batch size $\ds$, and
training data
    $(\datax_{n,i}, \mathcal{E}_{n_,i}, \randT_{n})_{(n,i) \in \N^2}$,
    \item we think of
    $\nn$
    as the number of training steps,
    \item we think of
    $Z_t$, $t\in\{1,\ldots,T\}$,
    as the noise components of the backward process, and
    \item we think of
    $\backX$
    as the backward process for the trained parameters $\Theta_{\nn}$
(cf.\ \cref{prop:ddpm:noise:item3} in \cref{prop:ddpm:noise}).
\end{enumerate}
Under suitable assumptions, we expect 
the terminal value $\backX_0$ of the trained backward process to be approximately distributed according to the distribution we would like to sample from.
In other words, we think of the random variable $\backX_0$ as the generative sample produced by \cref{setting:dnn3}.

Note that the training objective that the \SGD\ process aims to minimize is given
for all 
    $\theta \in \R^{ \fd } $
by
\begin{equation}
\label{remark:dnn3:eq1}
\begin{split}
    \Eb{}{\fL(\theta,\datax_{1,1}, \mathcal{E}_{1,1}, \randT_{1})}
=
    \frac{1}{T}
    \pr*{
       \sum_{t = 1}^T 
         \Eb{}{\norm{\mathcal{E}_{1,1} - \bV^{\theta} (\sqrt{\tal_t} \datax_{1,1} + \sqrt{1-\tal_t} \mathcal{E}_{1,1} , t)}_2^2}
    }
\end{split}
\end{equation}
and does therefore not exactly correspond to the upper bound in \cref{prop:ddpm:noise}.
Specifically, the training objective in \cref{remark:dnn3:eq1} omits the weighting terms in the upper bound in \cref{prop:ddpm:noise} and adjust the term for the first step of the forward process to all other terms.
These simplifications are empirically justified in \cite[Section 3.4]{ho2020denoising}.

We note that in \cref{setting:dnn3} we have the natural choice that
\begin{equation}
    \tal_0 = 1 \qandq \tbe_1 = 0,
\end{equation}
despite this choice not being admissible in the context of \cref{prop:ddpm:noise}
(cf.\ \cref{remark:ddpm:noise}).
The fact that $\tbe_1 = 0$ implies that in the last step of the backward process in \cref{setting:dnn3:backx} we have that no noise is being added.
This makes intuitive sense as the result of the last step of the backward process is considered as the generative sample.

\end{remark}

\begin{remark}[Choice of noise intensity in \cref{setting:dnn3}] \label{remark:def:alpha}
We recall that in \cref{setting:dnn3} we have for all 
    $t \in \{1,\ldots,T\}$ 
that $(1-\al_t)$ is a measure for the amount of noise added in the $t$-th time step of the forward process (cf.\ \cref{remark:setting1}).
In \cite{ho2020denoising} the following choice for the parameters $(\al_t)_{t \in \{1,\ldots,T\}}$ in \cref{setting:dnn3}
is proposed:
Assume that $\al_1=1-10^{-4}$, $\al_T=0.98$, and
\begin{equation}\label{setting:dnn3:alpha}
    \al_t=\al_1-(t-1)\frac{\al_1-\al_T}{T-1}.
\end{equation}
The cummulative noise intensities $(\tal_t)_{t \in \{1,\ldots,T\}}$ in the case $T = 1000$ are graphically illustrated in \cref{fig:alphabar}.
Roughly speaking, this choice corresponds to adding very small amounts of noise in the initial steps of the forward process when the distribution of the forward process is still close to the distribution from which we want to sample from and adding more noise in the later steps of the forward process when the distribution of the forward process is already very noisy.

\begin{figure}[H]
\centering
{\includegraphics[width=.75\textwidth]{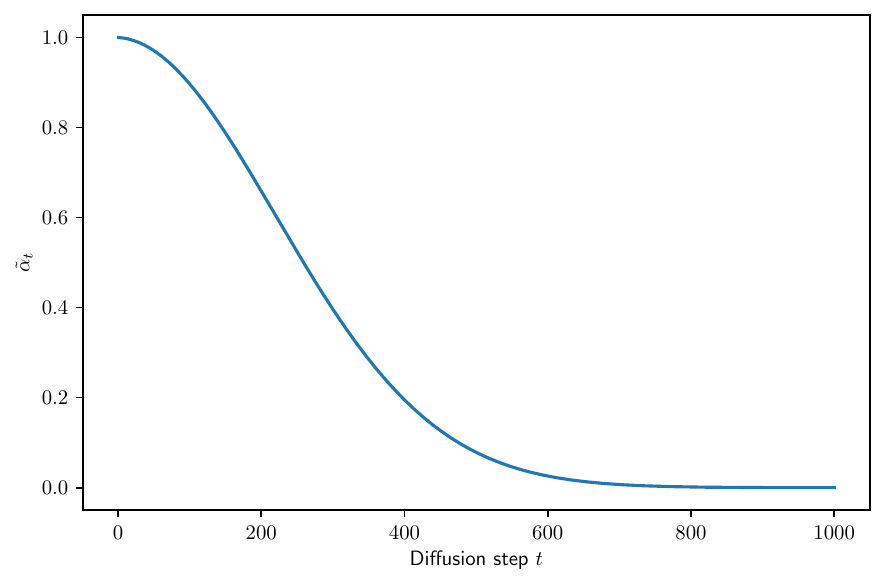}}
\caption{\label{fig:alphabar}
Graphical illustration of $(\tal_t)_{t \in \{1,\ldots,T\}}$ 
in \cref{setting:dnn3}
for $T=1000$
and $(\al_t)_{t \in \{1,\ldots,T\}}$ given as in 
\cref{setting:dnn3:alpha}.
}
\end{figure}
    
\end{remark}

\subsection{Network architectures for the backward process} 
\label{sec:architecture}
In this section we discuss the most popular choice for the architecture of the \ANN\ $(\bV^{\theta})_{\theta \in \R^{\fd}}$ from \cref{setting:dnn3}.
Specifically, we explain UNets in \cref{subsec:unet} and present how the temporal component is commonly incorporated in \cref{sec:time_embedding}.
    For general introductions to \ANN\ architectures we refer, \eg to 
        \cite[Section 9]{bach2024learning},
        \cite[Section 5]{10.5555/1162264},
        \cite[Section 1]{jentzen2023mathematical}, and
        \cite[Section 20]{shalev2014understanding}.

\subsubsection{UNets}\label{subsec:unet}

\begin{figure}[H]
    \centering
    {\includegraphics[width=1\textwidth]{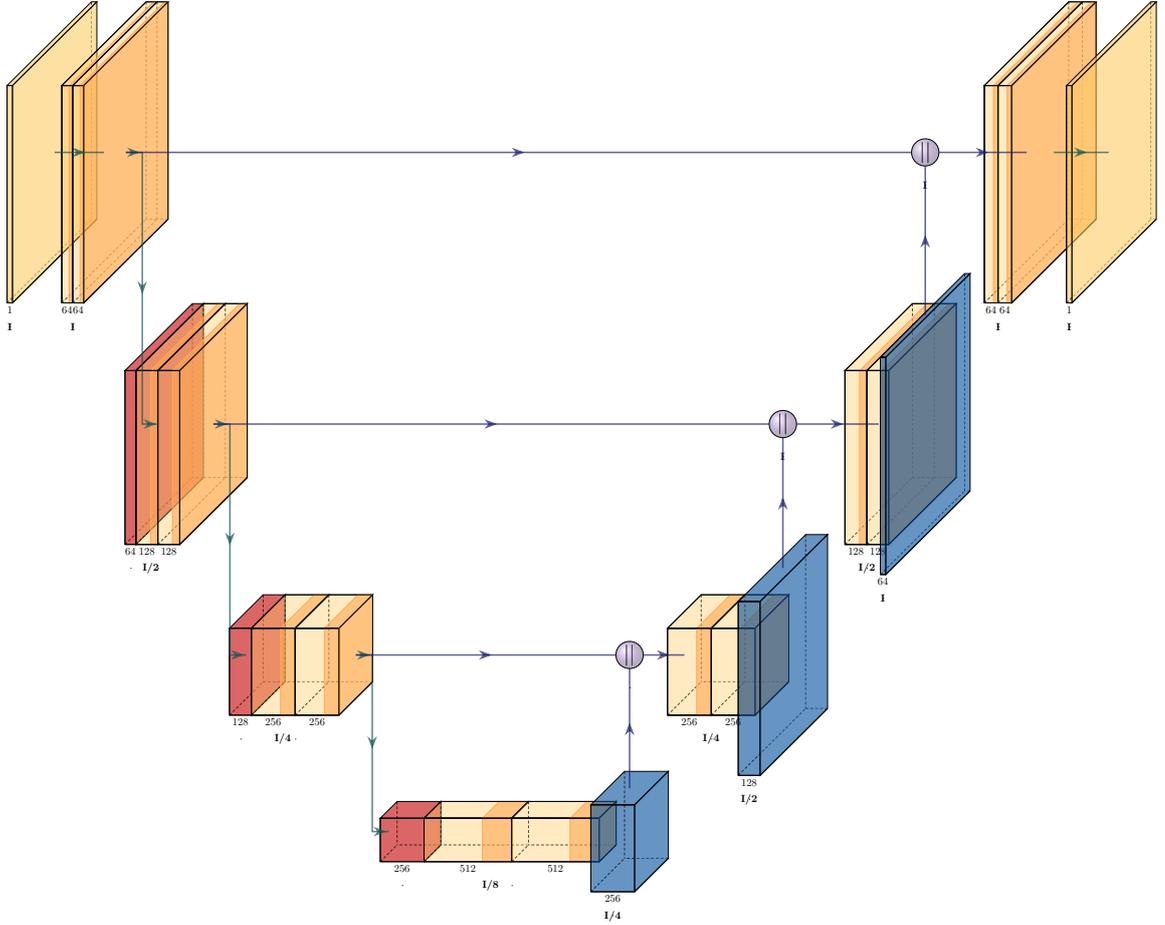}}
    \caption{Graphical illustration of a typical UNet architecture in case of two dimensional data (e.g\ images). In yellow the convolutions, in red the max pooling operations, in blue the transpose convolutions. 
    During each max pooling operation in the encoder network (left side), we increase the number of channels twofold and reduce the spatial dimensions by half. Conversely, in each transpose convolution in the decoder network (right side), we reduce the number of channels by half and double the spatial dimensions. In the decoder part we concatenate encoder's feature map with decoder's feature maps.}
    \label{image:unet}
\end{figure}

In the following we introduce the most common architecture used in diffusion models, the UNet architecture \cite{7298965}.
UNets have gained popularity in the field of computer vision, particularly for their effectiveness in semantic segmentation tasks  but it has also been applied in various other domains, see, \eg \cite{sohl2015deep, dhariwal2021diffusion, 9878449, nichol2022glide, saharia2022photorealistic}. 
Roughly speaking, UNets have an encoder-decoder structure made up of blocks.
We now provide some comments on major components and aspects of UNets. See \cref{image:unet} for a graphical illustration of its architecture.
\begin{enumerate}[label=(\roman*)]
\item The encoder network (contracting path) is responsible for diminishing the spatial dimensions and enlarging the number of channels using down-sampling operations.
It is made of blocks or levels that share the same structure and gradually compress the input. Each block typically involves convolutional layers, group or batch normalizations, and max-pooling. Optionally, before the max pooling an attention layer can be inserted. The encoder network corresponds to the left side of \cref{image:unet}.
\item At the bottom of UNets, after the encoder network, there is the bottleneck, the most compressed and abstracted form of the input's information (cf.\ bottom part of \cref{image:unet}).
\item The decoder network (expanding path), on the contrary, upsamples the  spatial information.
It is made up of blocks or levels that share the same structure and specularly mirror the one of the corresponding blocks in the encoder network.
The process also employs transposed convolutions to progressively reconstruct the original shape. Optionally, an attention layer can be inserted. The decoder network corresponds to the right side of \cref{image:unet}.
\item Skip connections have a crucial role in the model. These connections link the encoder's feature maps to the corresponding decoder's feature maps at the same spatial resolution (horizontal arrows in  \cref{image:unet}). They help the decoder to generate better features and prevent gradient degradation in the backpropagation.
\end{enumerate}

\subsubsection{Time embedding}
\label{sec:time_embedding}

 We now aim to describe how the temporal component is commonly incorporated in UNets.
The time step is a fundamental input since the model parameters are shared across time. Passing a structured temporal signal permits the model to capture at which particular time step we are operating.
The sinusoidal time embedding, defined in \cref{def:sin_time_emb}, is the embedding typically used (cf., \eg \cite{ho2020denoising, nichol2021improved, song2022denoising}),
which is inspired by positional encoding \cite{NIPS2017_3f5ee243}.
It introduces a continuous and periodic time signal, enabling the model to implicitly learn the sequence of events during the diffusion process. 

The time embeddings are typically added to the input features at various levels in the UNet architecture, particularly in the encoder and decoder paths (cf.\ \cref{subsec:unet}).

\begin{definition}[Sinusoidal time embedding]\label{def:sin_time_emb}
    Let $ d, c \in \N$ satisfy $d= 2 c $. Then we denote by  $\textnormal{TimeEmb}^{(d)}=(\textnormal{TimeEmb}_1^{(d)},\ldots,\textnormal{TimeEmb}_d^{(d)}) \colon \N \to  \R^{d}$ the function which satisfies for all
    $t\in \N$, $i \in \{1,\ldots,c\}$ that
    \begin{equation}
        \textnormal{TimeEmb}_i^{(d)}(t)= \sin \Bigg( \frac{t}{10000^{\frac{i}{c-1}}}  \Bigg) \qandq \textnormal{TimeEmb}_{c+i}^{(d)}(t)= \cos \Bigg( \frac{t}{10000^{\frac{i}{c-1}}}  \Bigg)
    \end{equation}
and we call  $\textnormal{TimeEmb}^{(d)}$ the sinusoidal time embedding with embedding dimension $d$.
\end{definition}
\begin{figure}[H]
\centering
{\includegraphics[width=.7\textwidth]{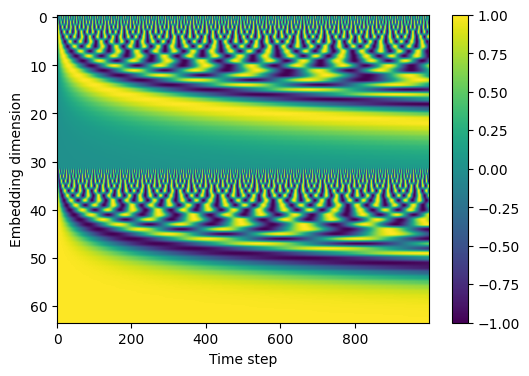}}
\caption{Sinusoidal time embedding for $1000$ time step using as embedding dimension $64$.}
\end{figure}

\section{Evaluation of generative models}
\label{sec:evaluation}
In the context of generative modelling and in particular in diffusion models, evaluating the quality and performance of generated data is essential. Therefore, finding robust evaluation metrics is crucial to ensure the models are producing the desirable outcomes. 
In this section we consider two types of metrics used for this purpose, content variant metrics and content invariant metrics. 
These metrics provide an understanding of the model's capabilities in different aspects.
In \cref{content_variant_metrics} we provide a detailed explanation of two content invariant metrics: the \IS\ and \FID, in \cref{content_invariant_metrics} we present an overview of the most commonly used content invariant metrics.

\subsection{Content variant metrics}\label{content_variant_metrics}

In the following we elucidate two content variant metrics: \IS\ in \cref{def:score:var} and \FID\ in \cref{def:score:var2}.  %
Content-invariant metrics are tools to measure a model's ability to generate diverse images.

\subsubsection{Inception score}\label{sec:content_metrics}
The \IS, introduced in \cite{salimans2016improved}, has become very popular, see, \eg \cite{ho2020denoising, 9878449, 9887996, Zhou_2022_CVPR}.  It measures the quality and diversity of generated images. 
The quality refers to the realism and clarity of the image, while diversity signifies the variety within the generated images. The model should possess the capability to produce a diverse range of images within a given category.
The \IS\ is based on Inception \cite{szegedy2016rethinking}, an image classification network that returns probability distribution of labels.
Authors of \cite{salimans2016improved} suggest to have $50000$ generated images, divide them in batches, and calculate the mean and standard deviation of \IS\ across them, obtaining a more stable estimate of the \IS.
The goal is to achieve a high \IS, which occurs when the Inception predicts labels with high confidence, suggesting that the generated images are clear and well-defined,
and the discrete \KL\ divergences between the predicted label distributions and the average distribution are high, implying that the generated images are both high-quality and diverse.

\begin{definition}[\KL\ divergence in the discrete case]\label{def:KL_divergence_discrete}
    Let $\di \in \N$ and
    let $v=(v_1,\ldots, v_\di), w= (w_1,\ldots,w_\di)\in (0,\infty)^\di $.
    Then we denote by $\infdivdisc{v}{w} \in \R$ the number given by 
    \begin{equation}
    \infdiv{v}{w} = \sum_{i=1}^{\di} \ln\!\left(\frac{v_i}{w_i}\right)v_i
\end{equation}
    and we call $\infdivdisc{v}{w}$ the \KL\ divergence of $v$ from $w$.
\end{definition}
\begin{definition}[Inception score]\label{def:score:var}
    Let $ K \in \N \backslash \{1\}$, $\di, N \in \N$, $x_1,\ldots,x_K\in \R^{\di}$ and
    let $\bI \colon \allowbreak \R^{\di} \to (0,1)^N$ be a function.
    Then we say that $\textnormal{IS}$ is the Inception score based on the Inception model $\bI$ for the generated images $x_1,\ldots,x_M$ if and only if $\textnormal{I}$ is the real number which satisfies
    \begin{equation}\label{def:score:var:IS}
        \textnormal{I}=\exp\bigg(\frac1{K} \sum_{i=1}^K\Big( \infdivdisc{\bI(x_i)}{\tfrac{1}K\smallsum_{j=1}^K \bI(x_j)} \Big) \bigg)
    \end{equation}
    \cfout.
    \end{definition}

\begin{remark}[Explanations for \cref{def:score:var}]
    In this remark we provide some explanations for \cref{def:score:var}.
    In \cref{def:score:var} we think of  $x_1,\ldots,x_K \in \R^{\di}$ as the new images created by the generative model we aim to evaluate and
    we think of $\bI$ as the pretrained Inception-v3 model \cite{szegedy2016rethinking} 
    which outputs the probability of the input belonging to each of the $N$ possible classes. This model, in particular, has $N=1000$ possible classes. 
    In \cref{def:score:var:IS} the label distributions $\bI(x_i)\in (0,1)^{N}$, $i \in \{1,\ldots,K\}$, are compared to the average of all label distributions $\tfrac{1}K\smallsum_{j=1}^K \bI(x_j)$ using the discrete \KL\ divergence.
    Averaging these \KL\ divergences and  exponentiating gives the \IS. 
 \end{remark}

\subsubsection{Fréchet inception distance}

The \FID, introduced in \cite{Heusel2017GANsTB}, compares the distribution of generated images with the distribution of real images.
Like with \IS, the pretrained Inception model \cite{szegedy2016rethinking} is employed.
However, the model is used without its output layer, that is the activations of the last hidden layer are extracted as the output distribution. 
A lower \FID\ score means that the generated images are closer in distribution to the real images, which is a desirable outcome. The \FID\ score takes into account both the distance between the means of the output distributions (how well the tendency of the generated images matches that of the real images) and the difference in their covariances (how well the variability in the generated images match that of the real images).
This metric is more widely used than \IS\ and represents a common evaluation method, see, \eg \cite{gafni2022makeascene, nichol2022glide}.
\\
\begin{figure}[H]
\centering
\begin{minipage}{0.55\textwidth}\label{table:IS_FID}
  \centering 
  \vspace{-1em}
  \begin{tabular}{lccc}
    \toprule
    Model  & \FID \\
    \midrule
    DALL-E \cite{pmlr-v139-ramesh21a} & $17.89$ \\
    Stable Diffusion \cite{9878449} & $12.63$ \\
    GLIDE \cite{nichol2022glide} & $12.24$ \\
    DALL-E 2 \cite{ramesh2022hierarchical} & $10.39$ \\
    Imagen \cite{saharia2022photorealistic}  & $7.27$ \\
    \bottomrule
  \end{tabular}
\end{minipage}\hfill
\captionof{table}{Evaluation of text-conditional image synthesis on the
$256\times256$ sized MS-COCO \cite{lin2015microsoft}.} 
\end{figure}
\begin{definition}[Fréchet Inception Distance]\label{def:score:var2}
    Let $ K,M \in \N \backslash \{1\}$, $\di, D \in \N$, $x_1,\ldots,x_K, y_1,\ldots, \allowbreak y_M \in \R^{\di}$, let $\bI^-= \allowbreak (\bI^-_1,\ldots,\bI^-_D) \colon \allowbreak \R^{\di} \to\R^D$ be a function,
    and let $\mu^x=(\mu^x_1,\ldots,\mu^x_D), \mu^y=(\mu^y_1,\ldots,\mu^y_D)  \in \R^D$, $\Sigma^x=(\Sigma^x_{j,k})_{(j,k)\in \{1,\ldots,D\}^2},\Sigma^y=(\Sigma^y_{j,k})_{(j,k)\in \{1,\ldots,D\}^2} \allowbreak \in \R^{D \times D}$ satisfy for all $j,k\in \{1,\ldots,D\}$ that
    \begin{equation}
        \mu^x_j=\frac1K \sum_{i=1}^{K}\bI^-_j(x_i), \qquad \Sigma^x_{j,k}=\frac{1}{K-1}\sum_{i=1}^{K}(\bI^-_{j}(x_i)-\mu^x_j)(\bI^-_{k}(x_i)-\mu^x_k), 
    \end{equation}
    \begin{equation}
        \mu^y_j=\frac1M \sum_{i=1}^{M}\bI^-_j(y_i), \qandq \Sigma^y_{j,k}=\frac{1}{M-1}\sum_{i=1}^{M}(\bI^-_{j}(y_i)-\mu^y_j)(\bI^-_{k}(y_i)-\mu^y_k). 
    \end{equation}
    Then we say that $\textnormal{F}$ is the Fréchet inception distance based on the inception model without the last layer $\bI^-$ for the generated images $x_1,\ldots,x_K$ and the reference images $y_1,\ldots, \allowbreak y_M$ if and only if $\textnormal{F}$ is the real number which satisfies
    \begin{equation}\label{def:score:var:FID}
        \textnormal{F}^2=\norm{\mu^x-\mu^y}^2+ \text{tr}(\Sigma^x+\Sigma^y-2(\Sigma^x \Sigma^y)^{\nicefrac12}).
    \end{equation}
    \end{definition}
\begin{remark}[Explanations for \cref{def:score:var2}]
    In this remark we provide some explanations for \cref{def:score:var2}.
    In \cref{def:score:var2} we think of  $x_1,\ldots,x_K \in \R^{\di}$ as the new images created by the generative model we aim to evaluate, we think of $ y_1,\ldots,y_M \in \R^{\di}$ as the real reference images, and
    we think of $\bI^-=(\bI^-_1,\ldots,\bI^-_D) $ as the pretrained Inception-v3 model \cite{szegedy2016rethinking} without the output layer. The last inner dimension of this model, i.e.\ the output dimension of the function $\bI^-$, is $D=2048$. 
    Moreover, we think of $\mu^x, \mu^y \in \R^D$ as the means 
    of the multidimensional gaussian distributions which arise in the last hidden layer of the Inception model from the generated data and the reference data respectively
    and we think of $\Sigma^x, \Sigma^y \in \R^{D \times D}$ as the corresponding covariance matrices. 
    We select the last hidden layer because it captures high-level information.
    The Fréchet inception distance $\textnormal{F} \in \R$ is based on the Fréchet distance between these two multidimensional gaussian distributions. %
    The use of gaussian distributions allows us to  explicitly solve the Fréchet distance, yielding \cref{def:score:var:FID}.
    This choice is motivated by the property of representing the maximum entropy distribution for a given mean and covariance.
 \end{remark}

\subsection{Content invariant metrics}\label{content_invariant_metrics}
We now offer an overview of the most commonly used content invariant metrics, which evaluate the quality of generated images without considering the variety of their content. These metrics focus on how closely the generated images resemble the reference images in terms of structure, detail, and overall quality.

\textbf{Structured Similarity Index Metric.} \SSIM, introduced in \cite{1284395}, is a technique used to measure the similarity between two images, focusing on the structural and visual aspects. It has found many applications in various fields, such as image compression to estimate the quality of compressed images, or image restoration tasks like denoising or super resolution, where it is used to compare the quality of the restored image with the original.
It takes into account how humans perceive images and is known to match well with human judgment of image quality. To do that it divides the images into small, non-overlapping patches and for each corresponding patches it calculates three comparison terms: luminance, contrast, and structure. Then these term are combined together and finally, by averaging over patches, the \SSIM\ is obtained. A higher \SSIM\ score suggests greater similarity between the two images in terms of structure and perception. See \cite{nilsson2020understanding} for more in depth treatment of \SSIM.

\textbf{Peak Signal-to-Noise Ratio.} \PSNR\ compares the level of a desired signal to the level of background noise. 
It is commonly used to quantify reconstruction quality for images and videos subject to loss compression considering as signal the original data and as noise the error introduced by the compression. It is based on the mean squared error between the original and distorted images. A higher \PSNR\ value indicates that the distorted image is more similar to the original image.
This method is widely used as metric but has some limitations. 
It may not consistently be aligned with human perception, 
it relies on pixel-wise differences and it doesn't consider visual elements when evaluating image quality.
In situations where human perception is an important factor, metric like the \SSIM\ is often preferred.
See \cite{5596999,Al-Najjar} for more in depth treatment of \PSNR.

\textbf{Learned Perceptual Image Patch Similarity.} \LPIPS\ \cite{zhang2018unreasonable} measures perceptual similarity rather than focusing on the quality.
Trained on large datasets to closely align with human visual perception, \LPIPS\ uses a deep neural network to achieve a perceptual similarity metric. This metric goes beyond pixel-wise distinctions, capturing high-level structural information. The goal of the training is to minimize perceptual differences between image pairs, guided by human judgment. \LPIPS\ is widely recognized for its ability to better match human perception, making it a valid metric, see, \eg \cite{9156570}.

\section{Advanced variants and extensions of DDPMs}
\label{sec:advanced}
In this section, we explore some successful improvements of the \DDPM\ scheme in \cite{sohl2015deep,ho2020denoising} from the scientific literature.
We begin by discussing the innovations introduced in the so-called Improved \DDPM\ \cite{ho2020denoising} in \cref{sec_improved_ddpm}.
Next, in \cref{sec_ddim} we present and explain the \DDIM\ scheme in \cite{song2022denoising}.
In \cref{sec_classifier_free_guidance} we introduce the classifier-free diffusion guidance from \cite{ho2022classifierfree} and we highlight how class information is integrated into the model architecture.
Thereafter, in \cref{sec_stable_diffusion} stable diffusion \cite{9878449} is presented, explaining how textual information can be incorporated in image generation.
Finally, in \cref{sec_stae_of_art} we explore additional state of the art techniques at a high level.

\subsection{Improved DDPM}\label{sec_improved_ddpm}
In \cite{ho2020denoising} the authors find that \DDPMs\ can generate high fidelity samples according to \FID\ and \IS\ but it fails to achieve competitive \ENLL\ (cf.\ \cref{sec:training_obj}). This suggests that the scheme generates high-quality outputs but does not capture the diversity of the data distribution.
Motivated by this observation, the authors of \cite{nichol2021improved} 
investigate the reasons behind the high \ENLL\ and propose several modifications to improve the algorithm.
\begin{itemize}
    \item They learn the variances in the backward process %
    rather than assuming they are fixed, %
    as in \cref{setting:dnn3}. %
    \item They replace the linear rate scheduler described in \cref{remark:def:alpha} with a cosine scheduler. 
    \item They increase the number of time steps during training while attempting to reduce the number of steps during sampling.
\end{itemize}
This new algorithm is known as Improved \DDPM. We present its methodology, following a similar structure to \cref{setting:dnn3}, in \cref{setting:improvedDDPM}. The proposed scheme is based on the work of \cite{nichol2021improved}.

\cfclear
\begin{method}[Improved \DDPM\ generative method]
\label{setting:improvedDDPM}
    Let $\di ,\fd, \ds \in \N$, $T\in \N \backslash\{1\}$,  $\gamma \in (0, \infty)$, $\al_1,\ldots,\al_T \in (0,1)$, $ \tal_0,\tal_1, \ldots, \tal_T,\tbe_1,\ldots,\tbe_T \in [0,1]$, assume for all $t \in \{0,1, \ldots, T\}$ that $\tal_t= \textstyle\prod_{s=1}^t\al_s$,
    assume for all $t \in \{1, \ldots, T\}$ that 
    $\tbe_t=\left[\frac{1-{\tal}_{t-1}}{1-{\tal}_t}\right](1-\al_t)$,
	for every $\theta \in \R^{\fd}$ let 
	$\bV^\theta =(\bv^\theta_1,\bv^\theta_2)=((\bv^\theta_{1,1},\ldots,\bv^\theta_{1,\di}),(\bv^\theta_{2,1}\ldots,\bv^\theta_{2,\di})) \colon \R^{\di} \times \{1,\ldots,T\} \to\R^{\di} \times (-1,1)^{\di}$
	be a function,
let $\tilde{\mu}= (\tilde{\mu}_t)_{t \in \{1,\ldots,T\}}\colon \R^{\di} \times \R^{\di} \times \{1,\ldots,T\} \to \R^{\di}$ satisfy for all $x, y \in \R^{\di}$, $t \in \{1,\ldots,T\}$ that
\begin{equation}
    \tilde{\mu}_t(x,y)=\bigg[\frac{\sqrt{\al_t}(1-\tal_{t-1})}{1-\tal_t}\bigg] x + \bigg[\frac{\sqrt{\tal_{t-1}}(1-\al_t)}{1-\tal_t}\bigg]   y,%
\end{equation}
for every $\theta \in \R^{\fd}$ let $\mu^\theta=(\mu^\theta_t)_{t \in \{1,\ldots,T\}} \colon \R^{\di} \times \{1,\ldots,T\}\to \R^{\di}$ satisfy for all $x \in \R^{\di}$, $t \in \{1,\ldots,T\}$ that
\begin{equation}
    \mu^{\theta}_t(x)= \frac{1}{\sqrt{\al_t}} \bigg( x - \frac{1-\al_t}{\sqrt{1-\tal_t}} \bv_1^\theta(x,t)\bigg),
\end{equation}
for every $\theta \in \R^{\fd}$ let $\Sigma^\theta = (\Sigma^\theta_t)_{t \in \{1,\ldots,T\}} = ((\Sigma^\theta_{t,i,j})_{(i,j)\in\{1,\ldots,\di\}^2})_{t \in \{1,\ldots,T\}}\colon \R^{\di} \times \{1,\ldots,T\}\to \R^{\di\times\di}$ satisfy for all $x \in \R^{\di}$, $t \in \{1,\ldots,T\}$, $i,j\in\{1,\ldots,\di\}$ that
\begin{equation}\label{setting:improvedDDPM:sigmatheta}
    \Sigma^{\theta}_{t,i,j}(x) = 
    \begin{cases}
           \exp\big(\bv_{2,i}^{\theta}(x,t) \log(1-\al_t)+ (1-\bv_{2,i}^{\theta}(x,t))\log(\tbe_t)\big) & \colon i=j \\
           0 & \colon i\neq j,
    \end{cases}
\end{equation}
let $\delta_{+} \colon [-1,1] \to \R \cup \{\infty\}$ and $\delta_{-}\colon [-1,1] \to \R \cup \{-\infty\}$ satisfy for all $x \in [-1,1]$ that
\begin{equation}
\delta_{+}(x)  =\left\{\begin{array}{ll}
\infty & \text { if } x=1 \\
x+\frac{1}{255} & \text { if } x<1
\end{array} 
\qandq 
\delta_{-}(x)= \begin{cases}-\infty & \text { if } x=-1 \\
x-\frac{1}{255} & \text { if } x>-1,\end{cases} \right.
\end{equation}
let $L \colon \R^\fd \times \R^\fd \times [-1,1]^{\di} \times \R^{\di} \times \{1,\ldots,T\} \to \R$ satisfy for all $\theta, \tilde{\theta} \in  \R^\fd $, $x=(x_1,\ldots,x_\di )\in [-1,1]^{\di}$, $\varepsilon \in \R^{\di}$, $t \in \{1,\ldots,T\}$ that
\begin{equation}\label{setting:improvedDDPM:Lt}
L(\theta, \tilde{\theta}, x, \varepsilon, t)= 
    \begin{cases}
    \begin{aligned}
        & - \log\bigg(\int_{\delta_{-}(x_1)}^{\delta_{+}(x_1)}\ldots \int_{\delta_{-}(x_{\di})}^{\delta_{+}(x_{\di})} \n \Big(y , \mu_t^{\tilde{\theta}} \big(\sqrt{\tal_t} x + \sqrt{1-\tal_t} \varepsilon \big), \\
        & \quad  \Sigma_t^\theta \big(\sqrt{\tal_t} x + \sqrt{1-\tal_t} \varepsilon\big) \Big) \, \d y_1 \ldots \, \d y_\di \bigg)
    \end{aligned}
      &  \colon t=1\\
      \begin{aligned}
    & \infdivcomplex
    \bigg(
    \n\big(
    \cdot, \tilde{\mu}_t 
  (\sqrt{\tal_t} x + \sqrt{1-\tal_t} \varepsilon, x)  ,\tilde{\beta}_{t}\mathbb{I}
  \big) 
  \;\|\; \\
  & \quad \n\Big( \cdot,\mu_t^{\tilde{\theta}} \big(\sqrt{\tal_t} x + \sqrt{1-\tal_t} \varepsilon \big),  \Sigma^\theta_t\big(\sqrt{\tal_t} x + \sqrt{1-\tal_t} \varepsilon \big)
  \Big) 
  \bigg)
    \end{aligned}
      & \colon t>1,
\end{cases}
\end{equation}
let  
$\fL \colon \R^\fd \times \R^\fd \times [-1,1]^\di \times \R^\di \times \{1,\ldots,T\} \to \R$
satisfy 
for all $ \theta, \tilde{\theta} \in \R^{ \fd } $, $x\in [-1,1]^{\di}$, $\varepsilon \in \R^{\di}$, $t \in \{1,\ldots,T\}$
that
\begin{equation}\begin{split} \label{setting:improvedDDPM:loss}
    \textstyle  \fL( \theta,\tilde{\theta},x,\varepsilon, t ) =& 
   \bigl\|
   \varepsilon - 
   \bv_1^{ \theta }\bigl( \sqrt{\tal_t} x + \sqrt{1-\tal_t} \varepsilon, t \bigr)
    \bigr\|^2 + \lambda L(\theta, \tilde{\theta}, x, \varepsilon, t),
\end{split}
\end{equation}
let 
$
  \fG \colon \R^\fd \times \R^\fd \times [-1,1]^\di \times \R^\di \times \{1,\ldots,T\} \to \R^{ \fd }
$ 
satisfy for all 
$\tilde{\theta} \in \R^\fd$,
$x\in [-1,1]^{\di}$, 
$\varepsilon \in \R^{\di}$,
$t \in \{1,\ldots,T\}$, 
$\theta \in \R^\fd$ with 
$\fL(\cdot,\tilde{\theta},x,\varepsilon,t)$ differentiable at $\theta$
that
\begin{equation}
\textstyle 
  \fG( \theta, \tilde{\theta}, x, \varepsilon, t ) = 
 ( \nabla_{ \theta } \fL )( \theta, \tilde{\theta}, x, \varepsilon, t ),
\end{equation}
let $(\Omega, \cF, \mathbbm{P})$ be a probability space, 
    let
    $\datax_{n,i} \colon \Omega \to [-1,1]^\di$, $ n, i \in \N$, be random variables,
    let $\noise_{n_
    ,i} \colon \Omega \to \R^{ \di }$, $ n, i \in \N$, 
    be i.i.d.\ standard normal random variables,
let $\randT_{n} \colon \Omega \to \{1,2,\ldots,T\}$, $ n\in \N$,
be independent $\mathcal{U}_{\{1,2,\ldots,T\}}$-distributed random variables,
let 
$ \Theta  \colon \N_0 \times \Omega \to \R^{ \fd } 
$
be a stochastic process which satisfies 
for all $n\in\N$ that
\begin{equation}
\Theta_{n}=\Theta_{n-1}-\gamma \bigg[\frac{1}{\ds}\sum_{i=1}^{\ds}\fG( \Theta_{n-1}, \Theta_{n-1}, \datax_{n,i}, \noise_{n,i}, \randT_{n} )\bigg],
\end{equation}
let $N \in \N$, $K\in \{2,3,\ldots, T\}$, let $Z_k = (Z_{k,i})_{i \in \{1,\ldots,\di\}} \colon \Omega \to \R^{ \di }$, $k\in\{1,\ldots,K+1\}$,
be i.i.d.\ standard normal random variables, let $t_0,t_1,\ldots,t_K \in \{0,1,\ldots,T\}$ satisfy for all $k \in \{1,\ldots,K\}$ that $t_k=1+\lfloor \nicefrac{(k-1)(T-1)}{(K-1)} \rfloor$ and $t_0=0$,
let $\backX = (\backX_{k})_{k \in \{0,1,\ldots,K\}}= ((\backX_{k,i})_{i\in\{1,\ldots,\di\}})_{k \in \{0,1,\ldots,K\}} \colon \allowbreak \{0,1,\ldots,K\} \times \Omega \to \R^{\di}$ be a stochastic process, 
and assume for all $k \in \{1,\ldots,K\}$, $i \in \{1,\ldots,\di\}$ that
\begin{equation} 
\backX_{K}=Z_{K+1} 
\end{equation}
\begin{equation}\label{setting:improvedDDPM:backx}
\begin{split}
    \andq & \backX_{k-1,i}= 
\frac{1}{\sqrt{\nicefrac{\tal_{t_k}}{\tal_{t_{k-1}}}}} \bigg( \backX_{k,i} - \frac{1-(\nicefrac{\tal_{t_k}}{\tal_{t_{k-1}}})}{\sqrt{1-\tal_{t_k}}} \bv_{1,i}^{\Theta_N}(\backX_{k},t_k)\bigg) + \\
& \quad 
\bigg[\exp\bigg(\bv_{2,i}^{\Theta_N}(\backX_{k},t) \log\big(1-(\nicefrac{\tal_{t_k}}{\tal_{t_{k-1}}})\big)\\
& \quad + (1-\bv_{2,i}^{\Theta_N}(\backX_{k},t))\log\Big(\Big[\frac{1-\tal_{t_{k-1}}}{1-\tal_{t_{k}}}\Big]\big(1-(\nicefrac{\tal_{t_k}}{\tal_{t_{k-1}}})\big)
\Big)
\bigg)\bigg]^{\nicefrac{1}{2}}  Z_{k,i}
\end{split}
\end{equation}
\cfout.
\end{method}

\begin{remark}[Explanations for \cref{setting:improvedDDPM}]\label{setting:improvedDDPM:remark}  
    In this remark we provide some intuitive and theoretical explanations for \cref{setting:improvedDDPM}
    and describe in what sense \cref{setting:improvedDDPM} aims to improve \cref{setting:dnn3}.

    Roughly speaking, the approach outlined in \cref{setting:improvedDDPM}, similar to \cref{setting:dnn3}, aims to minimize the \ENLL\ by reducing the upper bound in \cref{lemma:upperboundE} with the final goal of generating new samples that follow the initial data distribution. However, in this case the upper bound cannot be rewritten as simply as in \cref{prop:ddpm:noise}.
    The key difference is that the variances of the backward process $(\Sigma^\theta)_{\theta\in\R^\fd}$ are not fixed, unlike in \cref{sec:Obj_gaussian}. In \cite{ho2020denoising}, authors of \DDPM\ found that directly predicting the backward variances lead to unstable training and lower sample quality compared to using fixed variances. This problem arises because the variance values are very low and \ANNs\ often fail to predict them due to vanishing gradients.
    To obtain $(\Sigma^\theta)_{\theta\in\R^\fd}$ we now interpolate for every $t \in \{1,\ldots,T\}$ the numbers $(1-\al_t) $ and $\tbe_t $ (cf.\ \cite[Section 3.2]{ho2020denoising} for the extreme choices in the interpolation) in the logarithmic domain which results in more stable variance predictions.
    This is achieved using the interpolation parameter $(\bv^\theta_2)_{\theta \in \R^{\fd}}$ that arises from
    $(\bV^\theta)_{\theta \in \R^{\fd}} =(\bv^\theta_1,\bv^\theta_2)_{\theta \in \R^{\fd}}$.
     Simulations show that the choice of $(\Sigma^\theta)_{\theta\in\R^\fd}$ becomes less significant as the diffusion step increases, since $(1-\al_t)_{t \in \{1,\ldots,T\}}$ and $(\tbe_t)_{t \in \{1,\ldots,T\}}$  are nearly identical except for early time steps. Nevertheless, selecting appropriate backward variances can help to reduce the \ENLL\ during the first diffusion steps which are shown to contribute the most (cf. \cite{nichol2021improved}). 
     We think of $(\bV^\theta)_{\theta \in \R^{\fd}} $ as the \ANN\ which has a double output dimension compared to \cref{setting:dnn3}, we think of $(\bv^\theta_1)_{\theta \in \R^{\fd}}$ as the usual prediction of the noise component of the noisy data, and we think of $(\bv^\theta_2)_{\theta \in \R^{\fd}}$ as the object needed to calculate the learnable variance. 

    Furthermore, we think of $\fL$ as the loss used during the training, compared to \cref{setting:dnn3} it is adjusted by adding the term $L$ due to the previous changes. 
    This new term corresponds to an explicit version of the upper bound found in \cref{lemma:upperboundE} and it is designed to guide the learning of the variance, without any influence of the mean $(\mu^\theta)_{\theta \in \R^{\fd}}$.
    To achieve this, a new parameter is introduced in $L$ specifically to block the backpropagation process of the mean.
    Note that in the case $t=1$, assuming that the input data consists of values in $\{0,1,\ldots,255\}$ (\eg images) rescaled to $[-1,1]$, the term $L$ is the log-probability of returning to the correct bins. This final step ensures that the backward process is performed consistently with the original data distribution. 
    In the experiments we assume $\lambda = 0.001$  to prioritize the error between the true and the predicted noise rather than the prediction of the variance.

    Consistently with \cref{setting:dnn3}, we think of
        $\fG$ 
    as the generalized gradient of the loss $\fL$ with respect to the trainable parameters,
    we think of 
        $\datax_{n,i}$, $ n, i \in \N$, 
    as random samples of the initial value of the forward process used for training,
    we think of
        $\mathcal{E}_{n,i}$, $ n, i \in \N$,
    as the noise components of the forward process used for training,
    we think of 
        $\randT_{n}$, $ n\in \N$,
    as random times used to determine which terms of the upper bound are considered in each training step,
    we think of
        $(\Theta_{n})_{n\in\N_0}$
    as the training process for the parameters of the backward process given by an \SGD\ process for the generalized gradient $\fG$
    with learning rate $\gamma$,
    batch size $\ds$, and
    training data
    $(\datax_{n,i}, \mathcal{E}_{n_,i}, \randT_{n})_{(n,i) \in \N^2}$,
    we think of
        $\nn$
    as the number of training steps,
    we think of 
        $K \in \{2,3,\ldots,T\}$
    as the time steps in the backward process,
    we think of
        $Z_k$, $k\in\{1,\ldots,K+1\}$,
    as the noise components of the backward process,
    and we think of
        $\backX$
    as the backward process for the trained parameters $\Theta_{\nn}$.
     Compared to \cref{setting:dnn3}, the backward process has been optimized by reducing the number of steps. In the sampling phase we select $K$ evenly space real numbers between $1$ and $T$, rounding them down to the nearest integer to obtain the sampling steps $t_1,\ldots,t_K$. 
    This adjustment impacts the structure of the means $\mu^{\Theta_N}$ and variances $\Sigma^{\Theta_N}$ in the backward process (cf. \cref{lemma:constructive:item2} in \cref{lemma:constructive}), requiring a slightly modified versions of these functions, with coefficient rescaled to account for a shorter diffusion process.
  The model needs only $K = 100$ sampling steps to achieve almost the same \FID\ reached using the $T = 4000$ sampling steps.
 Given these assumptions we expect that the terminal value $X_0$ of the trained backward process will be roughly aligned with the distribution we aim to sample from.
\end{remark}

\begin{remark}[Choice of noise intensity in \cref{setting:improvedDDPM}] \label{remark:def:alpha2}
Another significant improvement in \cite{nichol2021improved} is the introduction of the following cosine scheduler to define $(\tal_t)_{t\in\{0,1,\ldots,T\}}$ in \cref{setting:improvedDDPM}. Assume \cref{setting:improvedDDPM}, let $s \in (0,1)$ and assume for all $t \in \{1,\ldots,T\}$ that 
\begin{equation}
    \tal_t=\cos\bigg(\frac{(\nicefrac{t}{T}+s)\pi}{(1+s)2}\bigg)^2 \cos\bigg(\frac{s\pi}{(1+s)2}\bigg)^{-2}.
\end{equation}
This choice, assuming  for all $t \in \{1,\ldots,T\}$ that $\tal_t= \textstyle\prod_{s=1}^t\al_s$, allows to define $(1-\al_t)_{t \in \{1,\ldots,T\}}$, which represent the measurements of noise added in the t-th time step.
The linear noise scheduler (cf.\ \cref{remark:def:alpha}) worked well for high resolution inputs but is sub-optimal for low resolution (\eg $64 \times 64$ and $32 \times 32$), too quickly in the forward process the input is not far from pure gaussian noise, making it difficult to learn the backward process. The new cosine scheduler permits to add noise slower preserving input information for later time steps.
The offset $s\in (0,1)$ is introduced to prevent $(1-\al_t)_{t \in \{1,\ldots,T\}}$ from becoming too low near $t=0$. 
Authors of \cite{nichol2021improved} assume $s=0.008$. Another precaution taken in practise is to clip $(1-\al_t)_{t \in \{1,\ldots,T\}}$ to be no larger than $0.999$. This clipping helps to avoid singularities near the terminal time step $T$. 
\end{remark}

\begin{figure}[H]
\centering
\subfigure%
{\includegraphics[width=1.\textwidth]{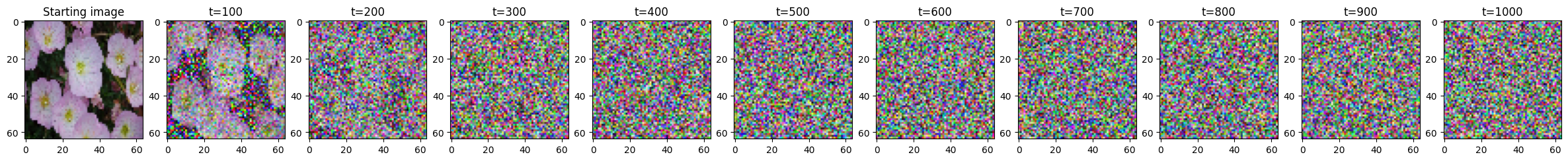}}\\
\subfigure%
{\includegraphics[width=1.\textwidth]{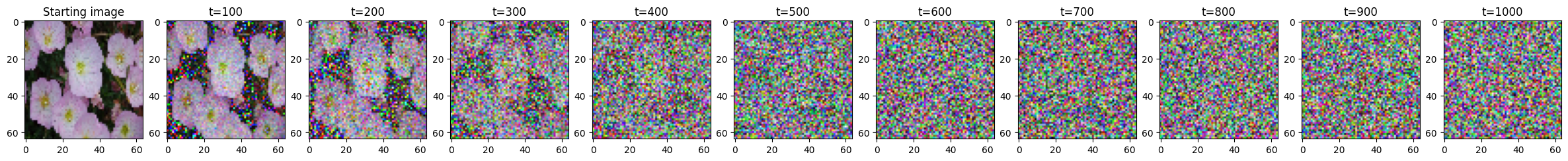}}
\caption{Forward diffusion process using a linear scheduler on top and a cosine scheduler at the bottom. The code to generate these plots can be found in \url{https://github.com/deeplearningmethods/diffusion_model}.}
\label{fig:ReLU,h20}
\end{figure}

\subsection{Denoising Diffusion Implicit Model (DDIM)}\label{sec_ddim}
\DDPMs\ have demonstrated impressive generation quality. However, they necessitate the simulation of a Markov process over numerous time steps to generate a sample. \DDIMs\ presented in \cite{song2022denoising}, introduce a more efficient way to generate data redefining the diffusion process as a non-Markovian process while maintaining the same training objective as \DDPMs.
In \cref{math_ddim_scheme}, we present a new mathematical framework without the Markov assumptions (cf.\ \cref{setting_ddim}). We justify the use of the same training objective as \DDPMs\ in \cref{ddim:exp_train_obj}. Finally, in \cref{ddim_scheme} we discuss the methodology employed in \DDIMs\ (cf.\ \cref{setting:DDIM}).

\subsubsection{Framework for DDIM}\label{math_ddim_scheme}

\cfclear
\begin{setting}[General framework for \DDIMs]\label{setting_ddim}
Assume \cref{setting0},  let $\sigma_1,\ldots,\sigma_T, \tal_1, \ldots, \tal_T \in (0,1)$ satisfy for all $t \in \{2,3, \ldots, T\}$ that $\sigma_t^2\leq 1- \tal_{t-1}$,
for every $\theta \in \R^{\fd}$ let $\bV^\theta \colon \R^{\di} \times \{1,\ldots,T\} \to\R^{\di}$ be a function,
for every $\theta \in \R^{\fd}$ let $\f\theta \colon \R^{\di} \times \{1,\ldots,T\} \to\R^{\di}$ satisfy for all $x \in \R^{\di}$, $t \in \{1,\ldots,T\}$ that $\f{\theta}(x,t) = (\sqrt{\tal_t})^{-1}(x-\sqrt{1-\tal_{t}} \bV^\theta(x, t)) $,
and assume for all $\theta \in \R^{\fd}$, $t \in \{2,3,\ldots,T\}$, $x_0,x_1,\ldots,x_T \in \R^{\di}$ that
\begin{equation}\label{setting_ddim:hp1}
        \pcond{\emptyset}{1,\ldots,T}{0}\argpcond{(x_1,\ldots,x_T}{x_0)}=\textstyle  \pcond{\emptyset}{T}{0}\argpcond{(x_T}{x_0)} \prod_{s=2}^{T} \pcond{\emptyset}{s-1}{s,0}\argpcond{(x_{s-1}}{x_{s},x_0)},
\end{equation}
\begin{equation}\label{setting_ddim:hp2}
    \pcond{\emptyset}{T}{0}\argpcond{(x_T}{x_0)}= \n(x_T, \sqrt{\tal_T} x_0, (1-\tal_T) \mathbb{I}),
\end{equation}
\begin{equation}\label{setting_ddim:hp3}
    \pcond{\emptyset}{t-1}{t,0}\argpcond{(x_{t-1}}{x_t,x_0)}=\n\bigg(x_{t-1}, \sqrt{\tal_{t-1}}x_0 + \sqrt{1-\tal_{t-1}-\sigma_t^2}\Big(\frac{x_t-\sqrt{\tal_t}x_0}{\sqrt{1-\tal_t}}\Big) , \sigma_t^2 \mathbb{I}\bigg),
\end{equation}
\begin{equation}\label{setting_ddim:hp4}
    \pjoint{\theta}(x_0,x_1,\ldots,x_T) = \p{\theta}{T}(x_T) \left[ \textstyle \prod_{s=1}^{T} \pcond{\theta}{s-1}{s} \argpcond{(x_{s-1}}{ x_{s} )} \right],
\end{equation}
\begin{equation}\label{setting_ddim:hp5}
    \pcond{\theta}{t-1}{t}\argpcond{(x_{t-1}}{x_{t})}=\pcond{\emptyset}{t-1}{t,0}\argpcond{(x_{t-1}}{x_t, \f{\theta}(x_t,t))},
\end{equation}
\begin{equation}\label{setting_ddim:hp6}
        \andq \pcond{\theta}{0}{1}\argpcond{(x_{0}}{x_1)}=\n(x_0,\f{\theta}(x_1,1), \sigma_1^2 \mathbb{I})
\end{equation}
\cfout.
\end{setting}

\begin{remark}[Explanations for \cref{setting_ddim}]
    In this remark we provide  intuitive explanations for \cref{setting_ddim}.
    Roughly speaking, in \DDIMs, differently from \DDPMs, we consider a non-Markovian forward process.
    The transition kernels for the backward process imitate the behaviour of $ \pcond{\emptyset}{t-1}{t,0}$, $ t \in \{2,3,\ldots,T\}$, where
    instead of the denoised data, a prediction based on $(\f\theta )_{\theta \in \R^{\fd}}$ is employed.
    We think of $(\bV^\theta)_{\theta \in \R^{\fd}}$ as the \ANN\ responsible for predicting the noisy component given a noisy input and a time step. This implies that $(\f\theta )_{\theta \in \R^{\fd}}$ represents the estimate of the initial data from an arbitrary time step.
    \Nobs that in the case $t=1$ the transition kernel formula for the backward process is adjusted to guarantee that the generative process is valid across the entire time range.
\end{remark}

\subsubsection{Distribution for the forward process in DDIM}
    In \cref{ddim:cond_prob} below we show that the means and variances of the conditional \PDFs\ for the forward process (cf.\ \cref{setting_ddim:hp3} in \cref{setting_ddim}) are chosen to ensure that 
    the conditional distribution of any time step of the forward process given the initial value of the forward process is again given by a Gaussian distribution.  This result coincides with the one found in \cref{lemma:prodq} for \DDPM\ and permits to accelerate the forward process skipping from the initial time step directly to the desired time step.

\cfclear
\cfconsiderloaded{def:gaussian}
\begin{lemma}[Multi-step transition density of the forward process]\label{ddim:cond_prob}
   Assume \cref{setting_ddim}. Then it holds for all $t\in \{1,\ldots,T\}$, $x_0,x_t \in \R^{\di}$ that \begin{equation}\label{ddim:cond_prob:thesis}
       \pcond{\emptyset}{t}{0}\argpcond{(x_t}{x_0)}= \n(x_t, \sqrt{\tal_{t}}x_0, (1-\tal_t)\mathbb{I}).
   \end{equation}
   \cfout
\end{lemma}
\begin{cproof}{ddim:cond_prob}
    We prove \cref{ddim:cond_prob:thesis} by induction.
    \Nobs that \cref{setting_ddim:hp2} assures that for all $x_{T},x_0 \in \R^{\di}$ it holds that $\pcond{\emptyset}{T}{0}\argpcond{(x_T}{x_0)}= \n(x_T, \sqrt{\tal_T} x_0, (1-\tal_T) \mathbb{I})$.
    For the induction step let $t \in \{1,\ldots,T-1\}$ and assume that for all $x_{t+1},x_0 \in \R^{\di}$ it holds that $\pcond{\emptyset}{t+1}{0}\argpcond{(x_{t+1}}{x_0)}= \n(x_{t+1}, \sqrt{\tal_{t+1}}x_0, (1-\tal_{t+1})\mathbb{I}).$ This, \cref{setting_ddim:hp3}, and \cref{lemma:condprob} assure that for all $x_t, x_0 \in \R^\di$ it holds that
    \begin{equation}
    \begin{split}
         \pcond{\emptyset}{t}{0}\argpcond{(x_t}{x_0)} &=
            \int_{\R^\di}\pcond{\emptyset}{t}{t+1,0}\argpcond{(x_{t}}{x_{t+1},x_0)}
            \pcond{\emptyset}{t+1}{0}\argpcond{(x_{t+1}}{x_0)} \, \d x_{t+1}\\
            &= \int_{\R^\di}\n\bigg(x_{t}, \sqrt{\tal_{t}}x_0 + \sqrt{1-\tal_{t}-\sigma_{t+1}^2}\Big(\frac{x_{t+1}-\sqrt{\tal_{t+1}}x_0}{\sqrt{1-\tal_{t+1}}}\Big) , \sigma_{t+1}^2 \mathbb{I}\bigg)\\
            & \quad \n(x_{t+1}, \sqrt{\tal_{t+1}}x_0, (1-\tal_{t+1})\mathbb{I}) \, \d x_{t+1}\\
            &=\n\bigg(x_t,\sqrt{\tal_{t}} x_0 + \sqrt{1-\tal_t - \sigma_{t+1}^2} \frac{\sqrt{\tal_{t+1}}x_0-\sqrt{\tal_{t+1}}x_0}{\sqrt{1-\tal_{t+1}}},\\
          & \quad \sigma_{t+1}^2 \mathbb{I} + \frac{1-\tal_t-\sigma_{t+1}^2}{1-\tal_{t+1}}(1-\tal_{t+1})\mathbb{I}\bigg) \\
          & = \n\big(x_t,\sqrt{\tal_{t}} x_0, (1-\tal_{t})\mathbb{I}\big).
    \end{split}
    \end{equation}
    Induction thus establishes \cref{ddim:cond_prob:thesis}.
\end{cproof}

\subsubsection{Explicit objective function in DDIM}\label{ddim:exp_train_obj}
In \cite[Theorem 1]{song2022denoising}, it is shown that \DDIMs\ can use the same training objective as \DDPMs,
despite being defined by a non-Markovian forward process.

\cfclear
\cfconsiderloaded{def:gaussian}
\begin{theorem}[Explicit bound for negative log-likelihood]\label{ddim:lemma:loss}
   Assume \cref{setting_ddim}, let $\theta \in \R^{\fd}$ and for every $t \in \{1,\ldots,T\}$ let $\mathcal{E}_t \colon \Omega \to \R^{ \di }$ satisfy for all $B \in \cB(\R)$ that
    $\Prob{}(\mathcal{E}_t \in B)= \int_B \n(x,0,\mathbb{I})\, \d x$, $\mathcal{E}_t$ and $X_0^\emptyset$ are independent, and $X_t^\emptyset= \sqrt{\tal_t} X_0^\emptyset+\sqrt{1-\tal_t} \mathcal{E}_t$.
    Then there exist $\gamma_1,\ldots,\gamma_T \in [0,\infty)$ and $C \in \R$ such that 
   \begin{equation}\label{ddim:lemma:loss:eq}
   \begin{split}
       & \negloglike[\big]{\p{\emptyset}{0}}{\p{\theta}{0}}=\Eb{}{-\ln (\p{\theta}{0}(X_{0}^\emptyset)) } \\
       &  \leq C +
        \sum_{t=1}^T \gamma_t \mathbb{E}_{}\left[\left\|\bV^\theta\left(\sqrt{\tal_t} X_0^\emptyset+\sqrt{1-\tal_t} \mathcal{E}_t, t \right)-\mathcal{E}_t \right\|^2\right]
    \end{split}
   \end{equation}
   \cfout.
\end{theorem}
\begin{cproof}{ddim:lemma:loss}
    \Nobs that \cite[Theorem 1]{song2022denoising} proves \cref{ddim:lemma:loss:eq}.
\end{cproof}

\subsubsection{Generative method}\label{ddim_scheme}
We now formulate the generative method based on the upper bound found in \cref{ddim:lemma:loss}. 
\DDIMs\ as a result of the non-Markovian formulation allow us to do the training employing the full number of training steps and to sample using fewer steps maintaining high quality.  The scheme was proposed in \cite{song2022denoising}.

\cfclear
 \begin{method}[\DDIM\  generative method]\label{setting:DDIM}
     Let $\di ,\fd, \ds \in \N$, $T\in \N \backslash\{1\}$,
 $\gamma \in (0, \infty)$, $\al_1,\ldots,\al_T\in (0,1)$, $\tal_0,\tal_1, \ldots, \tal_T \in (0,1]$, assume for all $t \in \{0, 1, \ldots, T\}$ that $\tal_t= \textstyle\prod_{s=1}^t\al_s$,
	for every $\theta \in \R^{\fd}$ let 
	$\bV^\theta \colon \R^{\di} \times \{1,\ldots,T\} \to\R^{\di}$
	be a function,
    let  $\fL \colon \R^\fd \times \R^\di \times \R^\di \times \{1,\ldots, T\} \to \R$
satisfy 
for all $ \theta \in \R^{ \fd } $, $x,\varepsilon \in \R^\di$, $t \in \{1,\ldots, T\}$
that
\begin{equation}
 \fL( \theta, x, \varepsilon, t )= \bigl\| \varepsilon - \bV^{ \theta }\bigl( \sqrt{\tal_t} x + \sqrt{1-\tal_t} \varepsilon, t \bigr)  \bigr\|^2,
\end{equation}
let 
$
  \fG \colon \R^\fd \times \R^\di \times \R^\di \times \{1,\ldots,T\} \to \R^{ \fd }
$ 
satisfy for all 
$x, \varepsilon \in \R^\di$, 
$t \in \{1,\ldots,T\}$, 
$\theta \in \R^\fd$ with $\fL( \cdot,x, \varepsilon, t)$ differentiable at $\theta$
that
\begin{equation}
  \fG( \theta, x, \varepsilon, t ) = 
 ( \nabla_{ \theta } \fL )( \theta, x, \varepsilon, t ),
\end{equation}
let $(\Omega, \cF, \mathbbm{P})$ be a probability space, 
    let $\datax_{n,i} \colon \Omega \to \R^\di$, $ n, i \in \N$, be random variables, 
    let $\noise_{n_,i} \colon \Omega \to \R^{ \di }$, $ n, i \in \N$, be i.i.d.\ standard normal random variables,
    let  $\randT_{n} \colon \Omega \to \{1,\ldots,T\}$, $ n\in \N$, be independent $\mathcal{U}_{\{1,2,\ldots,T\}}$-distributed random variables,  
let 
$ \Theta  \colon \N_0 \times \Omega \to \R^{ \fd } 
$
be a stochastic process which satisfies 
for all $n\in\N$ that
\begin{equation}
\Theta_{n}=\Theta_{n-1}-\gamma \bigg[\frac{1}{\ds}\sum_{i=1}^{\ds}\fG( \Theta_{n-1}, \datax_{n,i}, \noise_{n,i}, \randT_{n} )\bigg],
\end{equation}
let $N\in \N$, $K\in \{2,3,\ldots, T\}$, let $Z_k \colon \Omega \to \R^{\di}$, $k\in\{1,\ldots,K+1\}$,
be i.i.d.\ standard normal random variables,
let $\tk_0, \tk_1, \ldots, \tk_K \in \{0,1,\ldots\, T\}$ satisfy for all $k \in \{1,\ldots,K\}$ that $\tk_{k-1}<\tk_k$, $\tk_0=0$, and $\tk_K=T$,
let $\eta,\sigma_{\tk_1},\ldots,\sigma_{\tk_K} \in [0,1]$ satisfy for all $k\in\{1,\ldots,K\}$ that $\sigma_{\tk_k}=\eta \sqrt{(1 - \al_{\tk_k} )(1 - \tal_{\tk_{k-1}}) (1 - \tal_{\tk_k})^{-1} }$ and $\sigma_{\tk_k}^2\leq1-\al_{\tk_{k-1}}$,
let $\backX=(\backX_{k})_{k \in \{0,1,\ldots,K\}} \colon \{0,1,\ldots,K\} \times \Omega \to \R^{ \di }$ be a stochastic process,
and assume for all $k \in \{1,\ldots,K\}$ that
\begin{equation}
    \backX_K=Z_{K+1}
\end{equation}
\begin{equation} \label{setting:DDIM:backx}
\begin{split}
    \andq \backX_{k-1} & = \sqrt{\tal_{\tk_{k-1}}} \bigg[ \frac{1}{\sqrt{\tal_{\tk_k}}}\bigg(\backX_{k}-\sqrt{1-\tal_{\tk_k}}\bV^{\Theta_\nn}(\backX_{k}, \tk_k)\bigg)\bigg]\\ 
    & \quad + \sqrt{1-\tal_{\tk_{k-1}}-\sigma_{\tk_k}^2}\bV^{\Theta_\nn}(\backX_{k}, \tk_k)+\sigma_{\tk_k} Z_{k}.
\end{split}
\end{equation}
\cfload
 \end{method}

\begin{remark}[Explanations for \cref{setting:DDIM}]
In this remark we provide some intuitive and
theoretical explanations for \cref{setting:DDIM} and roughly explain in what sense the scheme in \cref{setting:DDIM} can be used for generative modelling.
The structure of the scheme remains consistent with \cref{setting:dnn3} due to \cref{ddim:lemma:loss} that permits to use the same training objective as in \cref{prop:ddpm:noise}, up to a constant. The key distinction lies in the sampling phase of the backward process.

    We think of $(\bV^{\theta})_{\theta \in \R^{\fd}}$ as the \ANN\ which is trained to predict the noise component of the noisy data at each time step,
    we think of
        $\fL$ 
    as the loss used in the training,
    we think of
        $\fG$ 
    as the generalized gradient of the loss $\fL$ with respect to the trainable parameters,
    we think of 
        $\datax_{n,i}$, $ n, i \in \N$, 
    as random samples of the initial value of the forward process used for training,
    we think of
        $\mathcal{E}_{n,i}$, $ n, i \in \N$,
    as the noise components of the forward process used for training,
    we think of 
        $\randT_{n}$, $ n\in \N$,
    as random times used to determine which terms of the upper bound are considered in each training step,
    we think of
        $(\Theta_{n})_{n\in\N_0}$
    as the training process for the parameters of the backward process given by an \SGD\ process for the generalized gradient $\fG$
    with learning rate $\gamma$,
    batch size $\ds$, and
    training data
    $(\datax_{n,i}, \mathcal{E}_{n_,i}, \randT_{n})_{(n,i) \in \N^2}$,
    we think of
        $\nn$
    as the number of training steps,
    we think of 
        $K \in \{2,3,\ldots,T\}$
    as the time steps in the backward process,
    we think of
        $Z_k$, $k\in\{1,\ldots,K+1\}$,
    as the noise components of the backward process,
    and we think of
        $\backX$
    as the backward process for the trained parameters $\Theta_{\nn}$.

 Roughly speaking, accordingly to the transition kernels for the backward process (cf.\ \cref{setting_ddim:hp5}), for every $k \in \{1,\ldots,K\}$ three distinct parts can be identified in the backward process:
 \begin{enumerate}[label=(\roman*)]
     \item  $(\sqrt{\tal_{\tk_k}})^{-1}(\backX_{k}-\sqrt{1-\tal_{\tk_k}}\bV^{\Theta_\nn}(\backX_{k}, \tk_k))$ represents the denoised data prediction from the time step $\tk_k$,
     \item $\sqrt{1-\tal_{\tk_{k-1}}-\sigma_{\tk_k}^2}\bV^{\Theta_\nn}(\backX_{k}, \tk_k)$ is the direction pointing back to $\backX_{k}$, and
     \item $\sigma_{\tk_k} Z_{k}$ is the gaussian noise.
 \end{enumerate}
We compute the state at the previous time step by re-scaling the denoised estimate from the current time step
and by summing up a scaled version of the predicted noise. 
To derive the \DDIM\ we assume $\eta=0$, making the denoising process completely deterministic, that is, no new noise is added during the backward process.
This guarantees consistency in the generative phase, ensuring that processes started from the same  initial state of the backward process exhibit similar high-level features
On the other hand, assuming $\eta =1$ reverts the process to the standard \DDPM. There is also the option of choosing $\eta \in (0,1)$, which creates an interpolation between a \DDIM\ and a \DDPM.
Note that although in \cref{setting_ddim} we have for all $t \in \{1,\ldots,T\}$ that $\sigma_t\in(0,1)$ we can approximate the case $\sigma_t=0$ assuming that $0<\sigma_t<<1$.

Moreover, note that we can consider forward processes of length $K\leq T$ as long as the conditional distribution at any time step, given the initial value, follows a Gaussian distribution of the same form as in \cref{ddim:cond_prob}, since, roughly speaking, the training objective depends solely on this.
This allows to accelerate the respective backward process by selecting fewer time steps $\{\tk_0,\tk_1,\ldots,\tk_K\}$ while keeping the number of steps large during training. For a mathematical justification we refer to \cite[Section 4.2]{song2022denoising}.
Under these assumptions, we expect that the terminal value $\backX_0$ of the trained backward process to be distributed according to the desired distribution.
\end{remark}

\subsection{Classifier-free diffusion guidance}\label{sec_classifier_free_guidance}

In the previous sections, the objective of the considered generative methods has been to generate new data points from one underlying distribution based on a dataset from that distribution. 
We now consider the situation where the considered dataset can be divided into multiple subsets, each containing samples coming from different (but possibly related) distributions and the goal is to generate new data points from each of these distributions.

Classifier-free diffusion guidance \cite{ho2022classifierfree} is an improvement of classifier guidance \cite{dhariwal2021diffusion} that uses a classifier to guide a diffusion model to generate data of a desired class.
By eliminating the need for a separate discriminator or classifier, classifier-free diffusion guidance simplifies the model architecture and the training process, leading to a more stable and efficient data generation.
In \cref{sec_classifier_free_guidance:adgn} we introduce \AdaGN\ (cf.\ \cref{def:adagn}), a widely used technique for directly incorporating class information into UNets. Next, we present a simplified training and generation scheme for classifier-free diffusion guidance in \cref{sec_classifier_free_guidance:gen} (cf.\ \cref{setting:classfreeguidance}).

\subsubsection{Controlling with adaptive group normalization}\label{sec_classifier_free_guidance:adgn}

We now consider class conditioning, focusing on how the class information is typically integrated into the \ANN. Roughly speaking, class conditioning refers to
incorporating additional information, in the form of categorical labels or classes, to influence data generation or transformation during the modelling process.
The integration of class conditioning enables a generative model to understand and capture the distinctive features associated with each class, leading to more controlled and targeted generation. 
In literature, numerous methods have been proposed for conveying this information within \ANNs.
Assuming we are using a UNet architecture, at each resolution level (cf.\ \cref{image:unet}), we transform the class information to match the corresponding dimension of that level. 
We then either add it to the time embedding (cf.\ \cref{def:sin_time_emb}) as in \cite{ho2020denoising}, multiply it with the feature maps, or apply \AdaGN\ \cite{dhariwal2021diffusion}, a new normalization technique.
\begin{definition}[Adaptive Group Normalization] 
\label{def:adagn}
Let $D,n,\di \in \N$ satisfy $D n = \di$, let $C,G \in \{1,\ldots,\di\}$, let  $\beta\in \mathbb{R}^{C}, \gamma\in \mathbb{R}^{C }$, $\varepsilon \in(0,1)$.
Then we denote by $\textnormal{AdaGN}_{\beta, \gamma, \varepsilon}^{\di, D} \in C\left(\R^{\di  } \times \R^D \times \R^D, \R^{\di}\right)$ the function which satisfies for all $x \in \R^{\di  }$ and $y^{(1)}=(y^{(1)}_1,\ldots,y^{(1)}_D),y^{(2)}=(y^{(2)}_1,\ldots,y^{(2)}_D) \in \R^D$ that
\begin{equation}
    \textnormal{AdaGN}_{\beta, \gamma, \varepsilon}^{\di, G, D}(x, y^{(1)},y^{(2)}) = \big(y^{(1)}_i \, (\textnormal {Groupnorm}^{\di,G}_{\beta, \gamma, \varepsilon}(x))_{i+Dj}+ y^{(2)}_i\big)_{(i,j)\in \{1,\ldots,D\}\times\{0,1,\ldots,n-1\}}
\end{equation}
(cf.\ definition of $\textnormal {Groupnorm}_{\beta, \gamma, \varepsilon}^{\di,G}$ in \cite[Section\ 3]{wu2018group}) and we call $\textnormal{AdaGN}_{\beta, \gamma, \varepsilon}^{\di,G, D}$ the Adaptive Group Normalization with learnable parameters $\beta$ and $\gamma$, regularization parameter $\varepsilon$, data embedding dimension $\di$, number of groups $G$, and class embedding dimension $D$.
\end{definition}
\begin{remark}[Explanations for \cref{def:adagn}]
    In this remark we provide some explanations for \cref{def:adagn}.
    In \cref{def:adagn} we think of $x $ as the intermediate representation of the input, we think of $y^{(1)}$ as the transformation of the timestep, and we think of $y^{(2)}$ as the transformation of the class information.
    \AdaGN\ is obtained by first applying to the vector $x$ a group normalization \cite{wu2018group}, characterized by learnable parameters $\beta$ and $\gamma$, regularization parameter $\varepsilon$, data embedding dimension $\di$, and number of groups $G$. The result is then multiplied by $y^{(1)}$ and  $y^{(2)}$ is added. To ensure dimension alignment, $y^{(1)}$ and $y^{(2)}$ are repeated $n$ times.
    Authors of \cite{dhariwal2021diffusion} observe that this technique leads to an enhancement of the diffusion model, resulting in an improved \FID\ score.
\end{remark}

\subsubsection{Generative method}\label{sec_classifier_free_guidance:gen}
We now introduce the generative method for \DDPMs\ with class conditioning. In classifier-free diffusion guidance, the model is trained with the class information, allowing control over the generation of different types of data. The scheme was proposed in \cite{ho2022classifierfree}.

\cfclear
\begin{method}[Classifier-free diffusion guidance generative method]
\label{setting:classfreeguidance}
Let $\di ,\fd, \ds, C \in \N$, $T\in \N \backslash\{1\}$,  $\gamma \in (0, \infty)$, $p \in [0,1]$, $\al_1,\ldots,\al_T \in (0,1)$, $\tal_0, \tal_1, \ldots, \tal_T \in (0,1]$, assume for all $t \in \{0,1, \ldots, T\}$ that
$ \tal_t= \textstyle\prod_{s=1}^t\al_s$,
	for every $\theta \in \R^{\fd}$ let 
	$\bV^\theta \colon \R^{\di} \times \{0,1\}^C \times \{1,\ldots,T\} \to\R^{\di}$
	be a function,
let  
$\fL \colon  \R^\fd \times \R^\di \times \R^\di \times \{0,1\}^C \times \{1,\ldots,T\} \to \R$
satisfy 
for all $ \theta \in \R^{ \fd } $, $x, \varepsilon \in \R^\di$, $c \in \{0,1\}^C$, $t \in \{1,\ldots,T\}$
that
\begin{equation} \label{setting:classfreeguidance:loss}
\fL( \theta, x, \varepsilon, c, t )= 
  \bigl\| \varepsilon - \bV^{ \theta }\bigl(  \sqrt{\tal_t} x + \sqrt{1-\tal_t} \varepsilon, c, t \bigr) \bigr\|^2,
\end{equation}
let 
$
  \fG \colon \R^\fd \times \R^\di \times \R^\di \times \{0,1\}^C \times \{1,\ldots,T\} \to \R^{ \fd }
$ 
satisfy for all 
$x, \varepsilon \in \R^\di$,
$c \in \{0,1\}^C$,
$t \in \{1,\ldots,T\}$,
$\theta \in \R^\fd$ with $\fL( \cdot,x, \varepsilon, t) $ differentiable at $\theta$
that
\begin{equation}
  \fG( \theta, x, \varepsilon, c, t ) = 
 ( \nabla_{ \theta } \fL )( \theta, x, \varepsilon, c, t ),
\end{equation}
let $(\Omega, \cF, \mathbbm{P})$ be a probability space, 
let  $\datax_{n,i} \colon \Omega \to \R^\di$, $ n, i \in \N$, be random variables,
let $\noise_{n_
,i} \colon \Omega \to \R^{ \di }$, $ n, i \in \N$, 
be i.i.d.\ standard normal random variables,
let $\bern_{n,i} \colon \Omega \to \{0,1\}$, $ n, i \in \N$, be independent Bernoulli random variables with parameter $p$,
let  $\datac_{n,i} \colon \Omega \to \{0,1\}^C$, $ n, i \in \N$, be random variables,
let  $\randT_{n} \colon \Omega \to \{1,\ldots,T\}$, $ n\in \N$, be independent $\mathcal{U}_{\{1,2,\ldots,T\}}$-distributed random variables, 
let 
$ \Theta  \colon \N_0 \times \Omega \to \R^{ \fd } 
$
be a stochastic process which satisfies 
for all $n\in\N$ that
\begin{equation}
\Theta_{n}=\Theta_{n-1}-\gamma \bigg[\frac{1}{\ds}\sum_{i=1}^{\ds}\fG( \Theta_{n-1}, \datax_{n,i}, \noise_{n,i},\bern_{n,i} \,\datac_{n,i}, \randT_{n} )\bigg],
\end{equation}
let $N\in \N$, $w \in [0,\infty)$, $c \in \{0,1\}^C$, let $Z_t \colon \Omega \to \R^{ \di }$, $t\in\{1,\ldots,T+1\}$,
be i.i.d.\ standard normal random variables,
let $\backX=(\backX_t)_{t \in \{0,1,\ldots,T\}} \colon \Omega \to \R^{ \di }$ be a stochastic process, and assume  for all $t \in \{1,\ldots,T\}$ that
\begin{equation}
    \backX_{T}=Z_{T+1}
\end{equation}
\begin{equation} \label{setting:classfreeguidance:backx}
\begin{split}
    \andq \backX_{t-1} & = \frac1{\sqrt{\al_t}}\bigg(\backX_t-\frac{1-\al_t}{\sqrt{1-\tal_t}}\Big((1+w)\bV^{\Theta_N}(\backX_t, c,t)-w \, \bV^{\Theta_N}(\backX_t, 0,t)\Big)\bigg) \\
    & \quad + \sqrt{\left[\frac{1-{\tal}_{t-1}}{1-{\tal}_t}\right](1-\al_t)}Z_t.
\end{split}
\end{equation}
\cfload
\end{method}
\begin{remark}[Explanations of \cref{setting:classfreeguidance}]
    In this remark, we offer intuitive explanations for \cref{setting:classfreeguidance}, outlining how this scheme can be applied to generate data of different classes.
    We also refer to \cref{setting:dnn3} for explanations of fundamentals aspects of \DDPMs.
    
    We think of $(\bV^{\theta})_{\theta \in \R^{\fd}}$ as the \ANN\ which is trained to predict the noise component of the noisy data at each time step,
    we think of
        $\fL$ 
    as the loss used in the training,
    we think of
        $\fG$ 
    as the generalized gradient of the loss $\fL$ with respect to the trainable parameters,
    we think of 
        $\datax_{n,i}$, $ n, i \in \N$, 
    as random samples of the initial value of the forward process used for training,
    we think of 
        $\bern_{n,i} $, $ n, i \in \N$,
    as the Bernoulli random variables with probability $p$,
    we think of 
        $\datac_{n,i}$, $ n, i \in \N$, 
    as the one hot encoded vectors of the class information,
    we think of
        $\mathcal{E}_{n,i}$, $ n, i \in \N$,
    as the noise components of the forward process used for training,
    we think of 
        $\randT_{n}$, $ n\in \N$,
    as random times used to determine which terms of the upper bound are considered in each training step,
    we think of
        $(\Theta_{n})_{n\in\N_0}$
    as the training process for the parameters of the backward process given by an \SGD\ process for the generalized gradient $\fG$
    with learning rate $\gamma$,
    batch size $\ds$, and
    training data
    $(\datax_{n,i}, \mathcal{E}_{n_,i}, \bern_{n,i}\datac_{n,i}, \randT_{n})_{(n,i) \in \N^2}$,
    we think of
        $\nn$
    as the number of training steps,
    we think of
        $Z_t$, $t\in\{1,\ldots,T\}$,
    as the noise components of the backward process,
    and we think of
        $\backX$
    as the backward process for the trained parameters $\Theta_{\nn}$.
    
    Here the model $(\bV^\theta)_{\theta \in \R^{\fd} } $ requires three inputs: the noisy data, the class information, and the time step. The class information is provided as a one hot encoded vector of size $C$ where $C$ represents the number of classes. 
    The optimization process is slightly adjusted so that the model is effectively trained to generate data with and without class information. The number $p \in [0,1]$ defines the chances of replacing the one hot encoded vector $\datac_{n,i}$ with the zero vector, forcing the model to learn how to generate data also without class information. An optimal value for $p$ was determined to be $0.1$ or $0.2$, indicating that either $10\%$ or $20\%$ of the data will not be associated with any classes during the training.
    The backward process slightly differs from the one described in \cref{setting:dnn3}.
    After training the model, to generate a new data for the class $c$, 
    we interpolate the noise prediction given the desired class $\bV^{\Theta_N}(\backX_t, c, t)$ with the noise prediction without the class information $\bV^{\Theta_N}(\backX_t, 0,t)$.
    If $w=0$ the sampling phase coincides with a \DDPM\ with class information. 
    When $w \in (0,\infty)$ classifier-free diffusion guidance is applied. 
    Note that to strengthen the class information, the signal of the model without class information is removed. Theoretically, the more information without class is removed, the more information of the desired class is obtained.
    Given these assumptions we expect that the terminal value $\backX_0$ of the trained backward process will be roughly aligned with the class distribution we aim to sample from.
\end{remark}

\subsection{Stable Diffusion}\label{sec_stable_diffusion}
Stable diffusion model \cite{9878449} achieved state of the art results on image generation by combining diffusion model and autoencoder. In contrast to other works, this approach is able to manage high dimensional data 
limiting the demand of computational resources. It is primarily used to generate detailed images conditioned on text descriptions, but it can also be applied to other tasks such as inpainting, outpainting, and generating image-to-image translations guided by a text prompt.
Stable diffusion code and model weights have been released publicly, permitting further development.
In \cref{sec:cross_attention:layer} we define the Cross Attention layer (cf.\ \cref{setting:crossattention}), the mechanism by which word conditioning is incorporated into UNets and in
\cref{sec:stable_diff:gen}, coherently with the previous subsections, we introduce the generative method for stable diffusion (cf.\ \cref{setting:stable}).

\subsubsection{Controlling with cross attention layer}\label{sec:cross_attention:layer}
We now analyze how the encoded text data are used to influence the generation or transformation of data. 
The implementation of words conditioning  typically involves encoding the input words into a suitable representation, and then incorporating this information at each step of the diffusion process. 
The model learns to use the meaning of the input words to shape the data, ensuring the generated output fits the given context.
Nowadays many state of the art models use this technology (cf., \eg  \cite{9878449,pmlr-v139-ramesh21a,ramesh2022hierarchical,nichol2022glide,saharia2022photorealistic}).
Assuming a UNet architecture is used, the encoded texts are usually mapped to each intermediate level (cf.\ \cref{image:unet}). A common technique to pass this information to the model is cross attention \cite{lin2021catcrossattentionvision}, a variant of self-attention \cite{NIPS2017_3f5ee243}. We now present it.

\begin{definition}[Cross attention layer]
\label{setting:crossattention}
    Let $\di, \emb,\maxtok,\context,\headsdim,\heads \in \N$, $\inx \in \R^{\di \times \emb}$, $\iny \in \R^{\maxtok \times \context}$, 
    $W^Q=(W^Q_1,\ldots,W^Q_{\heads}) \in (\R^{\emb \times \headsdim})^{\heads}$, $W^K=(W^K_1,\ldots,W^K_{\heads}), W^V=(W^V_1,\ldots,W^V_{\heads}) \in (\R^{\context \times \headsdim})^\heads$,
    $Q=\allowbreak(Q_1,\ldots,Q_{\heads}) \in(\R^{\di \times \headsdim})^\heads $, $K=(K_1, \ldots,K_{\heads}), \allowbreak V=(V_1,\ldots,V_{\heads}) \in(\R^{\maxtok \times \headsdim})^\heads $ 
    satisfy for all $i \in \{1,\ldots,\heads\}$ that 
    $Q_i=\inx W^Q_i$, $K_i=\iny W^K_i$, and $V_i=\iny W^V_i$, and let $A \in \R^{\headsdim\heads \times \emb}$.
    Then we say that $\textnormal{crossatt}$ is the cross attention for the query $Q$ with weight matrix $W^Q$, the key $K$ with weight matrix $W^K$, the value $V$ with weight matrix $W^V$, the input data $\inx$, the encoded text  $\iny$, and the linear transformation $A$ if and only if $\textnormal{crossatt}$ is the matrix in $\R^{\di \times \emb}$ which satisfies
    \begin{equation}\label{setting:crossattention:attention}
        \textnormal{crossatt}=\left( \left(\textnormal{softmax}\left( \frac{ Q_1 K_1^*}{\sqrt{\headsdim}} \right)\right)V_1, \ldots, \left(\textnormal{softmax}\left( \frac{ Q_{\heads} K_{\heads}^*}{\sqrt{\headsdim}} \right)\right)V_{\heads} \right) A.
    \end{equation}
\end{definition}

\begin{remark}[Explanations for \cref{setting:crossattention}]
    In this remark we provide some explanations for \cref{setting:crossattention}. 
    In \cref{setting:crossattention} we think of $\di$ as the number of entries (or a latent representation of that number) of the input $\inx$ and we think of $\emb$ as the number of channel (or a latent representation of that number) of $\inx$ which is also referred to as the embedding size.
    Moreover, we think of $\context$ as the context dimension of the 
    encoded text or token embedding $\iny$. 
    Each token (a single unit of text) is represented as a vector of this length.
    Additionally, we think of $\maxtok$ as the maximum number of tokens allowed, defining the length limit of the text input that the model can handle.
    Next, we think of $\heads$ as the number of attention heads in a multi-head attention mechanism, each head independently processes the input and captures different aspects of the text's information (cf., \eg \cite{NIPS2017_3f5ee243}). Finally, we think of $\headsdim$ as the dimension of each head,
    which defines the size of the vector space in which each attention head operates.
    The query matrix $Q$ is computed by multiplying the input $\inx$ by the weight matrix $W^Q$. Similarly, the key matrix $K$ and the value matrix $V$ are derived from the encoded text  $\iny$ through the weight matrices $W^K$ and $W^V$. 
    The matrices $Q$, $K$, and $V$ are utilized to compute the attention using \cref{setting:crossattention:attention}. An optional linear transformation $A$ can be used to return to the initial dimensions of $\inx$. See \cite{NIPS2017_3f5ee243}  for more in depth treatment of $\textnormal{crossatt}$.
\end{remark}

\subsubsection{Generative method}\label{sec:stable_diff:gen}
We now formulate the generative method for stable diffusion with text conditioning. This approach is essential for learning the relationship between the text and the data, guiding the generation of new samples. The scheme was proposed in \cite{9878449}.

\cfclear
\begin{method}[Stable diffusion generative method]
\label{setting:stable}
Let $\Di,L,c,\di,l,\fd, \ds \in \N$, $T\in \N \backslash \{1\}$, $\gamma \in (0, \infty)$, $\al_1,\ldots,\al_T \in (0,1)$, $\tal_0, \tal_1, \ldots, \tal_T \in (0,1]$, assume for all $t \in \{1, \ldots, T\}$ that $\tal_t= \textstyle\prod_{s=1}^t\al_s$,
let $\cE \colon \R^{\Di} \to \R^{\di} $ be a function, let $\cD \colon \R^{\di} \to \R^{\Di}$ be a function, for every $\theta$ let $\tau^{\theta} \colon \{1,\ldots,L\}^{l} \to \R^{l \times c}$ be a function,
for every $\theta \in \R^{\fd}$ let 
$\bV^\theta \colon \R^{\di} \times \R^{l \times c} \times \{1,\ldots,T\} \to \R^{\di}$
be a function,
let  
$\fL \colon \R^\fd \times \R^\di \times \R^\di \times \R^{l \times c} \times \{1,\ldots,T\} \to \R$
satisfy 
for all $ \theta \in \R^{ \fd } $, $x, \varepsilon \in \R^\di$, $y \in \{1,\ldots, L\}^l$, $t \in \{1,\ldots,T\}$
that
\begin{equation}\label{setting:stable:loss}
\fL( \theta, x, \varepsilon, y, t)= 
  \bigl\|\varepsilon - \bV^{ \theta }\bigl( 
  \sqrt{\tal_t} \cE(x) + \sqrt{1-\tal_t} \varepsilon
,\tau^{\theta}(y), t \bigr)  \bigr\|^2,
\end{equation}
let 
$
  \fG \colon \R^\fd \times \R^\di \times \R^\di \times \R^{l \times c} \times \{1,\ldots,T\} \to \R^{ \fd }
$ 
satisfy for all 
$x, \varepsilon \in \R^\di$,
$y \in \{1,\ldots, L\}^l$,
$t \in \{1,\ldots,T\}$, 
$\theta \in \R^\fd$ with $\fL( \cdot,x, \varepsilon, t) $ differentiable at $\theta$
that
\begin{equation}
  \fG( \theta, x, \varepsilon, y, t ) = 
 ( \nabla_{ \theta } \fL )( \theta, x, \varepsilon, y, t ),
\end{equation}
let $(\Omega, \cF, \mathbbm{P})$ be a probability space, 
let  $\datax_{n,i} \colon \Omega \to \R^\di$, $ n, i \in \N$, be random variables,
let $\noise_{n_,i} \colon \Omega \to \R^{\di}$, $ n, i \in \N$, be i.i.d.\ standard normal random variables,
let  $\datal_{n,i} \colon \Omega \to \{1,\ldots, L\}^l$, $ n, i \in \N$, be random variables,
 let  $\randT_{n} \colon \Omega \to \{1,\ldots,T\}$, $ n\in \N$, be independent $\mathcal{U}_{\{1,2,\ldots,T\}}$-distributed random variables,  
let 
$ \Theta  \colon \N_0 \times \Omega \to \R^{ \fd } 
$
be a stochastic process which satisfies 
for all $n\in\N$ that
\begin{equation}
\Theta_{n}=\Theta_{n-1}-\gamma \bigg[\frac{1}{\ds}\sum_{i=1}^{\ds}\fG( \Theta_{n-1}, \datax_{n,i}, \noise_{n,i},\datal_{n,i}, \randT_{n} )\bigg],
\end{equation}
let $N\in \N$, $y \in \{1,\ldots, L\}^l$, let $Z_t \colon \Omega \to \R^{\di}$, $t\in\{1,\ldots,T+1\}$,
be i.i.d.\ standard normal random variables,
let $\eta,\sigma_1,\ldots,\sigma_t \in [0,1]$ satisfy for all $t\in\{1,\ldots,T\}$ that $\sigma_t=\allowbreak \eta \sqrt{(1 - \al_t )(1 - \tal_{t-1})} \allowbreak \sqrt{ (1 - \tal_t)^{-1} }$,
let $\backX=(\backX_t)_{t \in \{0,1,\ldots,T\}} \colon \Omega \to \R^{\di}$ be a stochastic process, and assume for all $t \in \{1,\ldots,T\}$ that
\begin{equation}
    \backX_T=Z_{T+1}
\end{equation}
\begin{equation} \label{setting:stable:backx}
\begin{split}
    \andq \backX_{t-1} & = \sqrt{\tal_{t-1}} \bigg[\frac{1}{\sqrt{\tal_{t}}}\bigg(\backX_t-\sqrt{1-\tal_t}\bV^{\Theta_N}(\backX_t,\tau^{\Theta_N}(y), t)\bigg)\bigg]\\
    & \quad + \sqrt{1-\tal_{t-1}-\sigma_t^2}\bV^{\Theta_N}(\backX_t,\tau^{\Theta_N}(y), t)+\sigma_t z_t.
\end{split}
\end{equation}
\cfload
\end{method}

\begin{remark}[Explanations for \cref{setting:stable}]
In this remark we provide some intuitive and
theoretical explanations for \cref{setting:stable} along with an overview of the principles behind stable diffusion data generation.
Roughly speaking, the stable diffusion model consists of three parts: the autoencoder made up of the encoder $\cE  $ and the decoder $\cD$, the \ANN\ $(\bV^{\theta})_{\theta \in \R^{\fd}}$, and the text encoder $(\tau^{\theta})_{\theta \in \R^{\fd}} $.
Diffusion models typically operate directly in pixel space,  consuming hundreds of GPU and the inference phase is expensive due to sequential evaluations.
Here we work in the latent space of pretrained autoencoders limiting the computational resources needed without losing resolution quality.  Imperceptible details are abstracted away while the most important information is kept.
Authors of \cite{9878449} train autoencoder models in an adversarial manner \cite{goodfellow2014generative}, such that a discriminator is optimized to differentiate original data from reconstructions.
To avoid arbitrarily scaled latent spaces, they
regularize the latent space to be zero centered and obtain small variance by introducing a regularizing loss. They experiment two different kinds of regularizations.
For text to image modelling, they choose a KL-penalty towards a standard normal on the learned latent. 
Note that in our generative method, the autoencoder is pretrained in an earlier phase and remains frozen.

    We think of $(\bV^{\theta})_{\theta \in \R^{\fd}}$ as the \ANN\ which is trained to predict the noise component of the noisy data at each time step,
    we think of
        $\fL$ 
    as the loss used in the training,
    we think of
        $\fG$ 
    as the generalized gradient of the loss $\fL$ with respect to the trainable parameters,
    we think of 
        $\datax_{n,i}$, $ n, i \in \N$, 
    as random samples of the initial value of the forward process used for training,
    we think of
    $\datal_{n,i}$, $ n, i \in \N$,
    as the labels of the random samples,
    we think of
        $\mathcal{E}_{n,i}$, $ n, i \in \N$,
    as the noise components of the forward process used for training,
    we think of 
        $\randT_{n}$, $ n\in \N$,
    as random times used to determine which terms of the upper bound are considered in each training step,
    we think of
        $(\Theta_{n})_{n\in\N_0}$
    as the training process for the parameters of the backward process given by an \SGD\ process for the generalized gradient $\fG$
    with learning rate $\gamma$,
    batch size $\ds$, and
    training data
    $(\datax_{n,i}, \datal_{n,i}, \mathcal{E}_{n_,i}, \randT_{n})_{(n,i) \in \N^2}$,
    we think of
        $\nn$
    as the number of training steps,
    we think of
        $Z_t$, $t\in\{1,\ldots,T\}$,
    as the noise components of the backward process,
    and we think of
        $\backX$
    as the backward process for the trained parameters $\Theta_{\nn}$.

The output of the text encoder $(\tau^{\theta})_{\theta \in \R^{\fd}}$ is an additional input of the model $(\bV^{\theta})_{\theta \in \R^{\fd}}$. Assuming a UNet architecture, the encoded text data is commonly mapped to each intermediate level (cf.\ \cref{image:unet}) via a cross-attention layer (cf.\ $\cref{setting:crossattention}$). 
Note that the text information  belongs to $\{1,\ldots,L\}^l$ where we think of $L$ as the total number of possible tokens and we think of $l$ as the length limit of the number of tokens allowed as input.

The major achievement of this scheme is the capability to generate high quality data conditioned on text descriptions. Specifically, we expect that the terminal value of the backward process $\backX_0 \in \R^{\di}$, when passed to the decoder $\cD$, will produce the data $\cD(\backX_0)$ that aligns with the given text prompt $y$ and the distribution from which we aim to sample.
\end{remark}

\subsection{Further state of the art diffusion techniques}\label{sec_stae_of_art}
We now explore further state of the art diffusion techniques.
We will focus on various diffusion models: GLIDE \cite{dhariwal2021diffusion} in \cref{sec_stae_of_art:glide},  DALL-E 2 and DALL-E 3 \cite{ramesh2022hierarchical,betker2023improving} in \cref{sec_stae_of_art:dalle}, and Imagen \cite{saharia2022photorealistic} in \cref{sec_stae_of_art:imagen}. 
Below, we roughly describe the advancements that distinguish these diffusion models.

\subsubsection{GLIDE}\label{sec_stae_of_art:glide}
GLIDE \cite{nichol2022glide} is a model that combines the capabilities of text-to-image generation and image editing, aiming to create realistic images that align with textual descriptions.

The authors employ the upsampling diffusion model architecture proposed in \cite{dhariwal2021diffusion}, which consists of $3.5$ billion parameters, with certain modifications. One key improvement is the inclusion of text captions as an additional input to the model.
They compare two methods for guiding diffusion models with text prompts: CLIP guidance \cite{DBLP00020, kim2022diffusionclip} and classifier-free diffusion guidance \cite{ho2022classifierfree}. Based on both human assessments and automated evaluations, they observe that the classifier-free diffusion guidance approach generates higher quality images.
To adapt classifier-free diffusion guidance for text, they encode the text prompt into tokens, pass these tokens into a Transformer architecture, and use the last token embedding as the encoded text information. Additionally, all output tokens are concatenated with the attention context at each level of UNets' attention layers. This effectively integrates the text into the generation process, allowing the model to guide image creation according to the meaning of the text.

\subsubsection{DALL-E 2 and DALL-E 3}\label{sec_stae_of_art:dalle}
DALL-E \cite{pmlr-v139-ramesh21a}, the first model in the DALL-E family developed by OpenAI, generates images based on text prompts, producing visuals that correspond closely to the provided descriptions. However, unlike its successors, DALL-E 2 and DALL-E 3, the original DALL-E does not utilize a diffusion model.

In 2022 OpenAI released DALL-E 2 \cite{ramesh2022hierarchical}, a $3.5$ billion parameters text-to-image model, surprisingly smaller than its predecessor ($12$ billion parameters). Despite its size, DALL-E 2 generates higher resolution images than DALL-E.
DALL-E 2 possesses the capability to modify existing images, generate variations that retain key features, and interpolate between two given images. 

DALL-E 2 consists of a prior model that generates an image embedding from a text embedding and a decoder that generates an image based on the image embedding. The text embeddings are derived from CLIP \cite{DBLP00020}, another model developed by OpenAI to select the most appropriate caption for a given image.
CLIP, composed of a text and an image encoder, is trained on a large collection of image-text pairs, maximizing the cosine similarity between their embeddings and remains frozen during the training of DALL-E 2.
The prior model utilizes the CLIP text embedding generated by the CLIP text encoder  from the provided prompt and is trained to predict the corresponding CLIP image embedding.
In \cite{ramesh2022hierarchical} authors explore two different options for the prior.
The first is an autoregressive prior where the CLIP image embedding is converted into discrete codes and then predicted autoregressively, conditioned on the caption and the CLIP text embedding.
The second is the diffusion prior, where a decoder-only Transformer predicts the denoised CLIP image embedding. In this approach, the Transformer processes a sequence that includes the encoded text, the CLIP text embedding, an embedding for the diffusion timestep, and the noisy CLIP image embedding. A final placeholder embedding is also included in the sequence, with the Transformer output at this position used to predict the denoised CLIP image embedding.
While both priors yielded comparable performance, the diffusion prior is more computationally efficient.
The last phase generates the actual image using the decoder, a modified version of another OpenAI diffusion model named GLIDE \cite{nichol2022glide}, cf.\ \cref{sec_stae_of_art:glide}. 
GLIDE was adapted by adding the CLIP image embeddings derived from the prior to the timestep embeddings and by projecting the CLIP image embeddings into four additional context tokens that are concatenated with the GLIDE text encoder’s output sequence, enhancing conditioning on the input text.
The decoder produces images at $64\times64$ pixels, which are upsampled in two stages to a final resolution of $1024 \times 1024$ pixels.
Although the presence of the prior may seem unnecessary, the authors show that training the decoder using only text or CLIP text embeddings alone reduces the image quality.

 In September 2023, OpenAI announced the newest version in the DALL-E series, known as DALL-E 3 \cite{betker2023improving}. The focus is no longer on the improvement of the model but on the caption.
 Authors realized that existing text-to-image models struggle with detailed image descriptions due to noisy and inaccurate image captions in the training dataset. 
 Therefore, a custom image captioner is trained and used to recaption a training dataset, which leads to improved and detailed prompts.
 This challenge can be addressed using large language models, \eg \cite{achiam2023gpt}, capable of expanding brief prompts to more detailed and informative ones.
  DALL-E 3 is trained with $95\%$ synthetic captions and $5\%$ ground truth captions.
 As shown in \cite{betker2023improving}, DALL-E 3 outperforms other text-to-image generation models in various evaluation metrics and benchmarks. 
 Unfortunately, OpenAI shared only high-level information and capabilities of the models, detailed architectural specifications have not been provided.

\subsubsection{Imagen}\label{sec_stae_of_art:imagen}
Similar to GLIDE \cite{nichol2022glide} and DALL-E 2 \cite{ramesh2022hierarchical}, Imagen \cite{saharia2022photorealistic} is a diffusion model with an architecture similar to GLIDE, involving the use of a text embedding to generate images from noise.
A significant discovery highlighted in \cite{saharia2022photorealistic} underscores the value of incorporating large, pre-trained language models (\eg T5 \cite{JMLR:v21:20-074}) that are trained on text-only data. This integration proves to be highly beneficial in deriving text representations for the synthesis of images from textual prompts.
Expanding on this observation, the authors analyze the impact of scaling the text encoder. Their investigation reveals that scaling the size of the language models contributes more significantly to improve results than scaling the size of the diffusion model itself. 
Additionally, the authors introduce a novel technique aimed at preventing saturated pixels in images generated through classifier-free diffusion guidance. A challenge associated with this guidance approach arises when the guidance weight is large, in such cases, pixels may reach saturation, compromising image quality to better align with text. To address this concern, the authors propose the incorporation of dynamic thresholding. 
In this method, saturated pixels are dynamically adjusted within the range of $[-1, 1]$. The magnitude of these adjustments is determined individually at each sampling step (hence, being dynamic), contributing to the adaptability of the process.
The authors assert that this dynamic thresholding yields substantial improvements in both photorealism and the alignment of images with textual guidance, particularly in scenarios involving high guidance during image generation.
Another important contribution in \cite{saharia2022photorealistic} is the introduction of DrawBench a challenging benchmark for text-to-image models that permits to compare and evaluate different generative models.

\subsection*{Acknowledgements}
This work has been partially funded by the National Science Foundation of China (NSFC) under
grant number 12250610192.
Moreover, we gratefully acknowledge the Cluster of Excellence EXC 2044-390685587, Mathematics Münster: Dynamics-Geometry-Structure funded by the Deutsche Forschungsgemeinschaft (DFG, German Research Foundation).

\newpage
{\small
\bibliography{ref}
\bibliographystyle{acm}}
\end{document}